\documentclass[10pt,journal,compsoc]{IEEEtran}

\ifCLASSOPTIONcompsoc
  \usepackage[nocompress]{cite}
\else
  \usepackage{cite}
\fi

\usepackage{amssymb}
\usepackage{multirow}
\usepackage{amsmath}
\usepackage{booktabs}
\usepackage{multirow}
\usepackage{graphicx}
\usepackage{subcaption}
\usepackage{hyperref}
\hypersetup{
    colorlinks=true,
    linkcolor=blue,
    filecolor=magenta,      
    urlcolor=cyan,
    }

\ifCLASSINFOpdf
\else
\fi
\begin{document}
%
\title{TIE-KD: Teacher-Independent and Explainable Knowledge Distillation for Monocular Depth Estimation}
%
%
%
%


\author{Sangwon Choi, Daejune Choi, Duksu Kim \\
\textit{\small Korea University of Technology and Education (KOREATECH)}
\IEEEcompsocitemizethanks{\IEEEcompsocthanksitem Sangwon Choi, Daejune Choi, Duksu Kim are with the Department
of Computer Engineering, Korea University of Technology and Education (KOREATECH), Cheonan,
Korea, 31253.\protect\\
E-mail: see \href{https://sites.google.com/view/hpclab/people}{http://hpc.koreatech.ac.kr}
}
}

\IEEEtitleabstractindextext{%
\begin{abstract}
Monocular depth estimation (MDE) is essential for numerous applications yet is impeded by the substantial computational demands of accurate deep learning models.
To mitigate this, we introduce a novel Teacher-Independent Explainable Knowledge Distillation (TIE-KD) framework that streamlines the knowledge transfer from complex teacher models to compact student networks, eliminating the need for architectural similarity.
The cornerstone of TIE-KD is the Depth Probability Map (DPM), an explainable feature map that interprets the teacher's output, enabling feature-based knowledge distillation solely from the teacher's response.
This approach allows for efficient student learning, leveraging the strengths of feature-based distillation.
Extensive evaluation of the KITTI dataset indicates that TIE-KD not only outperforms conventional response-based KD methods but also demonstrates consistent efficacy across diverse teacher and student architectures.
The robustness and adaptability of TIE-KD underscore its potential for applications requiring efficient and interpretable models, affirming its practicality for real-world deployment.
The code and pre-trained models associated with this research are available at \href{https://github.com/HPC-Lab-KOREATECH/TIE-KD.git}{here}.

\end{abstract}

\begin{IEEEkeywords}
lightweight deep learning, knowledge distillation, explainable feature map, depth estimation
\end{IEEEkeywords}}

\maketitle

\IEEEdisplaynontitleabstractindextext

%
\IEEEpeerreviewmaketitle

\ifCLASSOPTIONcompsoc
\IEEEraisesectionheading{\section{Introduction}\label{sec:introduction}}
\else
\section{Introduction}
\label{sec:introduction}
\fi


\begin{figure}[] 
    \centering
    \resizebox{0.5\textwidth}{!}{%
        \begin{tabular}
            {c@{\hspace{0.5em}}@{\hspace{0.5em}}c@{\hspace{0.5em}}c} \vspace{4pt}            
            \begin{tabular}{@{}c@{}} \bf \large Input \\ \bf \large image \end{tabular}& \raisebox{-0.4\height}{\includegraphics[width=0.45\textwidth]{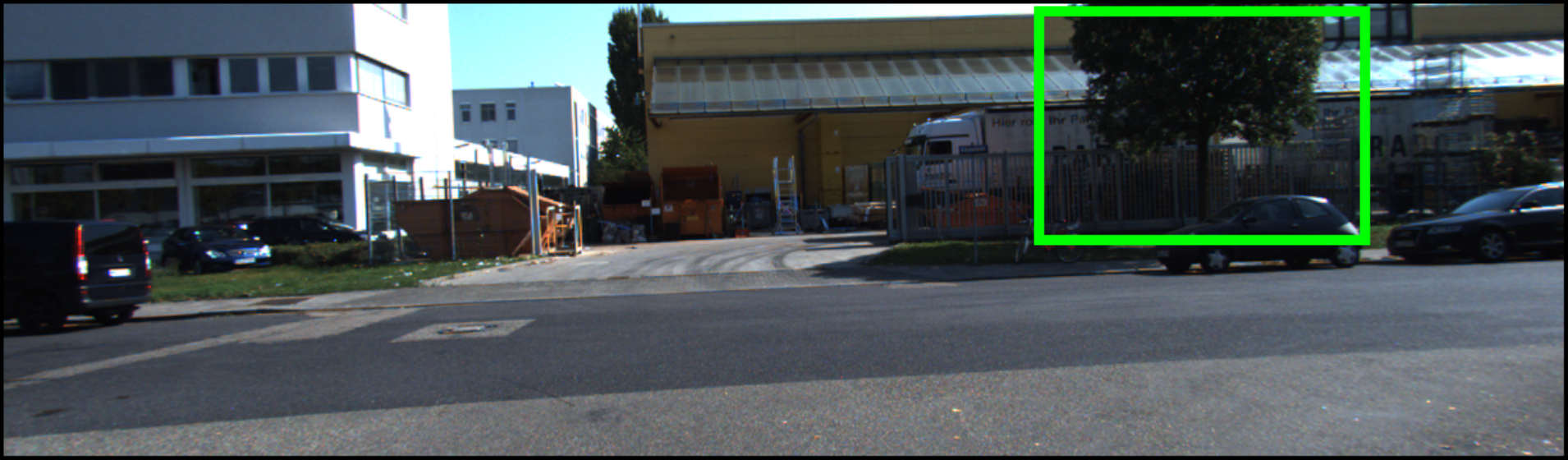}} & \raisebox{-0.4\height}{\includegraphics[width=0.184\textwidth]{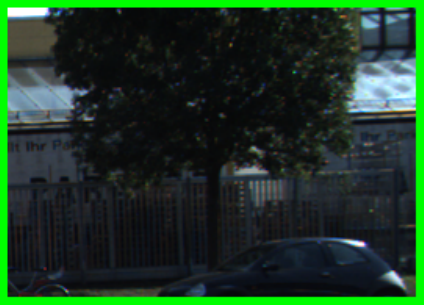}}\\ \vspace{4pt}
            \begin{tabular}{@{}c@{}} \bf \large Baseline \\ \bf \large (17.6 M) \end{tabular}& \raisebox{-0.4\height}{\includegraphics[width=0.45\textwidth]{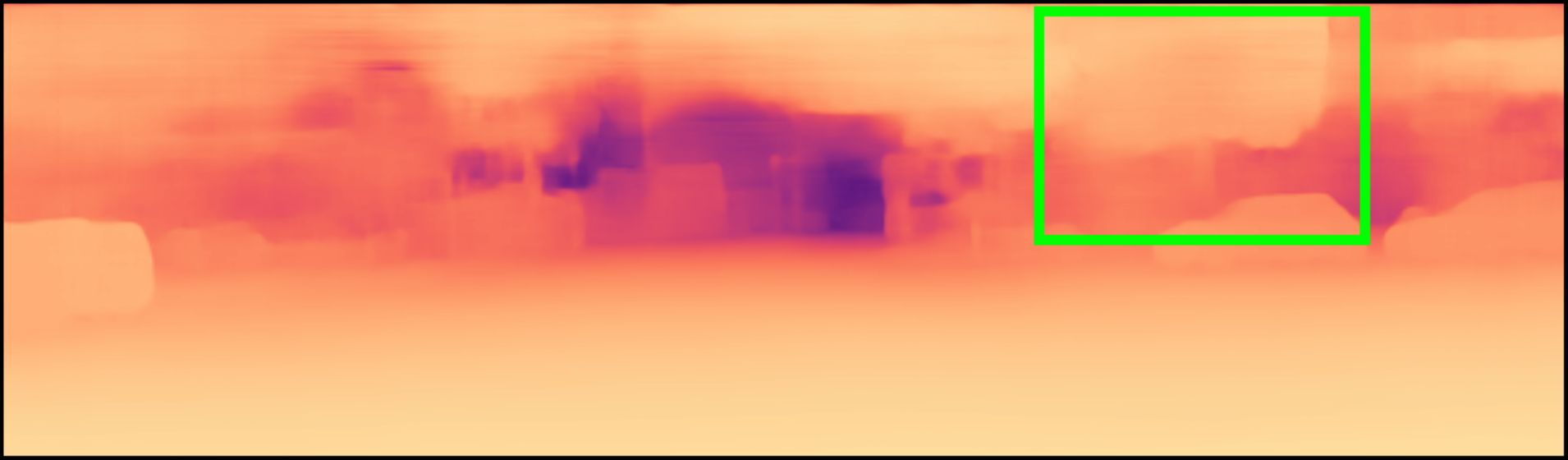}} & \raisebox{-0.4\height}{\includegraphics[width=0.184\textwidth]{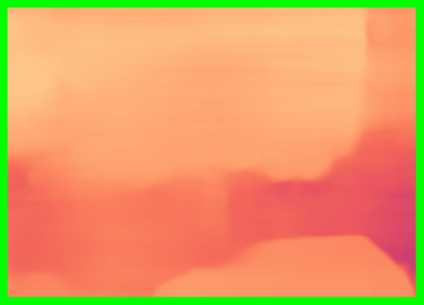}}\\ \vspace{4pt}
            \begin{tabular}{@{}c@{}} \bf \large Teacher \\ \bf \large (78 M) \end{tabular}& \raisebox{-0.4\height}{\includegraphics[width=0.45\textwidth]{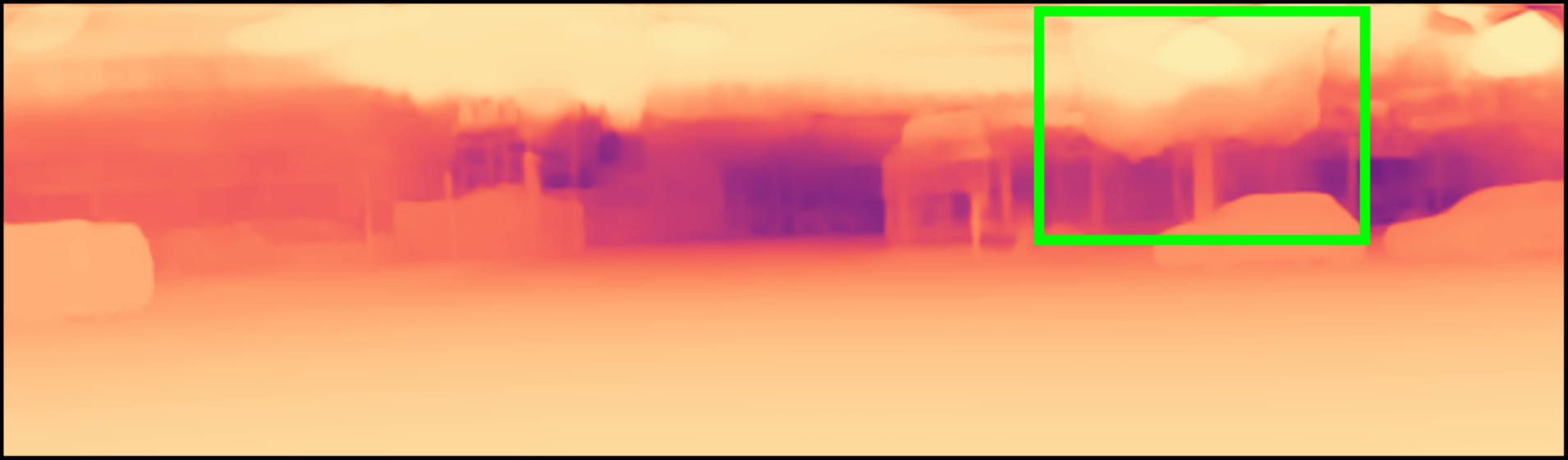}} & \raisebox{-0.4\height}{\includegraphics[width=0.184\textwidth]{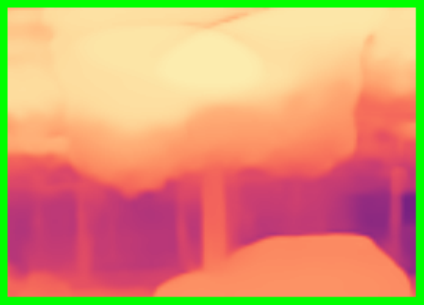}}\\ \vspace{4pt}
            \begin{tabular}{@{}c@{}} \bf \large Res-KD \\ \bf \large (17.6 M) \end{tabular}& \raisebox{-0.4\height}{\includegraphics[width=0.45\textwidth]{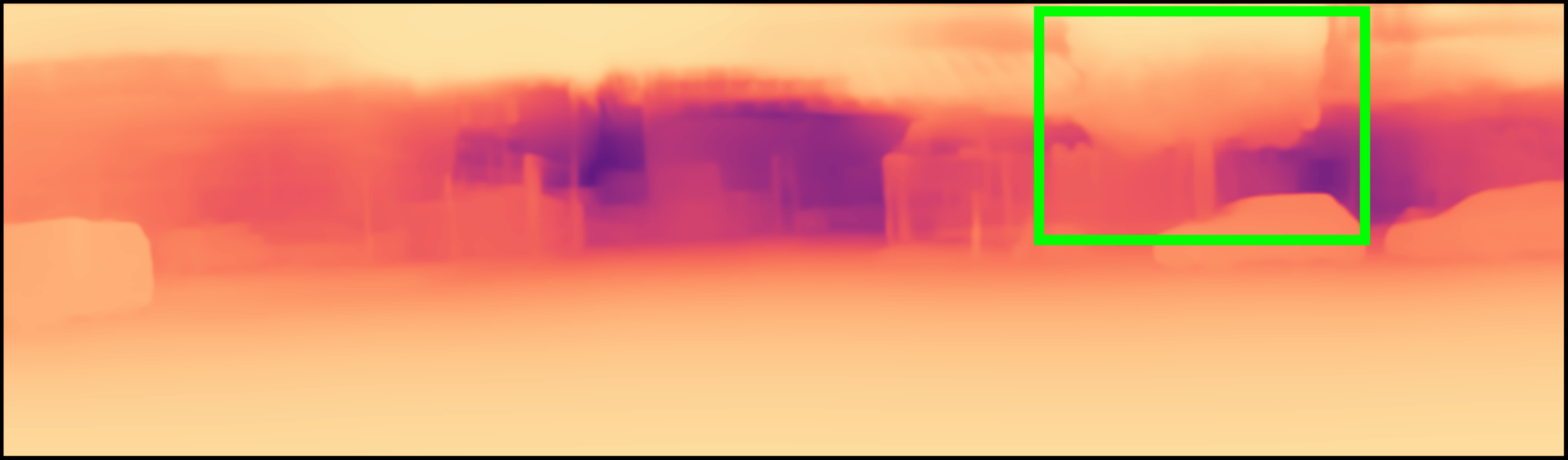}} & \raisebox{-0.4\height}{\includegraphics[width=0.184\textwidth]{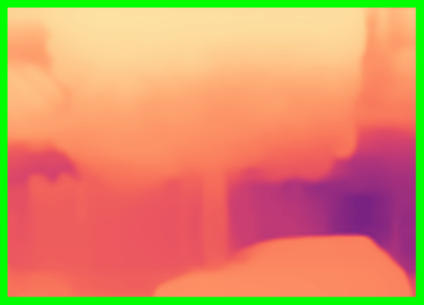}}\\ \vspace{4pt}
            \begin{tabular}{@{}c@{}} \bf \large TIE-KD \\ \bf \large (17.6 M) \end{tabular}& \raisebox{-0.4\height}{\includegraphics[width=0.45\textwidth]{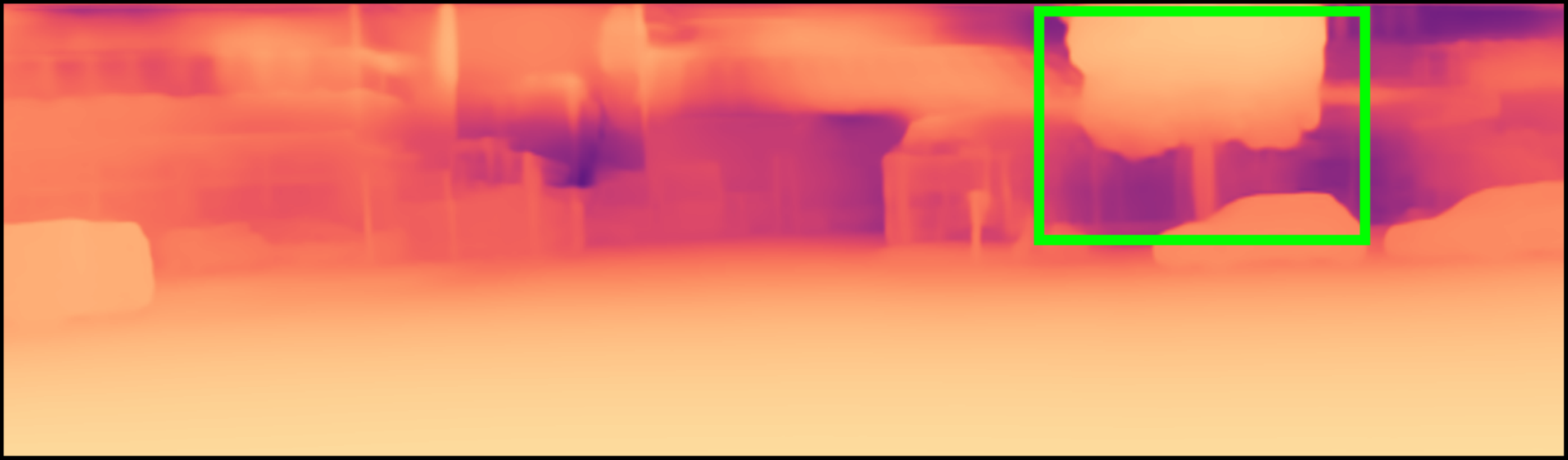}} & \raisebox{-0.4\height}{\includegraphics[width=0.184\textwidth]{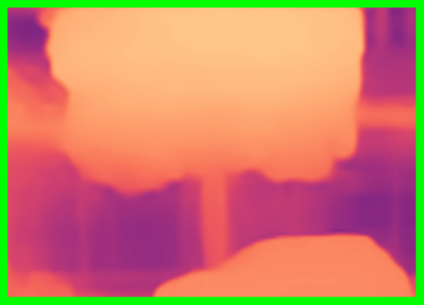}}          
        \end{tabular}%
    }

    \caption{Comparative visualization of depth estimation results showcasing the effectiveness of the proposed TIE-KD framework.
    The first row displays input images; the second row depicts outcomes from a baseline small model trained on ground truth. The third row shows results from the high-capacity teacher model, AdaBins~\cite{bhat2021adabins}. The fourth and fifth rows illustrate depth maps from students trained via a response-based KD and our TIE-KD, respectively.
    Our TIE-KD demonstrates a more effective knowledge distillation performance than prior response-based KD methods, achieving greater similarity to the teacher model, especially in preserving edge definition and depth accuracy.
    } 
    \label{fig:teaser}
\end{figure}

Monocular depth estimation (MDE) is pivotal in computer vision, with applications ranging from autonomous vehicles~\cite{geiger2013vision} to robotics~\cite{dudek2010computational} and 3D modeling~\cite{newcombe2011kinectfusion}. The integration of deep learning has notably enhanced MDE accuracy and efficiency ~\cite{eigen2014depth, laina2016deeper}.
However, state-of-the-art models like SQLdepth~\cite{wang2023sqldepth}, with 242 million parameters, pose challenges for real-time applications due to their computational demands.
Methods like parameter pruning~\cite{wu2016quantized, han2015learning}, low-rank factorization~\cite{yu2017compressing}, and compact convolution filters~\cite{zhai2016doubly} aim to alleviate this.

Knowledge Distillation (KD) is another strategy that efficiently condenses the knowledge from larger models into more compact ones~\cite{hinton2015distilling}.
Initially prevalent in classification tasks~\cite{hinton2015distilling, romero2014fitnets, zagoruyko2016paying, kim2018paraphrasing, heo2019knowledge, chen2021cross, zhao2022decoupled}, KD has expanded into other domains such as object detection~\cite{li2017mimicking}, visual odometry~\cite{saputra2019distilling}, and so on.
KD approaches are typically divided into response-based ~\cite{hinton2015distilling, zhao2022decoupled}, leveraging teacher outputs, and feature-based~\cite{romero2014fitnets, zagoruyko2016paying, heo2019knowledge, kim2018paraphrasing, chen2021cross}, where the student mimics the teacher's feature maps, often resulting in superior performance.
However, feature-based KD presents alignment challenges and typically requires similar network architectures between the teacher and student, unlike the more flexible response-based KD.

KD has been adapted for depth estimation~\cite{pilzer2019refine, wang2021knowledge, song2022learning, hu2023boosting}, effectively transferring knowledge from teacher to student models.
Despite their successes, these approaches are constrained by feature-based KD limitations, requiring knowledge of the teacher's architecture and meticulous feature map matching between teacher and student.

The fundamental question driving our research asks if it is possible to harness the advantages of feature-based KD using only the teacher's response (Sec.~\ref{sec:motivation}).
In response, we propose a novel knowledge distillation framework for monocular depth estimation called Teacher-Independent Explainable KD (TIE-KD).
Our method affords freedom from architectural constraints between teacher and student models by introducing an explainable feature map, the Depth Probability Map (DPM), generated directly from the teacher's depth map output (Sec.~\ref{sec:probability_map}).
Furthermore, we elaborate on a teacher-independent KD process that capitalizes on the DPM, incorporating two specially designed loss functions to ensure efficient knowledge transfer (Sec.~\ref{sec:KDproces}).

We validate our TIE-KD framework using three architecturally diverse teacher models, underscoring its robustness and adaptability (Sec.~\ref{sec:experiments}).
Utilizing the KITTI dataset, we demonstrate that TIE-KD consistently outperforms traditional response-based KD methods (Sec.~\ref{subsec:comparisonToRes}).
Moreover, TIE-KD shows remarkable flexibility in accommodating various backbone architectures within student models (Sec.~\ref{subsubsec:abal_backbone}).
An examination of the similarity between teacher and student outputs (Sec.~\ref{subsec:similarity}) reveals a closer alignment for TIE-KD-trained pairs, confirming the method's effectiveness in distilling knowledge (Fig.~\ref{fig:teaser}).
These findings collectively affirm the efficacy of our TIE-KD approach in monocular depth estimation.

In summary, our main contribution are the following:
\begin{itemize}
    \item Introduction of an innovative KD framework that operates independently of the teacher model's architecture.
    \item Utilization of explainable Depth Probability Map (DPM) derived from the teacher's output, enhancing the interpretability and efficiency of KD.
    \item Superior performance over traditional response-based KD methods, as confirmed by rigorous testing on the KITTI dataset.
    \item Proven adaptability and effectiveness across varying student model backbones, demonstrating the framework's versatility.
\end{itemize}

\section{Related work}
\label{sec:related_work}




{\bf Knowledge Distillation} enables lightweight models to enhance their performance by emulating more complex models.
It was first demonstrated by Ba and Caruana~\cite{ba2014deep} and formally defined by Hinton et al.~\cite{hinton2015distilling}, where the latter introduced a temperature-scaled softmax to transfer soft target probabilities via cross-entropy loss.
Beyond logits, feature map matching was pioneered by Romero et al.~\cite{romero2014fitnets}, while Zagoruyko and Komodakis~\cite{zagoruyko2016paying} introduced attention maps for distillation, and Heo et al.~\cite{heo2019knowledge} focused on the activation boundaries within neuron activations.
Recently, Zaho et al.~\cite{zhao2022decoupled} presented an innovative method for decoupling logits in classification tasks, further diversifying the applicability of knowledge distillation.
This technique has been extended to various domains, including object detection~\cite{li2017mimicking}, visual odometry~\cite{saputra2019distilling}, and so on.

{\bf Monocular Depth Estimation} (MDE) is the process of predicting the depth for each pixel in an image.
It is naturally a per-pixel regression problem, regression-based models have been extensively explored~\cite{eigen2014depth, godard2017unsupervised,lee2019big,ranftl2021vision, zhao2021transformer, li2023depthformer}.
Despite their effectiveness, these approaches can suffer from slow convergence and sub-optimal solutions~\cite{fu2018deep,li2022binsformer}.
Addressing these limitations, Fu et al.~\cite{fu2018deep} introduced DORN, a per-pixel classification-based MDE framework, assigning depth ranges to classes to improve efficiency.
Dias and Marathe~\cite{diaz2019soft} further refined this approach by introducing soft targets during training, enhancing the model's ability to generalize.
This classification concept has been widely adopted in subsequent research~\cite{liebel2019multidepth,johnston2020self,phan2021ordinal}, but it often results in decreased visual quality due to quantization effects.
The AdaBins model~\cite{bhat2021adabins} presents a solution to this challenge by utilizing adaptive bins for depth intervals, leading to more accurate depth predictions through a weighted sum of bin centers and bin's probabilities.
Li et al.~\cite{li2022binsformer} built upon this hybrid classification-regression approach by integrating a Transformer decoder for bin generation, pushing the boundaries of classification-regression MDE.

Our proposed KD framework aligns with this classification-regression paradigm while maintaining the versatility to work with teacher models outside this framework.

{\bf Knowledge Distillation for Depth Estimation} has been addressed in various studies~\cite{wang2021knowledge, song2022learning, pilzer2019refine, hu2023boosting}.
Pilzer et al.~\cite{pilzer2019refine} pioneered knowledge distillation in unsupervised monocular depth estimation, employing a self-distillation structure within a single model.
Wang et al.~\cite{wang2021knowledge} evolved from pixel-wise to pair-wise distillation, inspired by segmentation techniques~\cite{liu2019structured}, enabling student models to emulate teacher feature maps more effectively.
Song et al.~\cite{song2022learning} leveraged stereo-based teacher models and introduced selective distillation for multi-scale feature maps from student encoders.
Hu et al.~\cite{hu2023boosting} tackled the challenge of capacity disparities between student and teacher models by integrating auxiliary unlabeled data into the distillation process, a departure from traditional methods that focus solely on depth map features.

Our KD approach distinguishes itself from the aforementioned methods by generating an interpretable feature map directly from the teacher model’s depth map output.
This not only simplifies the distillation process but also provides an explicit representation of the distilled knowledge, paving the way for a more transparent and potentially more generalizable learning paradigm for depth estimation.

\section{Methodology}\label{sec:method}

\subsection{Motivation}
\label{sec:motivation}

Knowledge distillation techniques fall into two primary categories: response-based and feature-based KD.

In response-based KD, the student model is trained to mimic the teacher model's output (i.e., response).
This method permits varying architectures between the teacher and student models, as the alignment is exclusively based on the teacher's output.
The output could be a scalar value like pixel depth (hard label) or probability logits in classification tasks (soft label).
When hard labels approximate the ground truth (GT), the process resembles conventional training using the GT.
Soft labels, however, allow for knowledge transfer and induce regularization~\cite{hinton2015distilling}.

In feature-based KD, the emphasis shifts to replicating the teacher's feature maps, veering away from the response-based approach.
This method inherently provides a regularization effect and often boosts the student model's performance~\cite{romero2014fitnets, zagoruyko2016paying, heo2019knowledge}.
However, it usually benefits from similar architectures between the teacher and student models and faces challenges in aligning layers or feature maps effectively~\cite{gou2021knowledge}.
This is because it is difficult to define what specific knowledge a given feature map contains.

Depth estimation models output hard labels, in the form of depth maps, complicating the use of response-based KD in this domain.
While feature-based KD can be applied, it is still encumbered by the limitations inherent to feature-based KD.
Motivated by these challenges, we formulate the central question of our study: \textit{`Can the advantages of feature-based KD be replicated using only the teacher's response?'}
To address this, we propose a novel method to generate feature maps from the teacher's response, circumventing the need for extracting feature maps from the teacher model.
We also propose a knowledge distillation process that utilizes these generated feature maps.



\subsection{Explainable Depth Probability Map}
\label{sec:probability_map}

The conversion of depth regression tasks into classification problems was first pioneered by Fu et al.~\cite{fu2018deep}.
This methodology was later refined by AdaBins~\cite{bhat2021adabins}, which introduced adaptive binning.
Such classification-based strategies yield logits for each depth bin, thereby generating soft labels that are particularly beneficial for knowledge distillation.
Consistent with these developments, our work leverages this classification-based approach to maximize the advantages of using soft labels in the context of knowledge distillation.

Previous classification-regression-based techniques~\cite{bhat2021adabins,li2022binsformer} decode the final depth value by a specific decoding function like a weighted sum of logits along with the center values of the corresponding bins.
However, we observe that the logit value for a specific bin does not unambiguously indicate the likelihood of that bin representing the true depth value.
Various combinations of weights and center values can yield identical results, leading to a misalignment with the common expectation that the bin with the highest logit should naturally correspond to the actual depth.
This method of representation is both model-dependent and lacks intuitive interpretability, making it less suitable for response-based KD methods.

To address these issues, we introduce an easily interpretable feature map known as the \textit{depth probability map}.
This map is filled with probabilities that directly quantify the likelihood of each bin's contribution to the final depth value, offering a more transparent and model-independent representation.

Yuan et al.~\cite{yuan2020revisiting} proposed a label-smoothing technique that distributes probabilities uniformly across all classes while maintaining a higher probability for the class representing the GT.
Although effective in general classification problems, this method loses its efficacy in depth estimation tasks.
For depth estimation, the probability values in adjacent bins are not isolated but have a strong relationship owing to the continuous nature of depth.
We address this shortcoming by allocating higher probabilities to bins that are more proximate to the GT bin, thereby taking advantage of the inherent continuity in depth values.

To realize this idea, we uniformly divide the depth range into \(k\) bins and apply a Gaussian-like function, expressed in Eq.~\ref{eq:distribution}, to softly distribute probabilities around the GT depth value.
\begin{equation}
    f(x, GT) = \frac{1}{\sigma \sqrt{2\pi}} e^{-\frac{1}{2} \left(\frac{x - GT}{\sigma}\right)^2}
    \label{eq:distribution}
\end{equation}
The probability assigned to the \(i\)-th bin \(B_i\) is computed using integration, as presented in Eq.~\ref{eq:probability}:
\begin{equation}
    p(B_i) = \int_{B_i} f(x, GT) dx
    \label{eq:probability}
\end{equation}
To mitigate the dilution of probabilities in the core region due to extremely low values, we employ a cut-off threshold, such as \(10^{-16}\).
The probabilities of the remaining bins are then normalized using a softmax function.

Lastly, we apply this formulation to each pixel in the teacher's response, generating a depth probability map.
The resulting feature map's dimension is \([H, W, B]\), where \(H\) and \(W\) refer to the height and width of the input image, and \(B\) denotes the number of bins.

\subsection{Teacher-Independent Knowledge Distillation}\label{sec:KDproces}

\begin{figure*}[t] 
    \centering
    \includegraphics[width=\textwidth]{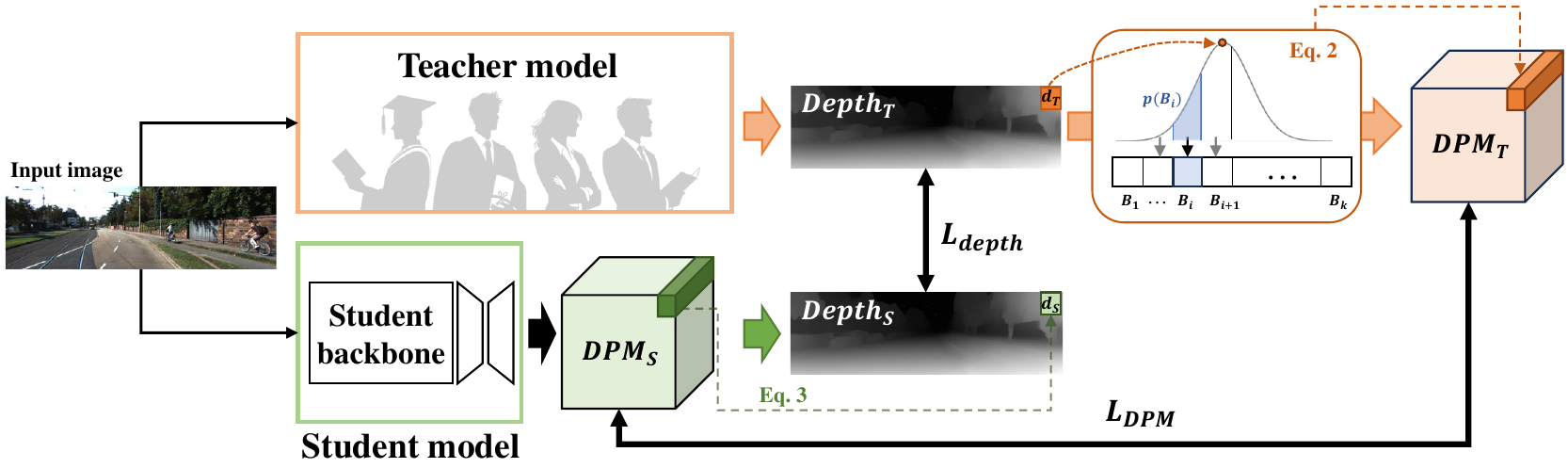}
    \caption{Overview of our teacher-independent and explainable KD process for single image depth estimation
    }
    \label{fig:overview}
\end{figure*}

Leveraging our explainable depth probability map, we propose a novel knowledge distillation method that is agnostic to the teacher model architecture while also benefiting from feature-based KD techniques.
The overview of our approach is illustrated in Fig.~\ref{fig:overview}.

{\bf Student network}
The student model features a flexible architecture, combining a backbone network with an encoder-decoder configuration. 
Its final output is a Depth Probability Map (DPM), tailored to capture depth information efficiently.
The backbone is adaptable, allowing for various architectures and sizes, and the DPM is dimensioned as \([H, W, B]\).
The depth estimation for each pixel is computed from the DPM, where the final depth value is derived through a weighted sum of the probabilities and the corresponding bin center depths.
This process is mathematically represented as:
\begin{equation}
    d = \sum_{i} center(B_i) \times p(B_i)
\label{eq:depthDecoding}
\end{equation}

\subsubsection{\bf KD process}
Both the teacher and student networks receive the same input image and produce depth maps as outputs.
The teacher network's output is a depth map, denoted as $Depth_T$, which is then converted into a depth probability map ($DPM_T$) using Eq.~\ref{eq:probability}.
On the other hand, the student network's last layer outputs a depth probability map ($DPM_S$), which is subsequently decoded into a depth map ($Depth_S$).
Thus, we obtain depth probability maps and depth maps from both models.

To optimize the student model, we employ two distinct loss functions: $L_{DPM}$ for the depth probability map and $L_{depth}$ for the depth map.
The overall loss, $L$, is a composite of these two functions, weighted and scaled as expressed in Eq.~\ref{eq:loss}. Here, $\alpha$ represents the weight assigned to $L_{DPM}$, and $\beta$ serves as a scaling factor to adjust the overall magnitude of the loss.
\begin{equation}
    L = \beta(\alpha L_{DPM} + (1-\alpha) L_{depth})
\label{eq:loss}
\end{equation}

\subsubsection{\bf Depth Probability Map Loss}
The $L_{DPM}$ term measures the divergence between $DPM_T$ and $DPM_S$ using Kullback-Leibler (KL) divergence at the pixel level.
The overall loss is an average of these pixel-wise divergences, as delineated in Eq.~\ref{eq:DPMloss}.
\begin{equation}
    L_{DPM} = \frac{1}{M} \sum_{m \in M} P_S(m) \log \left(\frac{P_S(m)}{P_T(m)}\right)
    \label{eq:DPMloss}
\end{equation}
Here, $M$ represents the set of all pixels, and $P_S(m)$ and $P_T(m)$ are the probabilities from the student's and teacher's depth probability maps for pixel $m$, respectively.

\subsubsection{\bf Depth Map Loss}
The $L_{depth}$ term quantifies the dissimilarity between depth maps $Depth_T$ and $Depth_S$, utilizing the Structural Similarity Index (SSIM) for this comparison.
The loss is calculated as specified in Eq.~\ref{eq:depthLoss}.
\begin{equation}
    L_{depth} = 1 - \text{SSIM}(Depth_T, Depth_S)
    \label{eq:depthLoss}
\end{equation}
This loss serves to minimize the discrepancy between the teacher's and the student's depth maps, thereby facilitating effective learning for the student model.

\section{Experiments}\label{sec:experiments}


Our comprehensive experiments, conducted using three unique teacher models, validate the efficacy of the TIE-KD framework.
In the following sections, we describe the datasets, performance metrics, and experimental details (Sec.~\ref{subsec:dataset} and Sec.~\ref{subsec:impl}).
We then compare TIE-KD's effectiveness against response-based KD methods, highlighting its robustness across different architectures (Sec.~\ref{subsec:comparisonToRes} and Sec.~\ref{subsec:similarity}).
Ablation studies on loss functions, student backbones, and hyper-parameters (Sec.~\ref{subsec:abalation}) further illustrate TIE-KD's flexibility and robustness.

\subsection{Dataset and evaluation metrics}\label{subsec:dataset}

\paragraph{\bf Dataset}
For our experiments, we utilized the well-established KITTI dataset, which provides stereo images and corresponding 3D LiDAR point clouds of street scenes~\cite{geiger2013vision}.
The RGB images are of high resolution ($1241 \times 376$ pixels), and the depth maps capture distances up to 80 meters.
Following the protocol established by Eigen et al.~\cite{eigen2014depth}, we trained our models on a set of approximately 26K left camera images and validated on a separate test set comprising 697 images.
During training, we augmented the data by randomly cropping the images to a resolution of $704 \times 352$ pixels.



\paragraph{\bf Evaluation metrics}
Our assessment adheres to the established evaluation protocol for depth estimation, following the precedent set by prior work~\cite{eigen2014depth, pilzer2019refine, song2022learning}.
Specifically, we measure performance using the Absolute Relative Error (AbsRel), Squared Relative Error (SqRel), Root Mean Squared Error (RMSE), and Root Mean Squared Logarithmic Error (RMSE$_{\log}$).
Additionally, we evaluate the accuracy under threshold, defined by the metric $\delta_i < 1.25^i$ for $i = 1, 2, 3$, to capture the proportion of depth estimates within specified error bounds.




\subsection{Implementation Details}\label{subsec:impl}
Our TIE-KD framework and five distinct response-based KD methodologies were developed using PyTorch.
For all methods, including our own and comparative approaches, we utilized the Adam optimizer with $\beta_1 = 0.95$ and $\beta_2 = 0.99$, implementing a one-cycle policy with a peak learning rate of 1e-3.
For  TIE-KD, we assign values of 0.1 and 10 to the weight parameters $\alpha$ and $\beta$, respectively, in Eq.~\ref{eq:loss}.
Additionally, the $\sigma$ in Eq.~\ref{eq:distribution} is set to 0.8.
Models were trained for 24 epochs with a batch size of 32, selecting the weights from the epoch that yielded the best performance.

\begin{itemize}
    \item \textbf{Student Network:} Our student model utilizes a UNet-inspired encoder-decoder structure~\cite{ronneberger2015u}, with MobileNetV2~\cite{Sandler_2018_CVPR} serving as the backbone.
    The final layer outputs a DPM with dimensions of $[704, 352, 257]$, thereby segmenting the depth range from 0 to 80 meters into 257 distinct bins.
    The total model size depends on the chosen backbone, adjusted accordingly within the encoder and decoder segments. Overall, the student network comprises roughly 17.6 million parameters.

    \item \textbf{Teacher models:}
    For our teacher models, we utilized three well-known monocular depth estimation architectures: AdaBins~\cite{bhat2021adabins}, BTS~\cite{lee2019big}, and DepthFormer~\cite{li2023depthformer}, with parameter counts of 78 million, 48 million, and 273 million, respectively.
    We utilized pre-trained models obtained from a public repository\footnote{https://github.com/zhyever/Monocular-Depth-Estimation-Toolbox}.

    \item \textbf{Baseline:}
    The baseline model shares the same architecture as the student network but replaces the DPM layer with a regression layer, resulting in a similar parameter count of around 17.6 million.
    The baseline was trained from scratch on the KITTI dataset using the scale invariant loss (SI) introduced by Eigen et al.~\cite{eigen2014depth}.

    \item \textbf{Response-based KD (Res-KD):}
    We implemented general response-based KD methods suitable for use with any teacher model architecture, in contrast to feature-based approaches.
    These methods entailed comparing depth maps generated by the teacher and student models, utilizing a range of loss functions including SSIM, MSE (Mean Square Error), and SI.
    Combinations of SSIM with SI, and SSIM with MSE, were also evaluated.
\end{itemize}

We noted that teacher models often show less consistent performance in the upper image regions, likely due to the sparsity of LiDAR data in the KITTI dataset, particularly in distant areas like the sky or mountains.
To address this and improve the reliability of our KD process, we excluded the top 110 pixels from the height of the images for loss computation.
This strategy focuses our training on regions with denser and more reliable depth data, ensuring a more consistent and dependable dataset.

\subsection{Comparison with Response-Based KD Methods}\label{subsec:comparisonToRes}

\begin{table*}[t]
\caption{Comparative evaluation of depth estimation performance across various models.
Bold values indicate the best performance among student models, and underlined values denote the second-best methods.}
\label{table:results}
\resizebox{\textwidth}{!}{%
\begin{tabular}{|ccc|cccc|ccc|}
\hline
\multicolumn{1}{|c|}{\multirow{2}{*}{\begin{tabular}[c]{@{}c@{}}Teacher\\      model\end{tabular}}} &
  \multicolumn{1}{c|}{\multirow{2}{*}{\begin{tabular}[c]{@{}c@{}}Method\\     (\# of parameters)\end{tabular}}} &
  \multirow{2}{*}{\begin{tabular}[c]{@{}c@{}}Loss\\      function(s)\end{tabular}} &
  \multicolumn{4}{c|}{Lower   is better ($\downarrow$)} &
  \multicolumn{3}{c|}{Higher   is better ($\uparrow$)} \\ \cline{4-10} 
\multicolumn{1}{|c|}{} &
  \multicolumn{1}{c|}{} &
   &
  \multicolumn{1}{c|}{AbsRel} &
  \multicolumn{1}{c|}{SqRel} &
  \multicolumn{1}{c|}{RMSE} &
  RMSE$_{log}$ &
  \multicolumn{1}{c|}{$\delta_1$} &
  \multicolumn{1}{c|}{$\delta_2$} &
  $\delta_3$ \\
  \hline \hline 
\multicolumn{3}{|c|}{Baseline (17.6 M)} &
  \multicolumn{1}{c|}{0.0663} &
  \multicolumn{1}{c|}{0.2340} &
  \multicolumn{1}{c|}{2.5625} &
  0.1017 & 
  \multicolumn{1}{c|}{0.9501} &
  \multicolumn{1}{c|}{0.9926} &
  0.9984 \\
  \hline \hline 
\multicolumn{1}{|c|}{\multirow{8}{*}{AdaBins~\cite{bhat2021adabins}}} &
  \multicolumn{2}{c|}{Teacher (78 M)} &
  \multicolumn{1}{c|}{0.0593} &
  \multicolumn{1}{c|}{0.1941} &
  \multicolumn{1}{c|}{2.3309} &
  0.0901 &
  \multicolumn{1}{c|}{0.9631} &
  \multicolumn{1}{c|}{0.9946} &
  0.9990 \\ \cline{2-10} 
\multicolumn{1}{|c|}{} &
  \multicolumn{1}{c|}{\multirow{5}{*}{Res-KD}} &
  SSIM &
  \multicolumn{1}{c|}{\underline{0.0697}} &
  \multicolumn{1}{c|}{\underline{0.2407}} &
  \multicolumn{1}{c|}{\underline{2.5639}} &
  {\underline{0.1041}} &
  \multicolumn{1}{c|}{0.9457} &
  \multicolumn{1}{c|}{\underline{0.9933}} &
  {\bf 0.9985} \\ 
\multicolumn{1}{|c|}{} &
  \multicolumn{1}{c|}{} &
  MSE &
  \multicolumn{1}{c|}{0.0786} &
  \multicolumn{1}{c|}{0.2793} &
  \multicolumn{1}{c|}{2.6964} &
  0.1155 &
  \multicolumn{1}{c|}{0.9319} &
  \multicolumn{1}{c|}{0.9911} &
  0.9981 \\ 
\multicolumn{1}{|c|}{} &
  \multicolumn{1}{c|}{} &
  SI &
  \multicolumn{1}{c|}{0.0739} &
  \multicolumn{1}{c|}{0.2747} &
  \multicolumn{1}{c|}{2.7371} &
  0.1112 &
  \multicolumn{1}{c|}{0.9382} &
  \multicolumn{1}{c|}{0.9916} &
  0.9980 \\ 
\multicolumn{1}{|c|}{} &
  \multicolumn{1}{c|}{} &
  SSIM,SI &
  \multicolumn{1}{c|}{0.0701} &
  \multicolumn{1}{c|}{0.2445} &
  \multicolumn{1}{c|}{2.5833} &
  0.1047 &
  \multicolumn{1}{c|}{\underline{0.9458}} &
  \multicolumn{1}{c|}{0.9932} &
  \bf 0.9985 \\ 
\multicolumn{1}{|c|}{} &
  \multicolumn{1}{c|}{} &
  SSIM,MSE &
  \multicolumn{1}{c|}{0.0808} &
  \multicolumn{1}{c|}{0.2807} &
  \multicolumn{1}{c|}{2.6943} &
   0.1158 &
  \multicolumn{1}{c|}{0.9328} &
  \multicolumn{1}{c|}{0.9906} &
   0.9983 \\ \cline{2-10}
\multicolumn{1}{|c|}{} &
  \multicolumn{1}{c|}{TIE-KD} &
  $L_{DPM}$, $L_{depth}$ &
  \multicolumn{1}{c|}{\bf 0.0654} &
  \multicolumn{1}{c|}{\bf 0.2179} &
  \multicolumn{1}{c|}{\bf 2.4315} &
   {\bf 0.0980} &
  \multicolumn{1}{c|}{\bf 0.9540} &
  \multicolumn{1}{c|}{\bf 0.9939} &
   {\bf 0.9985} \\
   \hline \hline 
\multicolumn{1}{|c|}{\multirow{8}{*}{BTS~\cite{lee2019big}}} &
  \multicolumn{2}{c|}{Teacher (47M)} &
  \multicolumn{1}{c|}{0.0586} &
  \multicolumn{1}{c|}{0.2060} &
  \multicolumn{1}{c|}{2.4798} &
  0.0916 &
  \multicolumn{1}{c|}{0.9602} &
  \multicolumn{1}{c|}{0.9940} &
  0.9986 \\ \cline{2-10} 
\multicolumn{1}{|c|}{} &
  \multicolumn{1}{c|}{\multirow{5}{*}{Res-KD}} &
  SSIM &
  \multicolumn{1}{c|}{0.0697} &
  \multicolumn{1}{c|}{\underline{0.2460}} &
  \multicolumn{1}{c|}{2.6357} &
  0.1050 &
  \multicolumn{1}{c|}{0.9434} &
  \multicolumn{1}{c|}{\underline{0.9930}} &
  {\bf 0.9985} \\ 
\multicolumn{1}{|c|}{} &
  \multicolumn{1}{c|}{} &
  MSE &
  \multicolumn{1}{c|}{0.0820} &
  \multicolumn{1}{c|}{0.2977} &
  \multicolumn{1}{c|}{2.7440} &
  0.1203 &
  \multicolumn{1}{c|}{0.9263} &
  \multicolumn{1}{c|}{0.9895} &
  0.9977 \\ 
\multicolumn{1}{|c|}{} &
  \multicolumn{1}{c|}{} &
  SI &
  \multicolumn{1}{c|}{0.0782} &
  \multicolumn{1}{c|}{0.2931} &
  \multicolumn{1}{c|}{2.8106} &
  0.1157 &
  \multicolumn{1}{c|}{0.9346} &
  \multicolumn{1}{c|}{0.9903} &
  0.9977 \\ 
\multicolumn{1}{|c|}{} &
  \multicolumn{1}{c|}{} &
  SSIM,SI &
  \multicolumn{1}{c|}{\underline{0.0690}} &
  \multicolumn{1}{c|}{0.2462} &
  \multicolumn{1}{c|}{\underline{2.6168}} &
  {\underline{0.1044}} &
  \multicolumn{1}{c|}{\underline{0.9467}} &
  \multicolumn{1}{c|}{0.9928} &
  \underline{0.9983} \\ 
\multicolumn{1}{|c|}{} &
  \multicolumn{1}{c|}{} &
  SSIM,MSE &
  \multicolumn{1}{c|}{0.0914} &
  \multicolumn{1}{c|}{0.3130} &
  \multicolumn{1}{c|}{2.7983} &
   0.1241 &
  \multicolumn{1}{c|}{0.9222} &
  \multicolumn{1}{c|}{0.9888} &
   0.9981 \\ \cline{2-10}
\multicolumn{1}{|c|}{} &
  \multicolumn{1}{c|}{TIE-KD} &
  $L_{DPM}$, $L_{depth}$ &
  \multicolumn{1}{c|}{\bf 0.0656} &
  \multicolumn{1}{c|}{\bf 0.2247} &
  \multicolumn{1}{c|}{\bf 2.4984} &
   {\bf 0.0995} &
  \multicolumn{1}{c|}{\bf 0.9523} &
  \multicolumn{1}{c|}{\bf 0.9933} &
   {\bf{0.9985}} \\
   \hline \hline 
\multicolumn{1}{|c|}{\multirow{8}{*}{DepthFormer~\cite{li2023depthformer}}} &
  \multicolumn{2}{c|}{Teacher (273 M)} &
  \multicolumn{1}{c|}{0.0513} &
  \multicolumn{1}{c|}{0.1511} &
  \multicolumn{1}{c|}{2.1038} &
  0.0783 &
  \multicolumn{1}{c|}{0.9752} &
  \multicolumn{1}{c|}{0.9970} &
  0.9993 \\ \cline{2-10} 
\multicolumn{1}{|c|}{} &
  \multicolumn{1}{c|}{\multirow{5}{*}{Res-KD}} &
  SSIM &
  \multicolumn{1}{c|}{0.0692} &
  \multicolumn{1}{c|}{\underline{0.2337}} &
  \multicolumn{1}{c|}{\underline{2.5009}} &
  0.1019 &
  \multicolumn{1}{c|}{\underline{0.9493}} &
  \multicolumn{1}{c|}{\underline{0.9937}} &
  {\textbf{0.9987}} \\ 
\multicolumn{1}{|c|}{} &
  \multicolumn{1}{c|}{} &
  MSE &
  \multicolumn{1}{c|}{0.0805} &
  \multicolumn{1}{c|}{0.2692} &
  \multicolumn{1}{c|}{2.6029} &
  0.1134 &
  \multicolumn{1}{c|}{0.9361} &
  \multicolumn{1}{c|}{0.9915} &
  0.9983 \\ 
\multicolumn{1}{|c|}{} &
  \multicolumn{1}{c|}{} &
  SI &
  \multicolumn{1}{c|}{0.0724} &
  \multicolumn{1}{c|}{0.2648} &
  \multicolumn{1}{c|}{2.6717} &
  0.1078 &
  \multicolumn{1}{c|}{0.9411} &
  \multicolumn{1}{c|}{0.9921} &
  0.9984 \\ 
\multicolumn{1}{|c|}{} &
  \multicolumn{1}{c|}{} &
  SSIM,SI &
  \multicolumn{1}{c|}{\underline{0.0682}} &
  \multicolumn{1}{c|}{0.2382} &
  \multicolumn{1}{c|}{2.5709} &
  {\underline{0.1018}} &
  \multicolumn{1}{c|}{0.9488} &
  \multicolumn{1}{c|}{\underline{0.9937}} &
  {\textbf{0.9987}} \\ 
\multicolumn{1}{|c|}{} &
  \multicolumn{1}{c|}{} &
  SSIM,MSE &
  \multicolumn{1}{c|}{0.0770} &
  \multicolumn{1}{c|}{0.2720} &
  \multicolumn{1}{c|}{2.6391} &
   0.1133 &
  \multicolumn{1}{c|}{0.9363} &
  \multicolumn{1}{c|}{0.9912} &
   0.9979 \\ \cline{2-10} 
\multicolumn{1}{|c|}{} &
  \multicolumn{1}{c|}{TIE-KD} &
  $L_{DPM}$, $L_{depth}$ &
  \multicolumn{1}{c|}{\bf 0.0657} &
  \multicolumn{1}{c|}{\bf 0.2208} &
  \multicolumn{1}{c|}{\bf 2.4402} &
   {\bf 0.0980} &
  \multicolumn{1}{c|}{\bf 0.9534} &
  \multicolumn{1}{c|}{\bf 0.9940} &
   \underline{0.9986}\\ \hline
\end{tabular}%
}
\end{table*}

Table~\ref{table:results} presents the performance of baseline models, a range of teacher models, and student models that have been trained using different KD methods.
Students trained under our TIE-KD framework consistently surpassed the performance of both the baseline and other response-based KD methods across all metrics, independent of the teacher model's architecture.
This achievement is noteworthy considering that TIE-KD students were not trained with GT data from the KITTI dataset, in contrast to the baseline model which was trained directly on the target dataset.

Among the response-based KD methods, those utilizing SSIM loss, or a blend of SSIM and SI loss, were typically the most effective.
Nevertheless, these methods often did not outperform the baseline, even when the students were instructed by highly capable teachers.
Our TIE-KD approach, in comparison, consistently exhibited superior performance, demonstrating its robust capacity to effectively harness the knowledge conveyed by teacher models.

Among the student models trained with three distinct teachers, the student instructed by AdaBins generally outperformed its counterparts.
It achieved up to a 7.4\% improvement (e.g., in SqRel) and an average enhancement of 4.5\% across the four performance metrics (AbsRel, SeRel, RMSE, and RMSE$_{log}$) when compared to the baseline.
Interestingly, the student trained under the guidance of DepthFormer, despite its larger model size and superior performance, did not exceed the performance of those trained by AdaBins.
This result aligns with the findings of Mirzadeh et al.~\cite{mirzadeh2020improved}, which suggest that a large disparity in parameters between teacher and student models can negatively impact performance.


\begin{figure*}[t]
    \centering
    \resizebox{0.94\textwidth}{!}{%
    \renewcommand{\arraystretch}{1}
    \setlength{\tabcolsep}{0pt}
    {\scriptsize
    \begin{tabular}{c@{\hspace{0.5em}}@{\hspace{0.5em}}r@{\hspace{0.5em}}c@{\hspace{0.5em}}c@{\hspace{0.5em}}c}
            \multicolumn{2}{c}{ \small Input\,}  & 
                \raisebox{-0.4\height}{\includegraphics[width=0.3\textwidth, height=0.05\textheight]{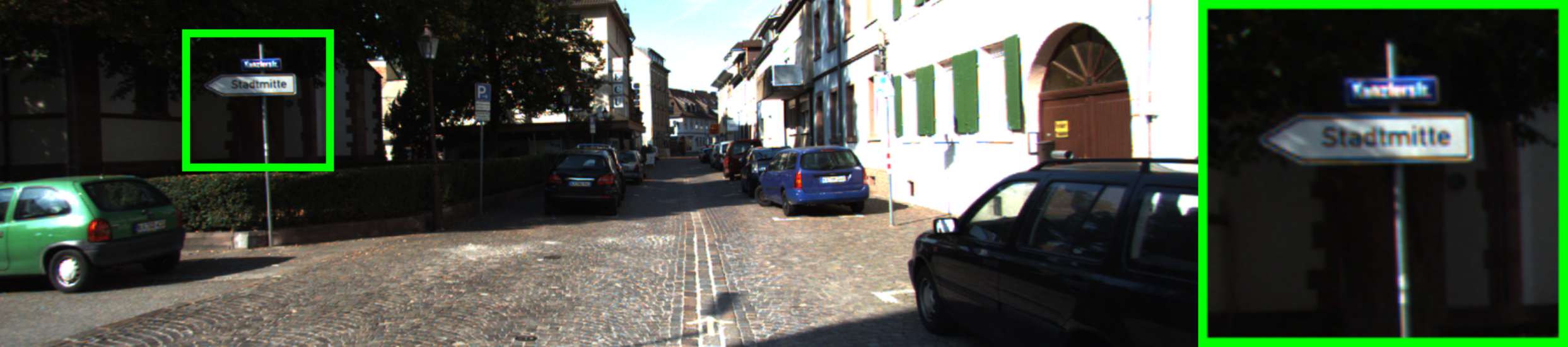}}& \vspace{0.5pt}
                \raisebox{-0.4\height}{\includegraphics[width=0.3\textwidth, height=0.05\textheight]{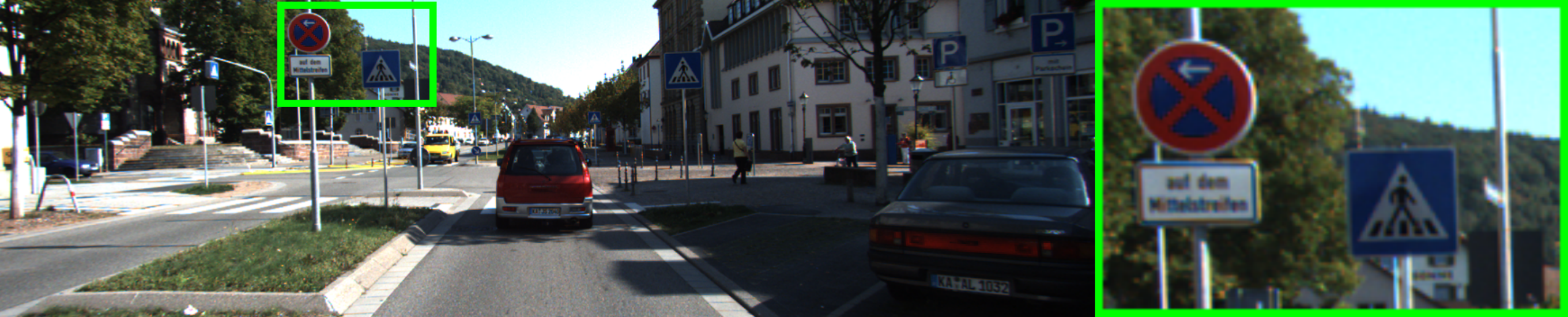}}& \vspace{0.5pt}
                \raisebox{-0.4\height}{\includegraphics[width=0.3\textwidth, height=0.05\textheight]{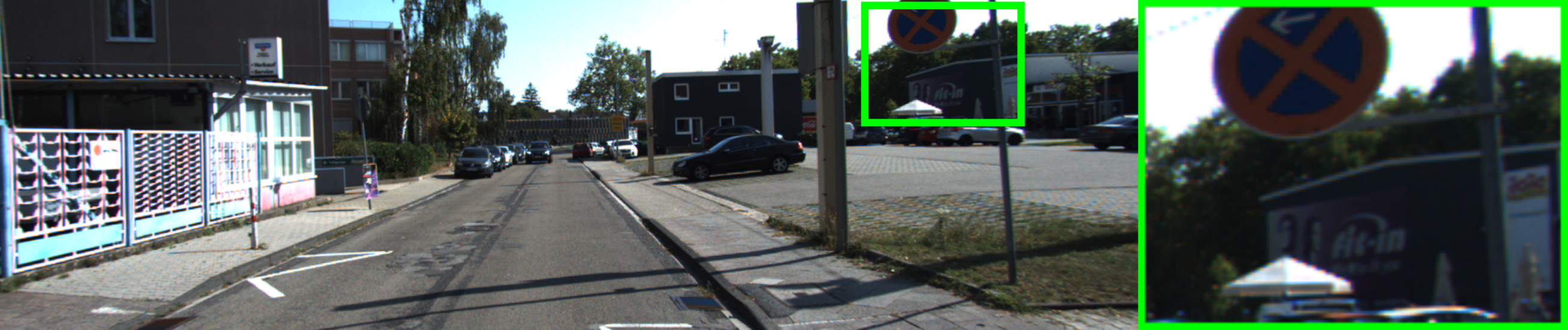}} \\ 
            \multicolumn{2}{c}{ \small Baseline\,}& 
                \raisebox{-0.4\height}{\includegraphics[width=0.3\textwidth, height=0.05\textheight]{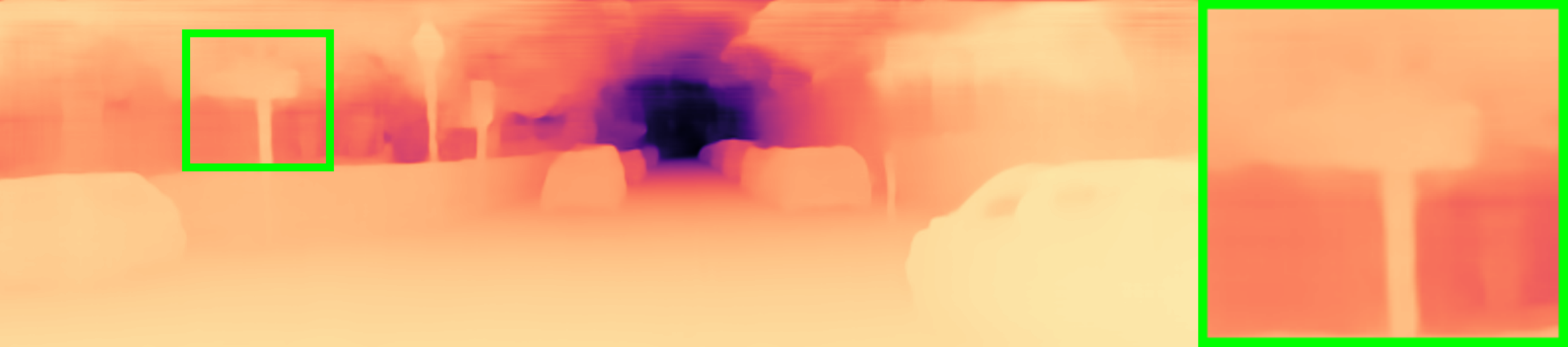}}& \vspace{0.5pt} 
                \raisebox{-0.4\height}{\includegraphics[width=0.3\textwidth, height=0.05\textheight]{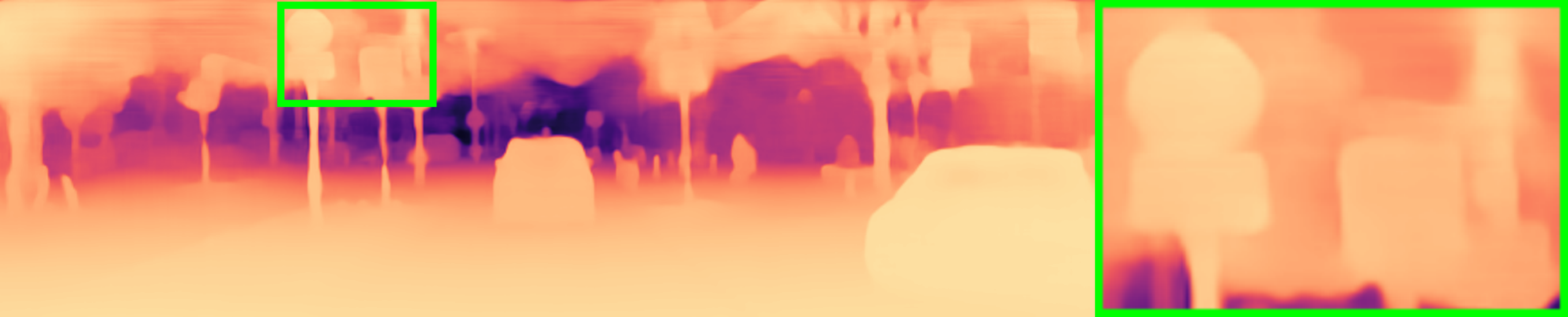}}& \vspace{0.5pt} 
                \raisebox{-0.4\height}{\includegraphics[width=0.3\textwidth, height=0.05\textheight]{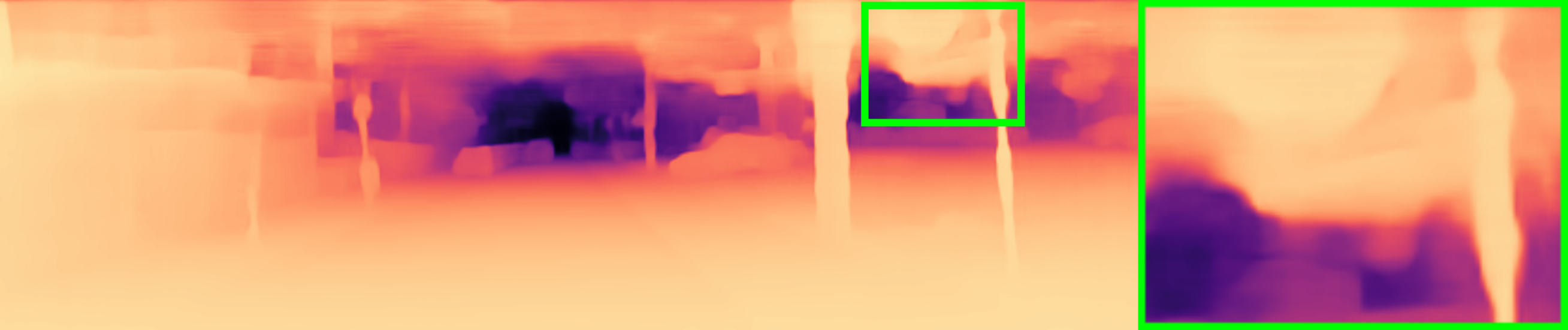}} \\ 
        \bottomrule[0.3pt]
        
        \toprule[0.3pt]
            \multirow{4}{*}{\rotatebox{90}{\parbox[c]{0.2\linewidth}{\centering \small AdaBins}}} 
            &  {\it Teacher}\,&  
                \raisebox{-0.4\height}{\includegraphics[width=0.3\textwidth, height=0.05\textheight]{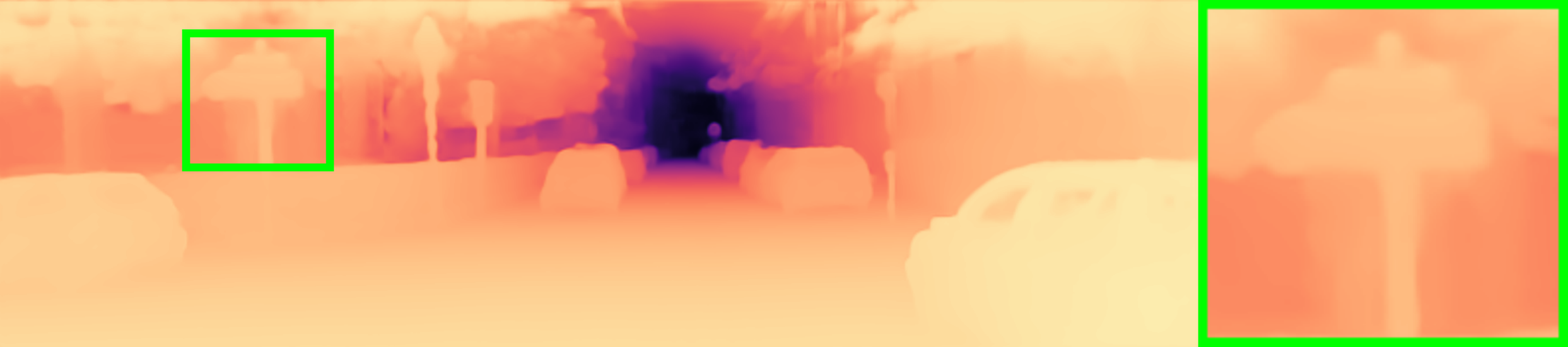}}& \vspace{0.5pt}
                \raisebox{-0.4\height}{\includegraphics[width=0.3\textwidth, height=0.05\textheight]{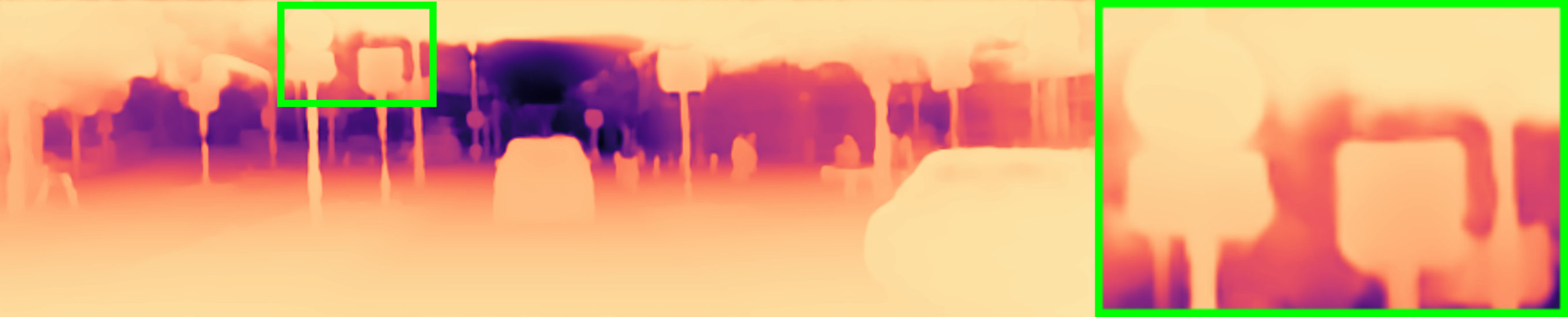}}& \vspace{0.5pt} 
                \raisebox{-0.4\height}{\includegraphics[width=0.3\textwidth, height=0.05\textheight]{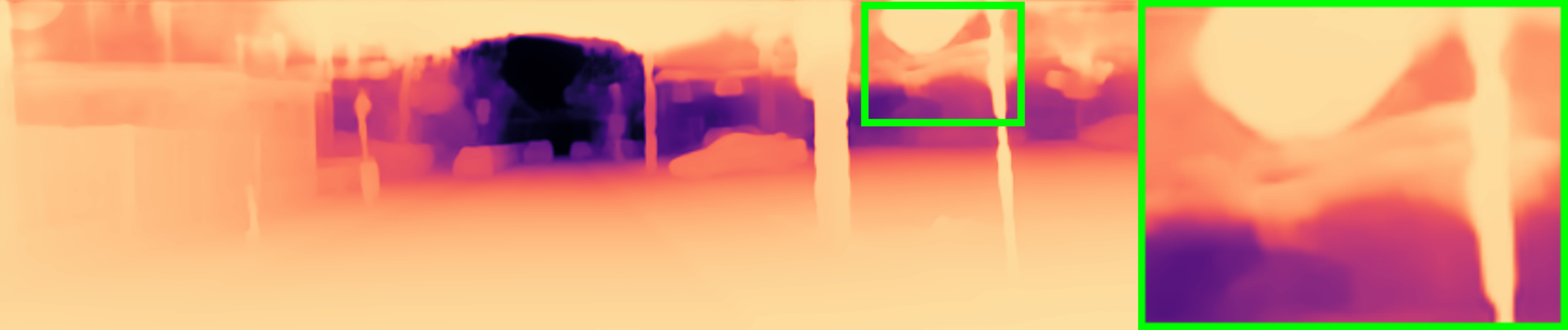}} \\ 
            &  \begin{tabular}{@{}c@{}}Res-KD \\ (SSIM)\end{tabular}\,&  
                \raisebox{-0.4\height}{\includegraphics[width=0.3\textwidth, height=0.05\textheight]{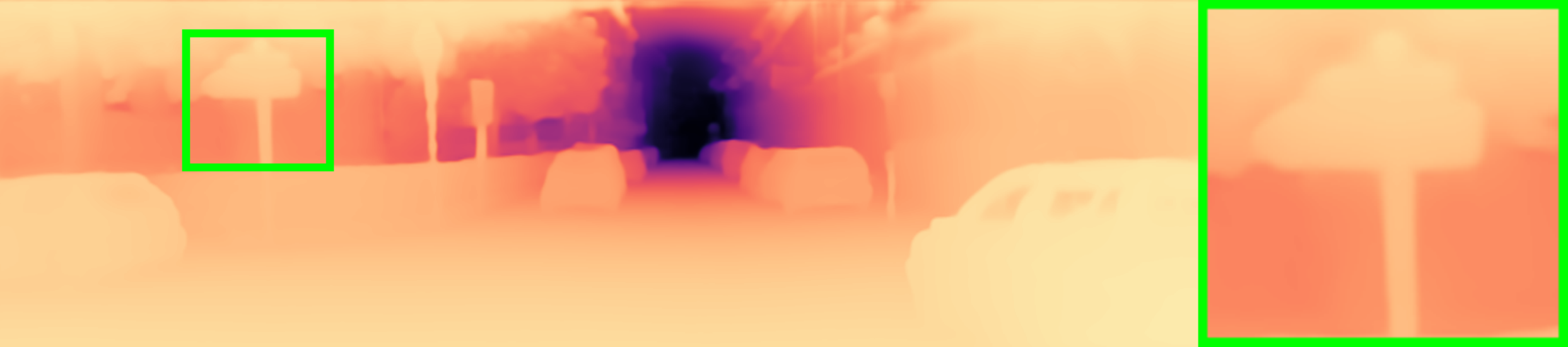}}& \vspace{0.5pt} 
                \raisebox{-0.4\height}{\includegraphics[width=0.3\textwidth, height=0.05\textheight]{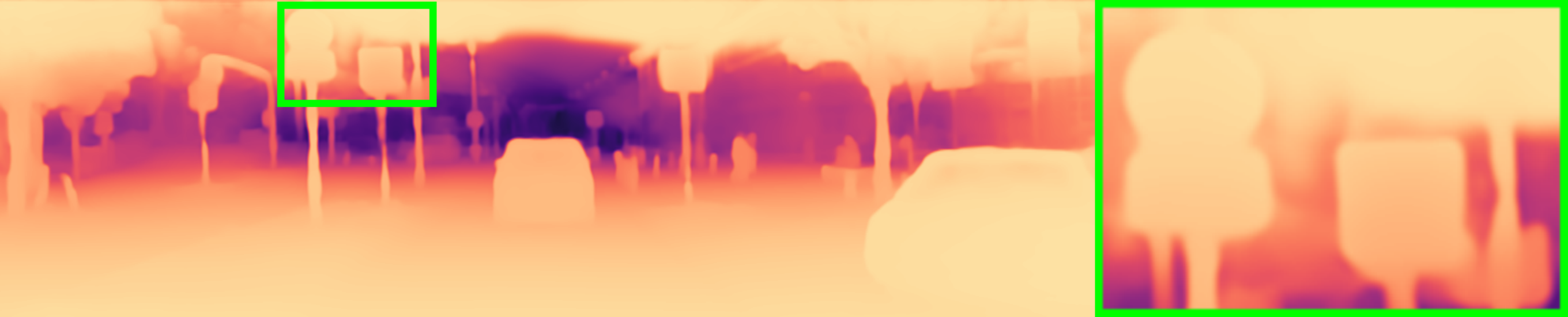}}& \vspace{0.5pt} 
                \raisebox{-0.4\height}{\includegraphics[width=0.3\textwidth, height=0.05\textheight]{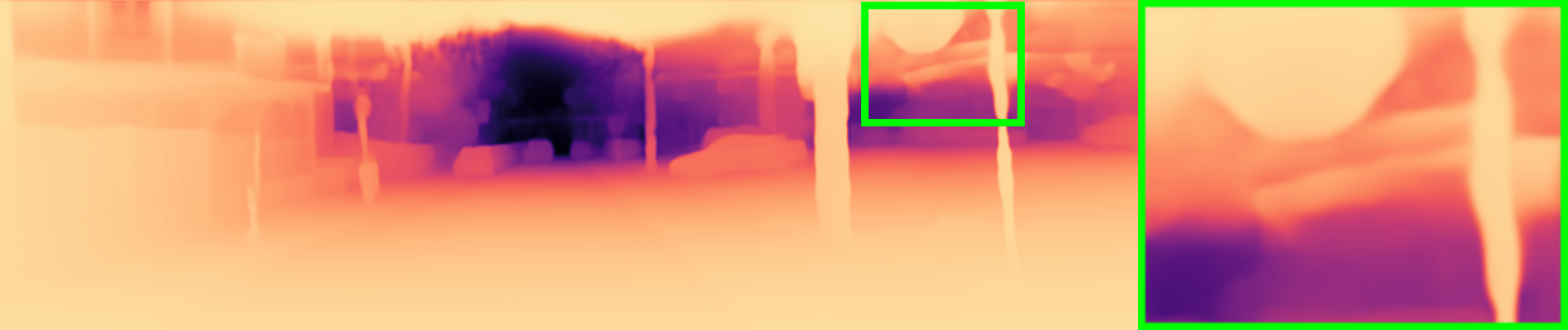}} \\ 
            &  \begin{tabular}{@{}c@{}}Res-KD \\ (SSIM, SI)\end{tabular}\,&  
                \raisebox{-0.4\height}{\includegraphics[width=0.3\textwidth, height=0.05\textheight]{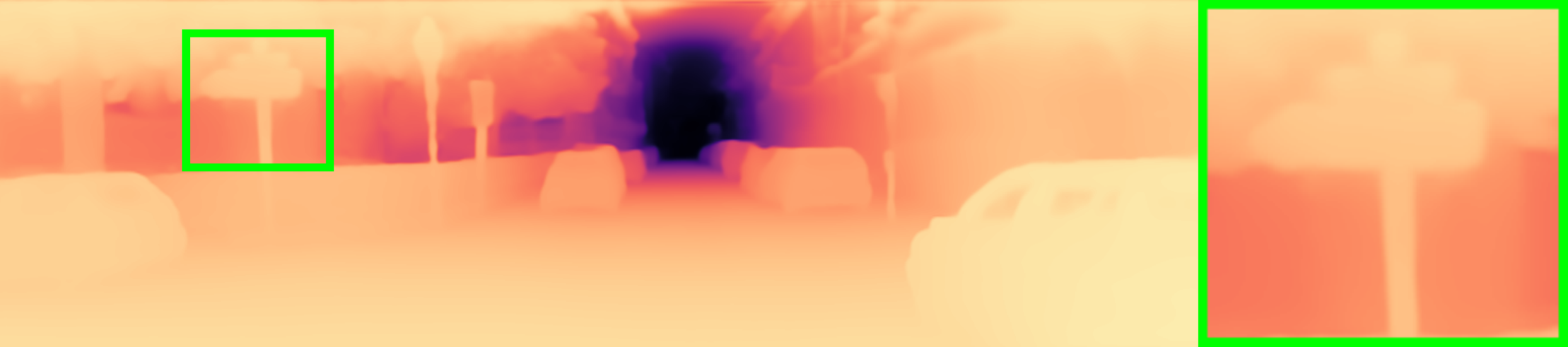}}&  \vspace{0.5pt} 
                \raisebox{-0.4\height}{\includegraphics[width=0.3\textwidth, height=0.05\textheight]{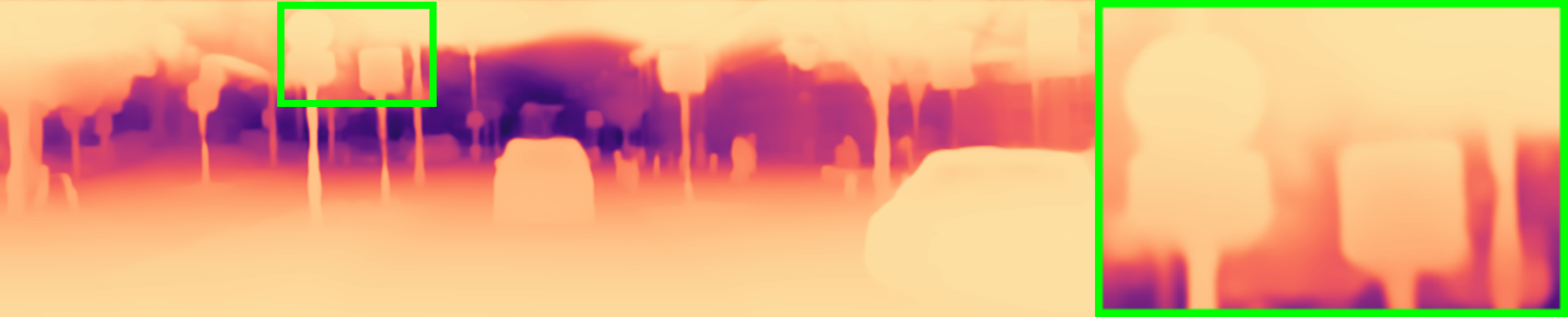}}& \vspace{0.5pt} 
                \raisebox{-0.4\height}{\includegraphics[width=0.3\textwidth, height=0.05\textheight]{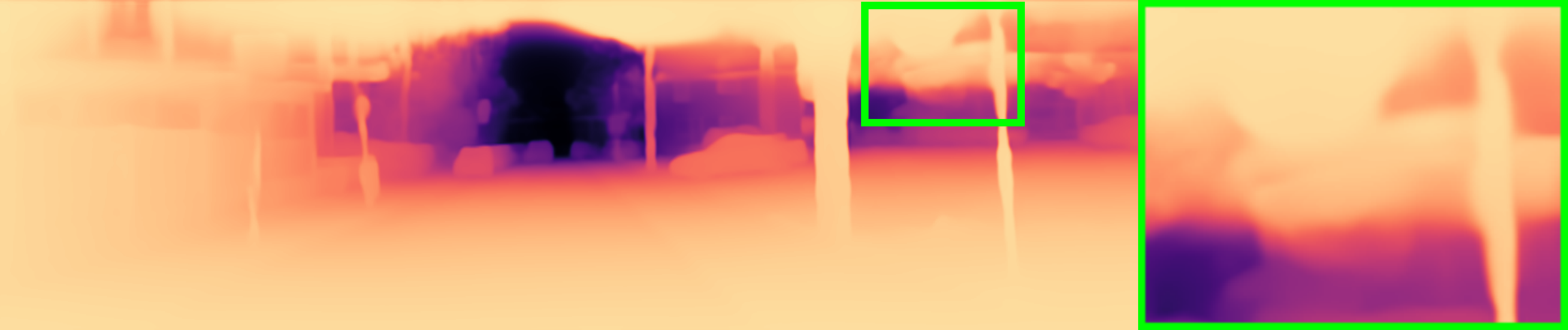}} \\ 
            &  {\bf TIE-KD}\,&  
                \raisebox{-0.4\height}{\includegraphics[width=0.3\textwidth, height=0.05\textheight]{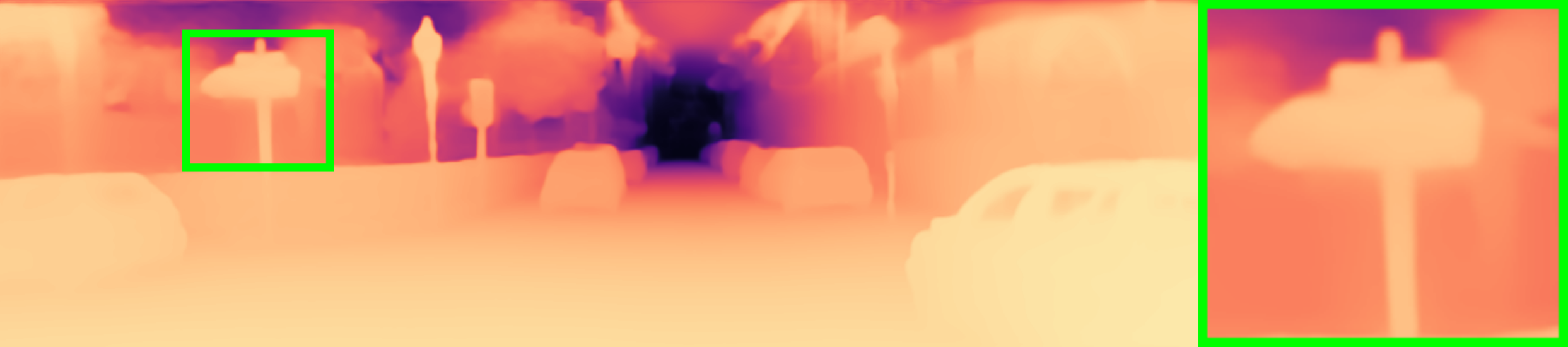}}& \vspace{0.5pt} 
                \raisebox{-0.4\height}{\includegraphics[width=0.3\textwidth, height=0.05\textheight]{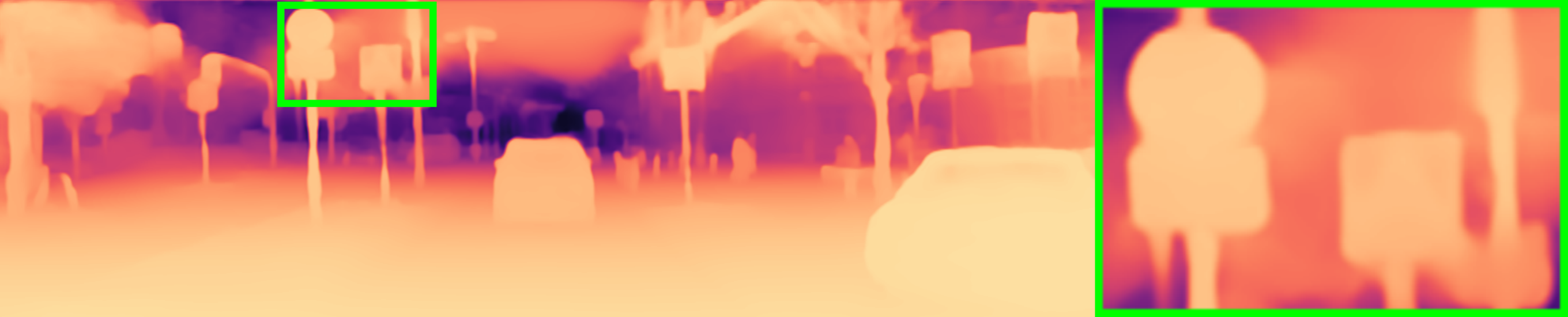}}& \vspace{0.5pt} 
                \raisebox{-0.4\height}{\includegraphics[width=0.3\textwidth, height=0.05\textheight]{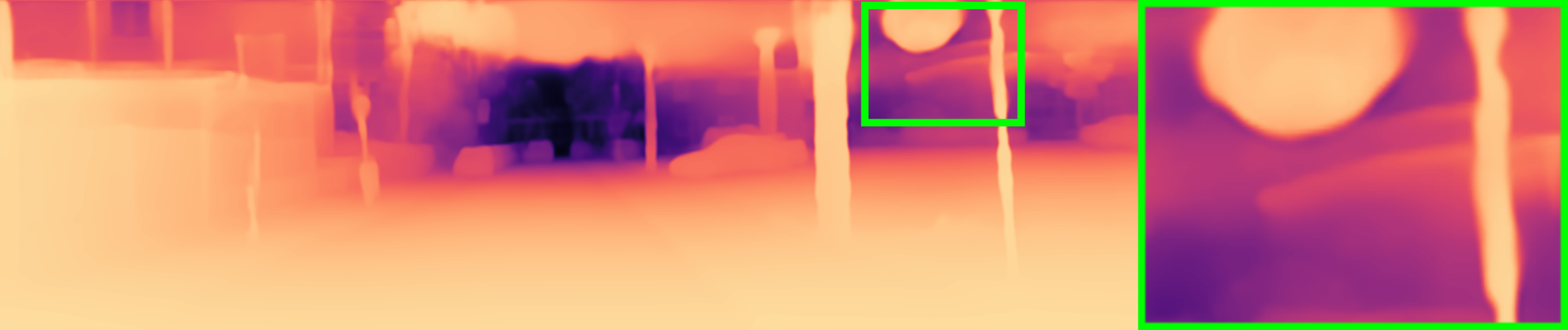}} \\ 
        \bottomrule[0.3pt]
        
        \toprule[0.3pt]
            \multirow{4}{*}{\rotatebox{90}{\parbox[c]{0.2\linewidth}{\centering \small BTS}}} 
            &  {\it Teacher}\,&  
                \raisebox{-0.4\height}{\includegraphics[width=0.3\textwidth, height=0.05\textheight]{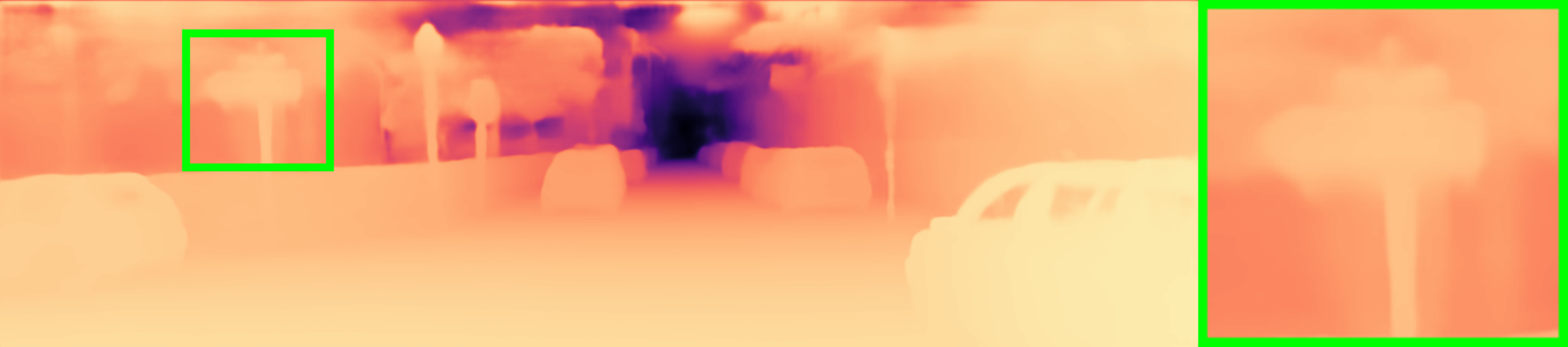}}& \vspace{0.5pt} 
                \raisebox{-0.4\height}{\includegraphics[width=0.3\textwidth, height=0.05\textheight]{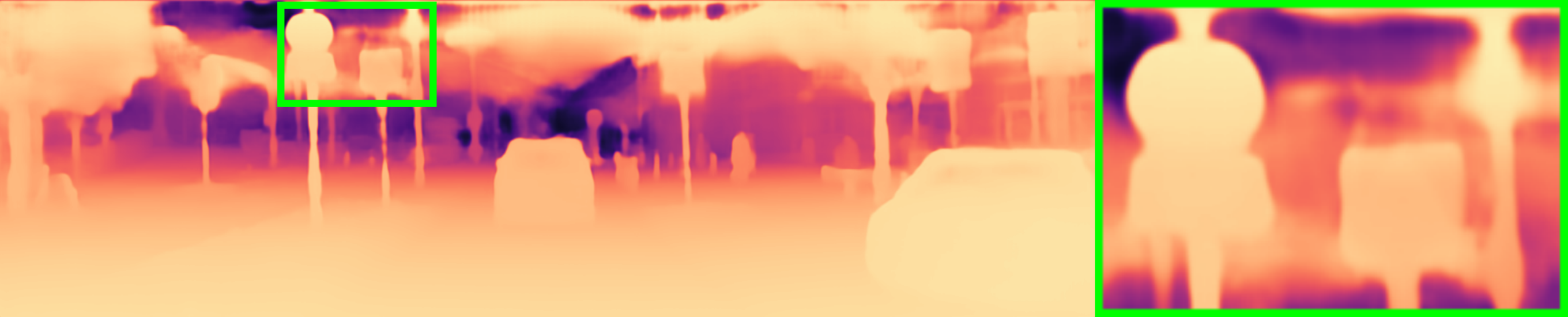}}& \vspace{0.5pt} 
                \raisebox{-0.4\height}{\includegraphics[width=0.3\textwidth, height=0.05\textheight]{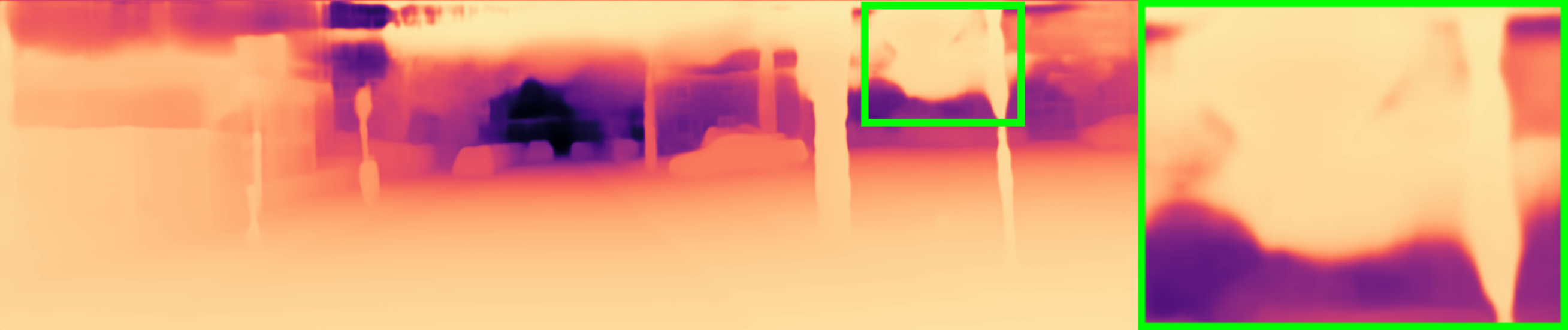}} \\ 
            &  \begin{tabular}{@{}c@{}}Res-KD \\ (SSIM)\end{tabular}\,&  
                \raisebox{-0.4\height}{\includegraphics[width=0.3\textwidth, height=0.05\textheight]{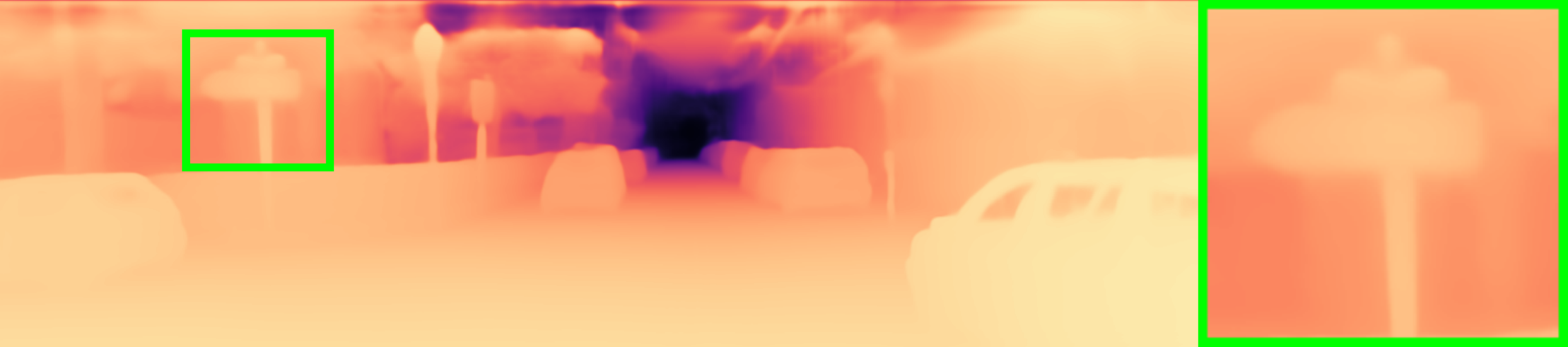}}& \vspace{0.5pt} 
                \raisebox{-0.4\height}{\includegraphics[width=0.3\textwidth, height=0.05\textheight]{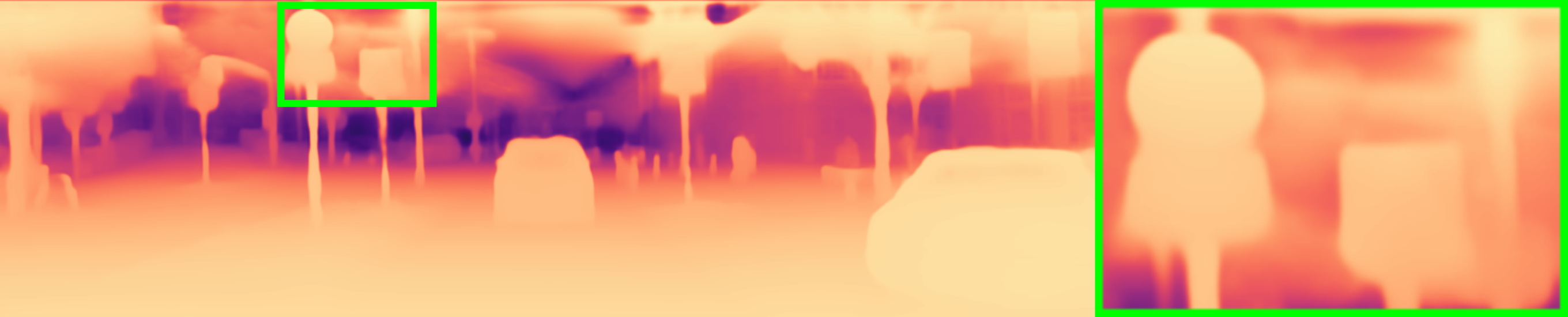}}& \vspace{0.5pt} 
                \raisebox{-0.4\height}{\includegraphics[width=0.3\textwidth, height=0.05\textheight]{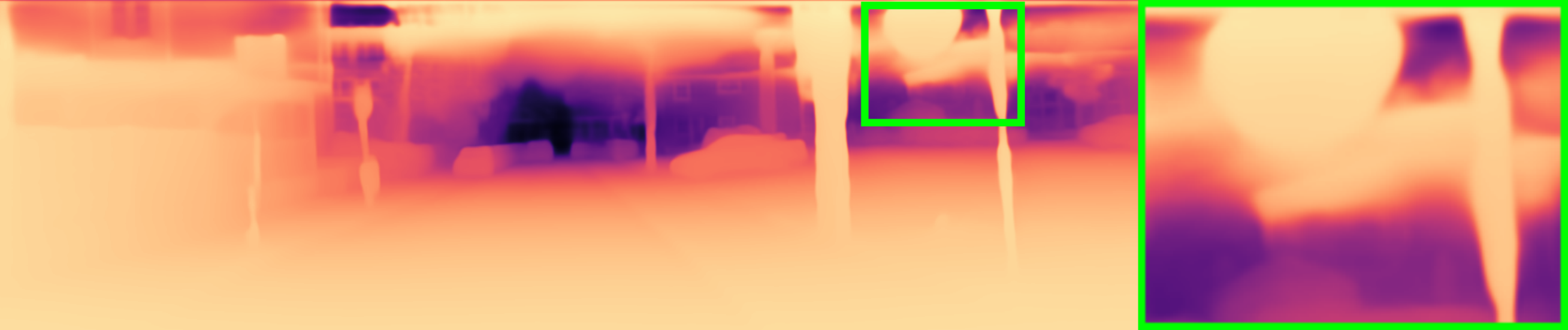}} \\ 
            &  \begin{tabular}{@{}c@{}}Res-KD \\ (SSIM, SI)\end{tabular}\,&  
                \raisebox{-0.4\height}{\includegraphics[width=0.3\textwidth, height=0.05\textheight]{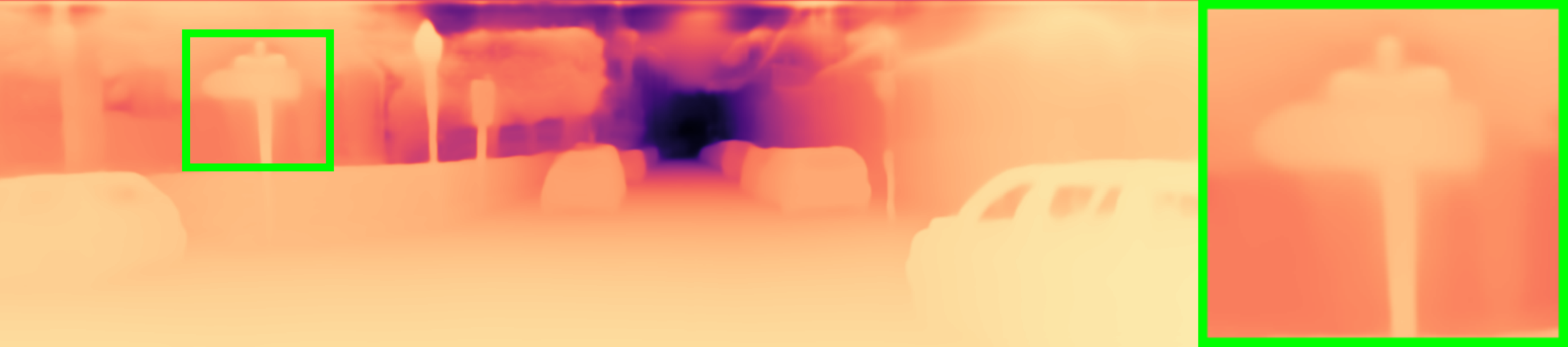}}& \vspace{0.5pt} 
                \raisebox{-0.4\height}{\includegraphics[width=0.3\textwidth, height=0.05\textheight]{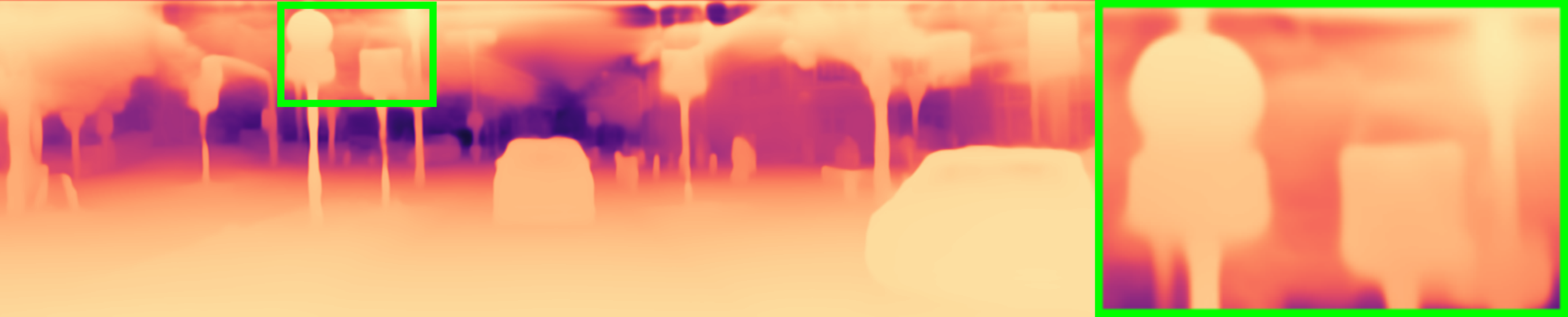}}& \vspace{0.5pt} 
                \raisebox{-0.4\height}{\includegraphics[width=0.3\textwidth, height=0.05\textheight]{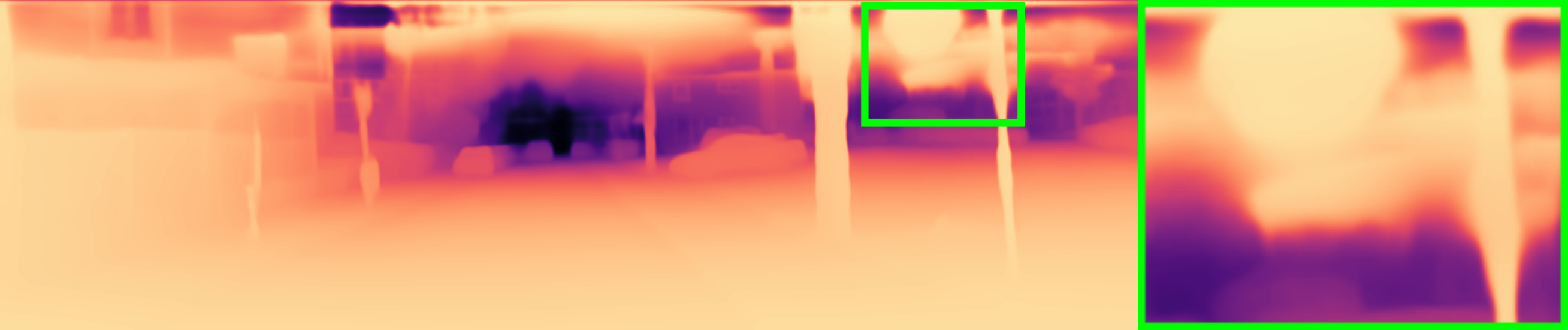}} \\ 
            &  {\bf TIE-KD}\,&  
                \raisebox{-0.4\height}{\includegraphics[width=0.3\textwidth, height=0.05\textheight]{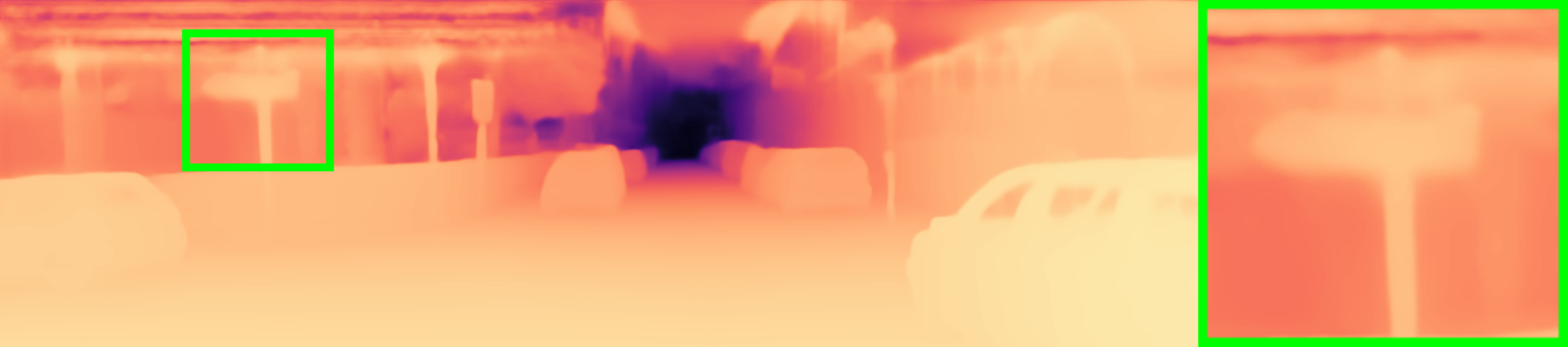}}& \vspace{0.5pt} 
                \raisebox{-0.4\height}{\includegraphics[width=0.3\textwidth, height=0.05\textheight]{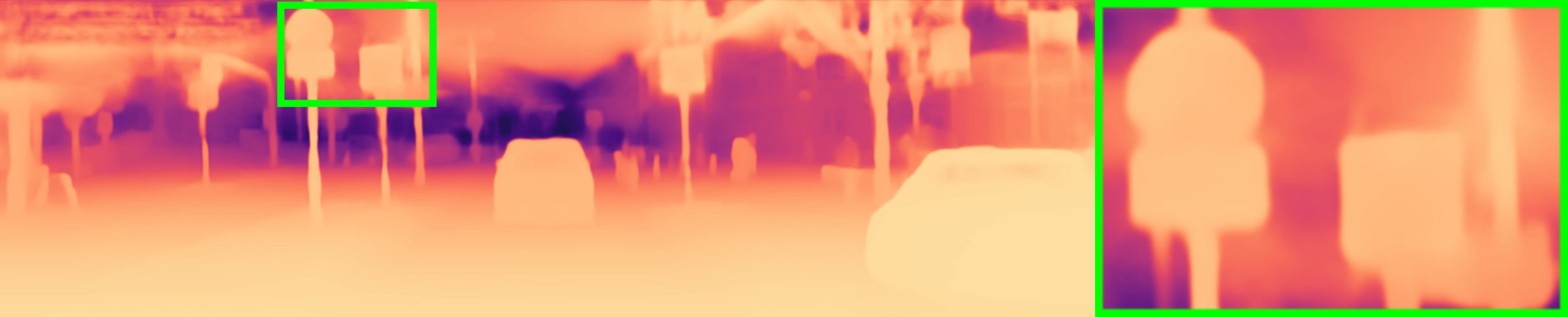}}& \vspace{0.5pt} 
                \raisebox{-0.4\height}{\includegraphics[width=0.3\textwidth, height=0.05\textheight]{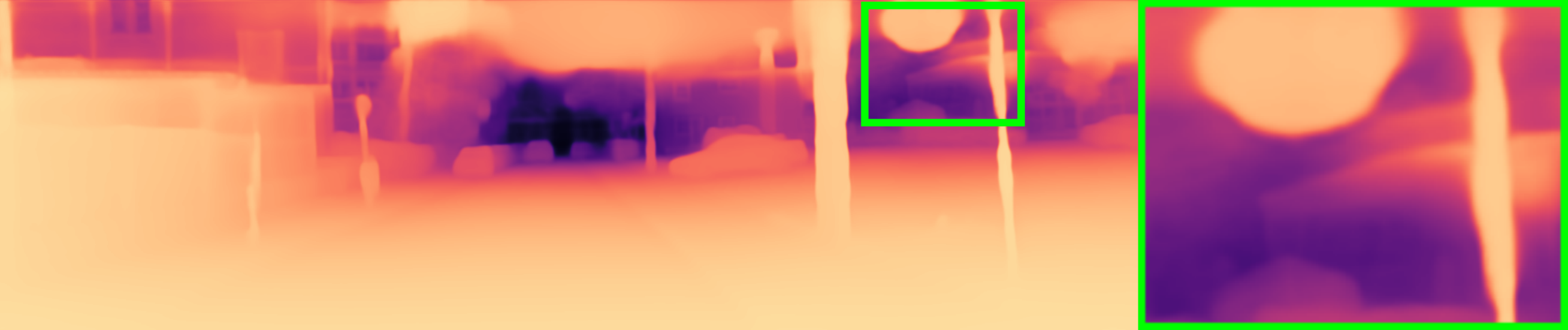}} \\ 
        \bottomrule[0.3pt]
        
        \toprule[0.3pt]
            \multirow{4}{*}{\rotatebox{90}{\parbox[c]{0.2\linewidth}{\centering \small DepthFormer}}} 
            &  {\it Teacher}\,&  
                \raisebox{-0.4\height}{\includegraphics[width=0.3\textwidth, height=0.05\textheight]{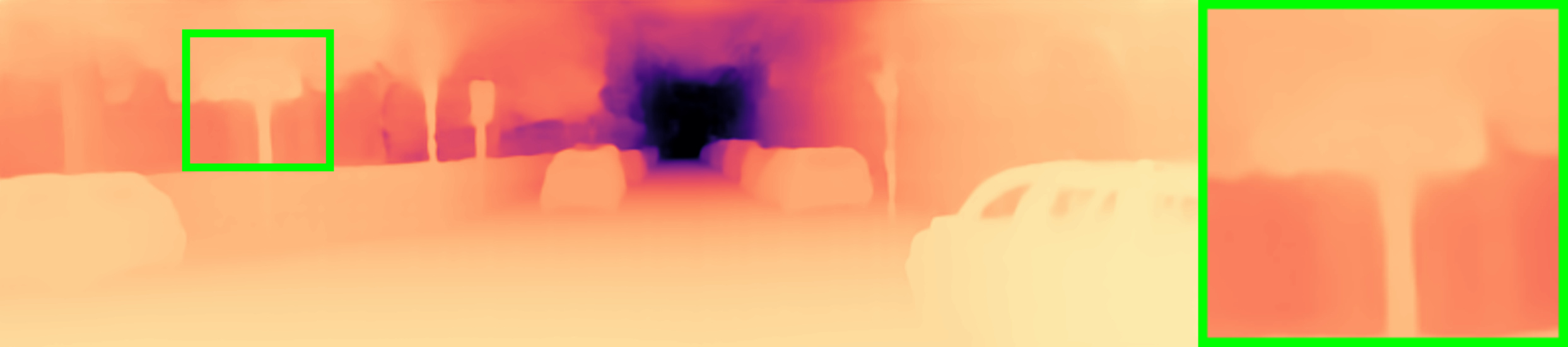}}& \vspace{0.5pt} 
                \raisebox{-0.4\height}{\includegraphics[width=0.3\textwidth, height=0.05\textheight]{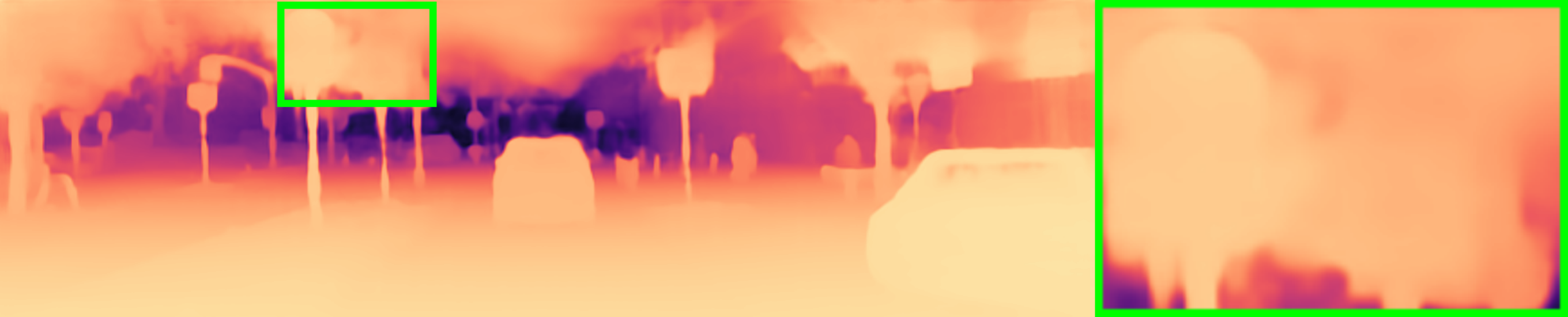}}& \vspace{0.5pt} 
                \raisebox{-0.4\height}{\includegraphics[width=0.3\textwidth, height=0.05\textheight]{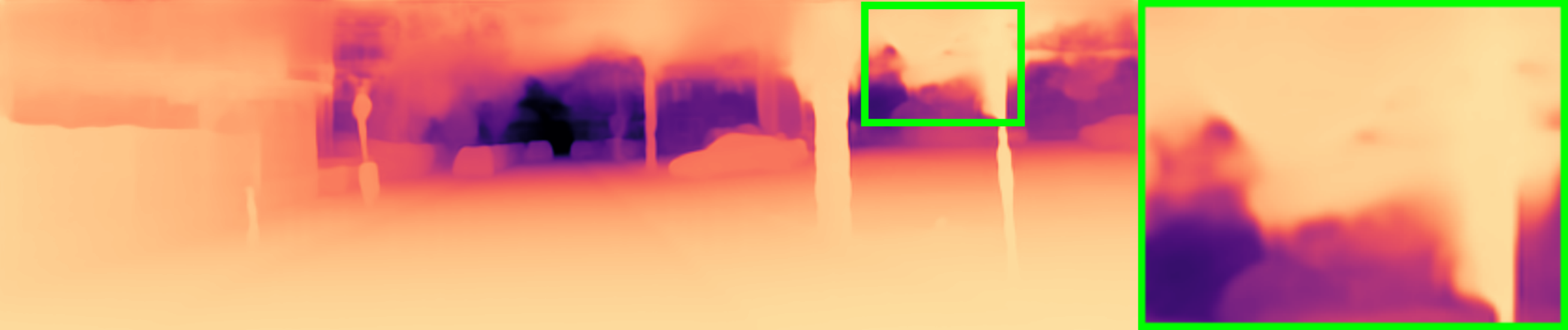}} \\ 
            &  \begin{tabular}{@{}c@{}}Res-KD \\ (SSIM)\end{tabular}\,&  
                \raisebox{-0.4\height}{\includegraphics[width=0.3\textwidth, height=0.05\textheight]{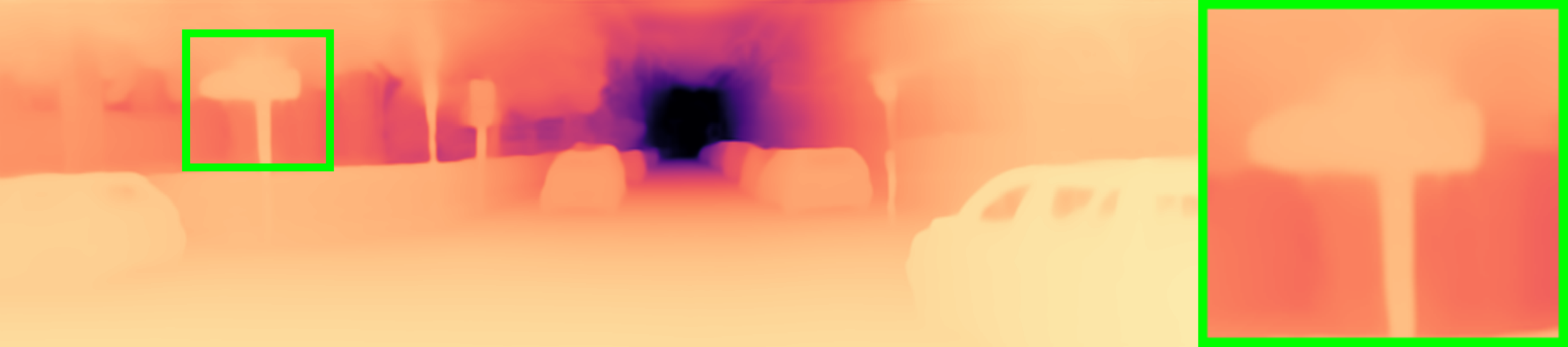}}& \vspace{0.5pt} 
                \raisebox{-0.4\height}{\includegraphics[width=0.3\textwidth, height=0.05\textheight]{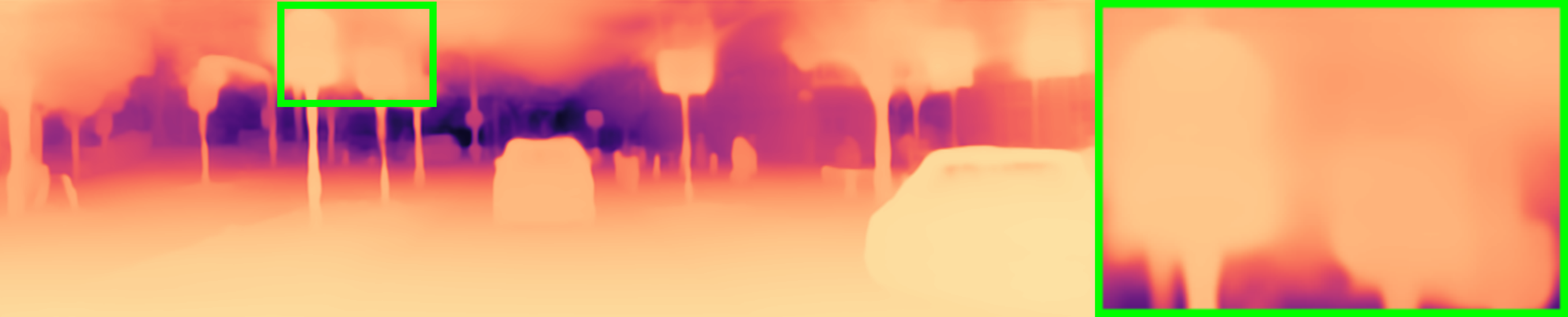}}& \vspace{0.5pt} 
                \raisebox{-0.4\height}{\includegraphics[width=0.3\textwidth, height=0.05\textheight]{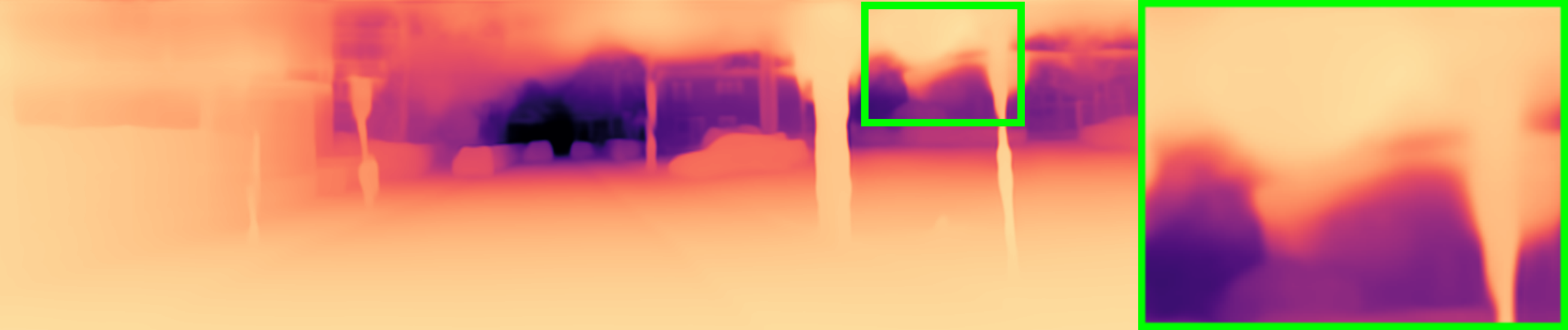}} \\ 
            &  \begin{tabular}{@{}c@{}}Res-KD \\ (SSIM, SI)\end{tabular}\, &
                \raisebox{-0.4\height}{\includegraphics[width=0.3\textwidth, height=0.05\textheight]{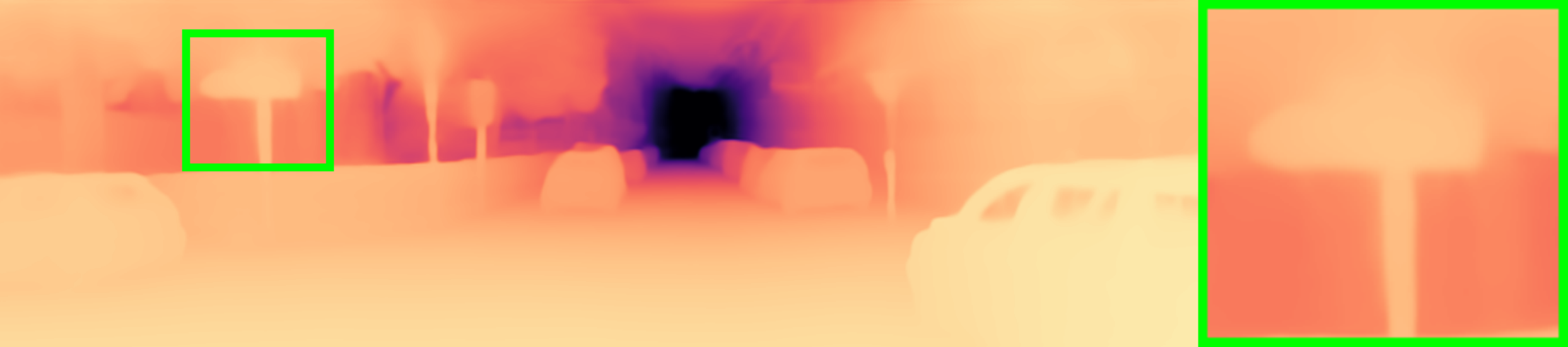}}& \vspace{0.5pt} 
                \raisebox{-0.4\height}{\includegraphics[width=0.3\textwidth, height=0.05\textheight]{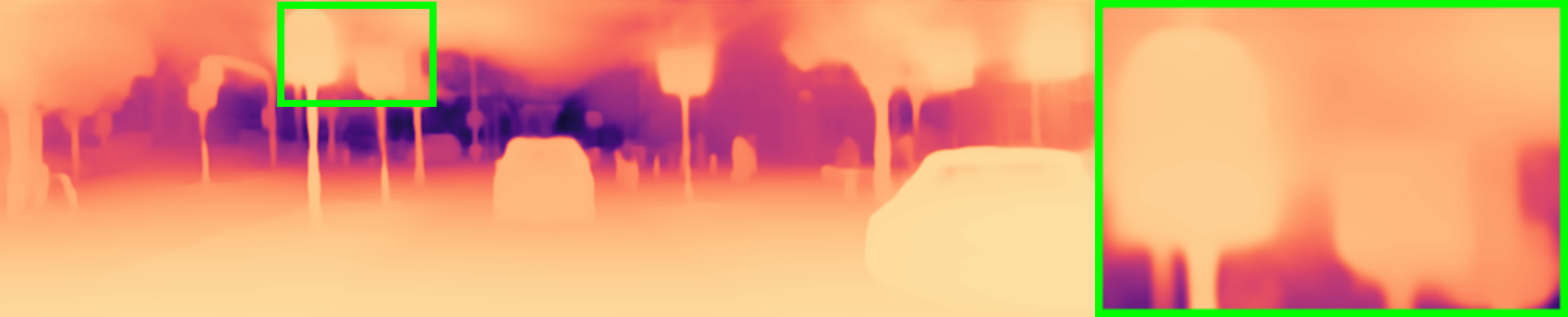}}& \vspace{0.5pt} 
                \raisebox{-0.4\height}{\includegraphics[width=0.3\textwidth, height=0.05\textheight]{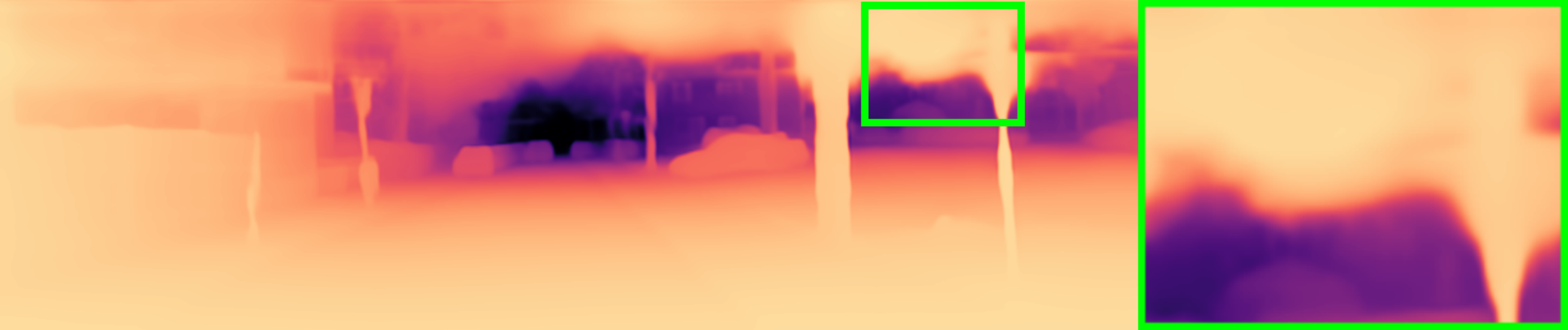}} \\
            &  {\bf TIE-KD}\,&  
                \raisebox{-0.4\height}{\includegraphics[width=0.3\textwidth, height=0.05\textheight]{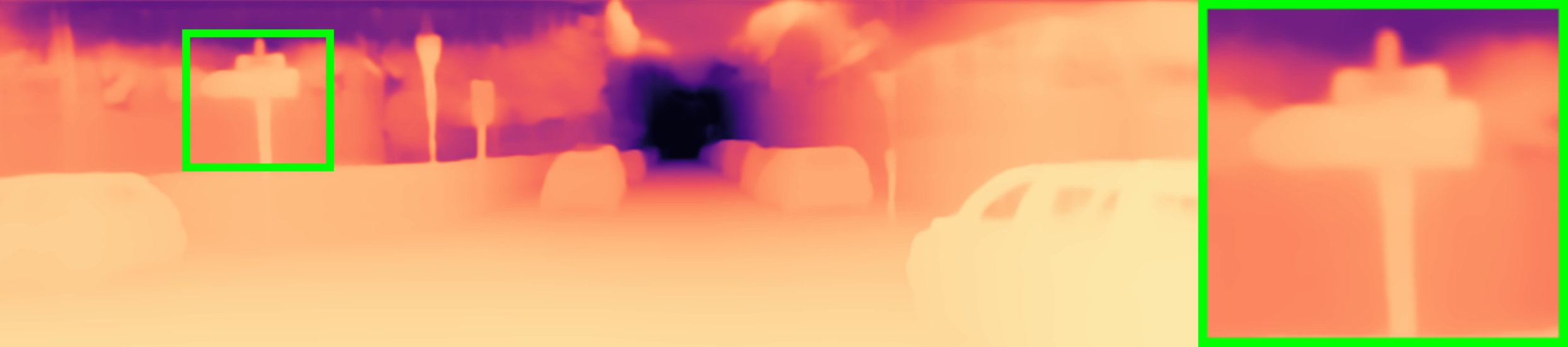}}& \vspace{0.5pt} 
                \raisebox{-0.4\height}{\includegraphics[width=0.3\textwidth, height=0.05\textheight]{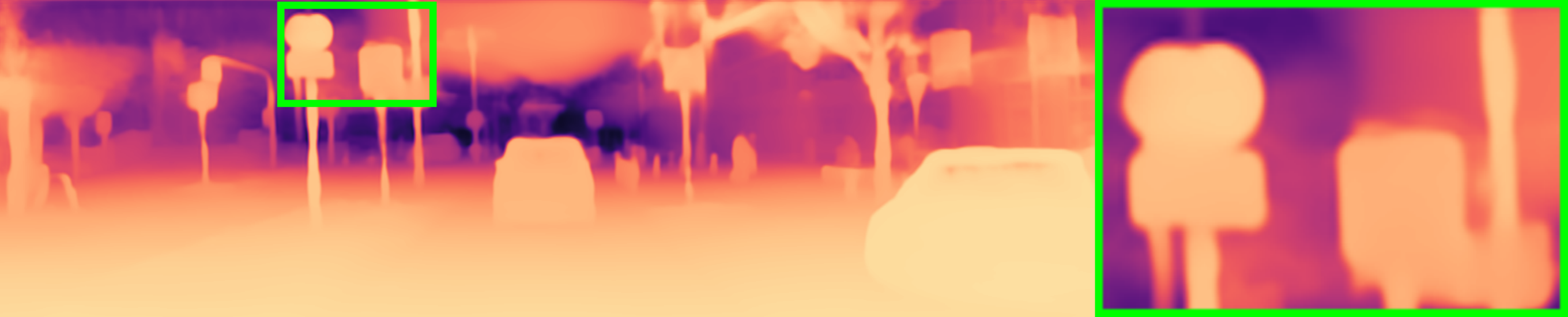}}& \vspace{0.5pt} 
                \raisebox{-0.4\height}{\includegraphics[width=0.3\textwidth, height=0.05\textheight]{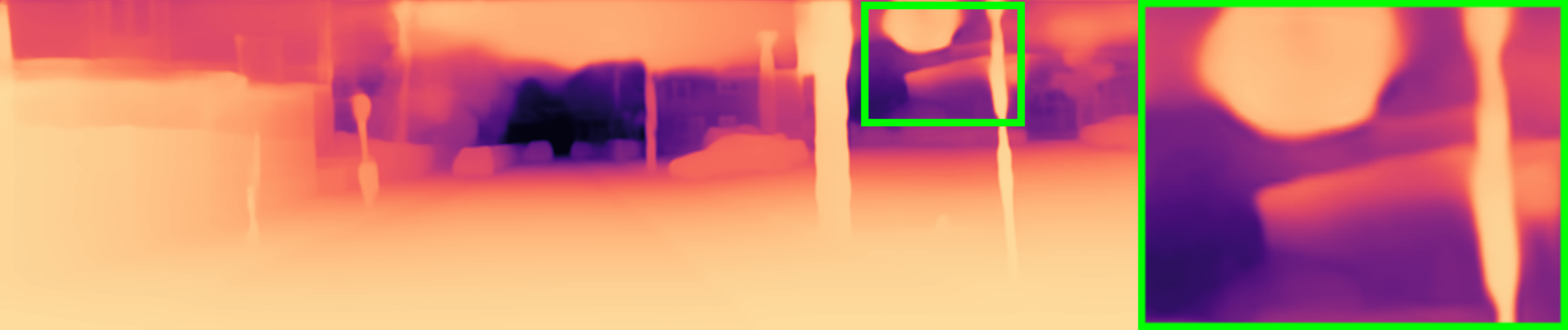}} \\ 
    \end{tabular}
    }
    }
    \caption{Visual comparison of depth maps across various models for three different scenes, highlighting detailed variances within the regions enclosed by green boxes.}
    \label{fig:inference}
\end{figure*}

\paragraph{\bf Qualitative Comparison}
Fig.~\ref{fig:inference} presents a visual comparison of depth maps generated by five models: the baseline, the teacher, and students trained with various KD methods.
The GT from the KITTI dataset is not depicted as it consists of sparse LiDAR points, which differ significantly from the continuous depth map representation.

The depth map from the baseline model is significantly blurrier, particularly around object edges, than that produced by the teacher model.
In contrast, the student model utilizing our TIE-KD framework presents depth estimations remarkably similar to the teacher's, with improved edge definition.
This improvement is exemplified in the middle image of Fig.~\ref{fig:inference}, where the TIE-KD$_{\text{AdaBins}}$ model delineates traffic sign boundaries with greater clarity than the teacher, as evidenced by visual comparisons.

Depth maps from students trained via response-based KD methods are closer to the teacher's output than the baseline but still fall short of the fidelity achieved by TIE-KD, particularly in capturing the fine details at object boundaries and transitions.

\subsection{Similarity to the Teacher Model}\label{subsec:similarity}

\begin{table*}[]

\caption{Comparative analysis of the similarity between teacher and student models based on their outputs.
Student models are identified by the subscript indicating their respective teacher models used in the KD process.``Res-KD" denotes the response-based KD method using SSIM and SI loss functions.
The best performances, marked in bold, signify the closest alignment with the teacher's output.
}
\label{table:similarityTeacher}
\centering
\resizebox{\textwidth}{!}{
\begin{tabular}{|c|ccc|ccc|ccc|}
\hline
 
 Evaluation target &
  \multicolumn{3}{c|}{AdaBins's output} &
  \multicolumn{3}{c|}{BTS's ouput} &
  \multicolumn{3}{c|}{DepthFormer's output} \\ \hline
Metric &
  \multicolumn{1}{c}{AbsRel$\downarrow$} &   \multicolumn{1}{c}{RMSE$\downarrow$} &   $\delta_1$$\uparrow$ &
  \multicolumn{1}{c}{AbsRel$\downarrow$} &   \multicolumn{1}{c}{RMSE$\downarrow$} &   $\delta_1$$\uparrow$ &
  \multicolumn{1}{c}{AbsRel$\downarrow$}  &  \multicolumn{1}{c}{RMSE$\downarrow$} &  $\delta_1$$\uparrow$ \\ \hline
Res-KD$_{AdaBins}$ &
  \multicolumn{1}{c}{0.0640} &   \multicolumn{1}{c}{2.2512} &     0.9598 &
  \multicolumn{1}{c}{0.0675} &   \multicolumn{1}{c}{2.3932} &  0.9564 &
  \multicolumn{1}{c}{0.0757} &   \multicolumn{1}{c}{2.4273} &  0.9495
  \\ 
TIE-KD$_{AdaBins}$ &
  \multicolumn{1}{c}{\bf{0.0612}} &  \multicolumn{1}{c}{\bf{2.0905}} &  \bf{0.9650} &
  \multicolumn{1}{c}{{0.0669}} &  \multicolumn{1}{c}{{2.3572}} &  {0.9573} &
  \multicolumn{1}{c}{{0.0733}} &  \multicolumn{1}{c}{{2.2698}} &  {0.9533}
  \\ \hline
Res-KD$_{BTS}$ &
  \multicolumn{1}{c}{0.0673} &  \multicolumn{1}{c}{2.5704} &  0.9528 &
  \multicolumn{1}{c}{0.0635} &  \multicolumn{1}{c}{2.2174} &  0.9619 &
  \multicolumn{1}{c}{0.0757} &  \multicolumn{1}{c}{2.6581} &  0.9482
  \\ 
TIE-KD$_{BTS}$ &
  \multicolumn{1}{c}{{0.0627}} &  \multicolumn{1}{c}{{2.4227}} &  {0.9588} &
  \multicolumn{1}{c}{\bf{0.0613}} &  \multicolumn{1}{c}{\bf{2.1385}} &    \bf{0.9642} &
  \multicolumn{1}{c}{{0.0726}} &  \multicolumn{1}{c}{{2.5311}} &  {0.9523}
  \\ \hline
Res-KD$_{DepthFormer}$ &
  \multicolumn{1}{c}{0.0667} &  \multicolumn{1}{c}{2.4252} &  0.9569 &
  \multicolumn{1}{c}{{0.0682}} &  \multicolumn{1}{c}{2.6015} &  {0.9576} &
  \multicolumn{1}{c}{0.0710} &  \multicolumn{1}{c}{2.2278} &  0.9580
  \\ 
TIE-KD$_{DepthFormer}$ &
  \multicolumn{1}{c}{\bf{0.0646}} &  \multicolumn{1}{c}{{2.3254}} &  {\bf{0.9604}} &
  \multicolumn{1}{c}{0.0683} &  \multicolumn{1}{c}{{2.5579}} &  0.9565 &
  \multicolumn{1}{c}{{0.0704}} &  \multicolumn{1}{c}{\bf{2.1605}} &  {0.9582}
  \\ \hline
\end{tabular}%
}
\end{table*}

A crucial goal of knowledge distillation is the accurate transfer of knowledge from the teacher to the student model.
To evaluate how well our TIE-KD framework preserves the teacher's knowledge, we compared the similarity between the outputs of the teacher and the student models using three metrics: AbsRel, RMSE, and $\delta_1$.
For this evaluation, we excluded the top 110 pixels, consistent with the rationale provided in Sec.~\ref{subsec:impl}.
The results, presented in Table~\ref{table:similarityTeacher}, show that students trained with TIE-KD more closely mirror their respective teacher models than those trained by other teachers not involved in their training process.
While Res-KD demonstrates some degree of correlation with the teachers' outputs, the TIE-KD students generally exhibit a higher level of similarity.
These results underscore the efficacy of the TIE-KD framework in faithfully transferring the teacher's knowledge to the student model.

Notably, TIE-KD$_{DepthFormer}$ and Res-KD$_{DepthFormer}$ students showed less alignment with their respective teacher model.
This divergence is likely attributed to the substantial differences in parameter counts, consistent with the findings of Mirzadeh et al.~\cite{mirzadeh2020improved}.
However, when applying the TIE-KD framework to a student model equipped with a larger ResNet50 backbone, which encompasses approximately 78M parameters, there is a notable increase in the similarity of the student model's output to that of the DepthFormer.
This observation, as presented in Table~\ref{table:sim_depthformer}, suggests that minimizing the disparity in parameters between teacher and student models can potentially improve the efficiency of knowledge transfer in the TIE-KD framework.


\begin{table}[]
\centering
\footnotesize
\caption{Comparison of TIE-KD student models with ResNet50 backbone (78.3M) against various teacher model outputs}
\label{table:sim_depthformer}
\begin{tabular}{|c|ccc|}
\hline
Evaluation target & AbsRel$\downarrow$ & RMSE$\downarrow$ & $\delta_1$$\uparrow$ \\ \hline
AdaBins's output     &  0.0561             & 2.1732           & 0.9675               \\ \hline
BTS's output         & {\bf0.0559}             & 2.3654           & 0.9718               \\ \hline
DepthFormer's output & 0.0597             & {\bf 1.9085}          & {\bf 0.9738}               \\ \hline
\end{tabular}%
\end{table}

\subsection{Ablation Study}\label{subsec:abalation}

\subsubsection{Impact of Loss Functions}\label{subsub:abal_loss}
\begin{table}[t]
\caption{
    Performance impact of different loss function configurations within the TIE-KD framework. Each row compares the outcomes when employing $L_{DPM}$, $L_{depth}$, or their combination.
    The best results are highlighted in bold.}
\label{table:loss}
\footnotesize
\centering
\begin{tabular}{|cccccc|}
\hline
\multicolumn{1}{|c|}{$L_{DPM}$} & \multicolumn{1}{c|}{$L_{depth}$} & \multicolumn{1}{c|}{AbsRel$\downarrow$} & \multicolumn{1}{c|}{SqRel$\downarrow$} & \multicolumn{1}{c|}{RMSE$\downarrow$} & $\delta_1$$\uparrow$
\\ \hline \hline 
\multicolumn{6}{|c|}{AdaBins}                                                                                                                                         \\ \hline
\multicolumn{1}{|c|}{\checkmark} & \multicolumn{1}{c|}{}           & \multicolumn{1}{c|}{0.0718} & \multicolumn{1}{c|}{0.2485} & \multicolumn{1}{c|}{2.5433} & 0.9398 \\ 
\multicolumn{1}{|c|}{}           & \multicolumn{1}{c|}{\checkmark} & \multicolumn{1}{c|}{0.0696} & \multicolumn{1}{c|}{0.2268} & \multicolumn{1}{c|}{2.4646} & 0.9468 \\ 
\multicolumn{1}{|c|}{\checkmark} & \multicolumn{1}{c|}{\checkmark} & \multicolumn{1}{c|}{\bf 0.0654} & \multicolumn{1}{c|}{\bf 0.2179} & \multicolumn{1}{c|}{\bf 2.4315} & {\bf 0.9540}
\\ \hline \hline 
\multicolumn{6}{|c|}{BTS}                                                                                                                                             \\ \hline
\multicolumn{1}{|c|}{\checkmark} & \multicolumn{1}{c|}{}           & \multicolumn{1}{c|}{0.0722} & \multicolumn{1}{c|}{0.2568} & \multicolumn{1}{c|}{2.6459} & 0.9385 \\ 
\multicolumn{1}{|c|}{}           & \multicolumn{1}{c|}{\checkmark} & \multicolumn{1}{c|}{0.0679} & \multicolumn{1}{c|}{0.2315} & \multicolumn{1}{c|}{2.5694} & 0.9477 \\ 
\multicolumn{1}{|c|}{\checkmark} & \multicolumn{1}{c|}{\checkmark} & \multicolumn{1}{c|}{\bf 0.0656} & \multicolumn{1}{c|}{\bf 0.2247} & \multicolumn{1}{c|}{\bf 2.4984} & {\bf 0.9523}
\\ \hline \hline 
\multicolumn{6}{|c|}{DepthFormer}                                                                                                                                     \\ \hline
\multicolumn{1}{|c|}{\checkmark} & \multicolumn{1}{c|}{}           & \multicolumn{1}{c|}{0.0713} & \multicolumn{1}{c|}{0.2438} & \multicolumn{1}{c|}{2.5241} & 0.9411 \\ 
\multicolumn{1}{|c|}{}           & \multicolumn{1}{c|}{\checkmark} & \multicolumn{1}{c|}{0.0698} & \multicolumn{1}{c|}{0.2299} & \multicolumn{1}{c|}{2.4805} & 0.9458 \\ 
\multicolumn{1}{|c|}{\checkmark} & \multicolumn{1}{c|}{\checkmark} & \multicolumn{1}{c|}{\bf 0.0657} & \multicolumn{1}{c|}{\bf 0.2208} & \multicolumn{1}{c|}{\bf 2.4402} & {\bf 0.9534} \\ \hline
\end{tabular}%
\end{table}

Table~\ref{table:loss} demonstrates the impact of utilizing our two proposed loss functions within the TIE-KD framework. When applied independently, both $L_{DPM}$ and $L_{depth}$ achieved performances comparable to the baseline, with $L_{depth}$ marginally outperforming $L_{DPM}$.
However, the combined application of $L_{DPM}$ and $L_{depth}$ led to the most substantial performance gains.
We also explored the influence of the weighting factor ($\alpha$) between these two loss functions, varying $\alpha$ from 0.1 to 0.9.
As a result, we found around 0.1 typically yields the best performance.

These results affirm our proposition that the depth probability map not only conveys the teacher's knowledge more effectively to the student but also acts as a beneficial regularizer, significantly enhancing the student's performance.

\subsubsection{Flexibility Across Different Student Backbones}\label{subsubsec:abal_backbone}

The adaptability of our TIE-KD framework to diverse network architectures was evaluated by employing alternative backbones for the student model.
We incorporated ResNet architectures with varying capacities, specifically ResNet18 and ResNet50~\cite{he2016deep}, into our student models.
The ResNet18-based student model was comparable in capacity to our original student model, whereas the ResNet50-based student model was the most capacious among those tested.

Performance outcomes, as delineated in Table~\ref{table:backbone}, confirm the anticipated trend that increased backbone capacity correlates with improved baseline performance.
Remarkably, the TIE-KD students, particularly those with the ResNet50 backbone, approached the performance of the teacher model (AdaBins).
Moreover, across all backbone architectures, TIE-KD-trained models consistently outperformed their baseline equivalents, attesting to the TIE-KD framework’s effectiveness and its robustness with various student architectures.

\begin{table}[]
\caption{Comparison of student network performance across different backbone architectures.
The baseline is trained on the KITTI dataset, while the student is trained using the TIE-KD framework with AdaBins as the teacher.
The term `model size' refers to the total number of parameters of the student model including the backbone network.
}
\label{table:backbone}
\centering
\footnotesize
\begin{tabular}{|c|c|c|c|c|}
\hline
\begin{tabular}[c]{@{}c@{}}Backbone\\      (model size)\end{tabular} & \begin{tabular}[c]{@{}c@{}}Method\end{tabular} & AbsRel$\downarrow$ & RMSE$\downarrow$ & $\delta_1$$\uparrow$ \\ \hline
\multirow{2}{*}{\begin{tabular}[c]{@{}c@{}}MobileNetV2\\      (17.6M)\end{tabular}} & baseline & 0.0663 & 2.5625 & 0.9501 \\
                                                                                   & TIE-KD   & 0.0654 & 2.4315 & 0.9540 \\ \hline
\multirow{2}{*}{\begin{tabular}[c]{@{}c@{}}ResNet18\\      (16.7M)\end{tabular}}   & baseline & 0.0634 & 2.5311 & 0.9531 \\
                                                                                   & TIE-KD   & 0.0628 & 2.4029 & 0.9559 \\ \hline
\multirow{2}{*}{\begin{tabular}[c]{@{}c@{}}ResNet50\\      (78.3M)\end{tabular}}   & baseline & 0.0605 & 2.4159 & 0.9576 \\
                                                                                   & TIE-KD   & 0.0596 & 2.3060 & 0.9605 \\ \hline
\hline
AdaBins (78M) & Teacher & 0.0593 & 2.3309 & 0.9631 \\
\hline
\end{tabular}%
\end{table}

    






\begin{table*}[h!]
\centering
\caption{Effect of the weight ($\alpha$) of Eq. 4 in the main paper (Sec. 3.3), with AdaBins~\cite{bhat2021adabins} as the teacher model.
Bold values indicate the best performance, and underlined values denote the second-best methods.
}
\label{table:alpha}
\footnotesize
\begin{tabular}{|c||c|c|c|c|c|c|c|}
\hline
$\alpha$ & AbsRel($\downarrow$) & SqRel($\downarrow$) & RMSE($\downarrow$) & RMSE$_{log}$($\downarrow$) & $\delta_1$($\uparrow$) & $\delta_2$($\uparrow$) & $\delta_3$($\uparrow$)\\ \hline
0.05 & \underline{0.0656} & \underline{0.2196} & \underline{2.4374} & \underline{0.0986} & \underline{0.9530} & \underline{0.9936} & 0.9985 \\ \hline
0.1      & \textbf{0.0654} & \textbf{0.2179} & \textbf{2.4315} & \textbf{0.0980} & \textbf{0.9540} & \textbf{0.9939} & 0.9985          \\ \hline
0.15 & 0.0657 & 0.2231 & 2.4564 & 0.0990 & 0.9524 & 0.9935 & 0.9984 \\ \hline
0.2      & 0.0660          & 0.2213          & 2.4376          & 0.0991          & 0.9515          & 0.9936          & \textbf{0.9986} \\ \hline
0.25 & 0.0658 & 0.2220 & 2.4464 & 0.0991 & 0.9516 & 0.9935 & \underline{0.9986} \\ \hline
0.3  & 0.0680 & 0.2281 & 2.4798 & 0.1010 & 0.9497 & 0.9931 & 0.9985 \\ \hline
0.35 & 0.0672 & 0.2312 & 2.4744 & 0.1008 & 0.9513 & 0.9927 & 0.9983 \\ \hline
0.4  & 0.0697 & 0.2319 & 2.4716 & 0.1026 & 0.9478 & 0.9926 & 0.9985 \\ \hline
0.45 & 0.0680 & 0.2327 & 2.4861 & 0.1016 & 0.9491 & 0.9927 & 0.9985 \\ \hline
0.5  & 0.0680 & 0.2336 & 2.4932 & 0.1018 & 0.9485 & 0.9926 & 0.9984 \\ \hline
\end{tabular}
\end{table*}
\begin{figure*}[h!]
\centering
\begin{subfigure}{0.30\textwidth}
    \includegraphics[width=\textwidth]{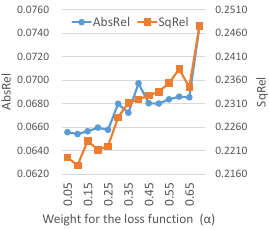}
    \caption{AbsRel and SqRel ($\downarrow$)}
    \label{fig:first}
\end{subfigure}
\hfill
\begin{subfigure}{0.30\textwidth}
    \includegraphics[width=\textwidth]{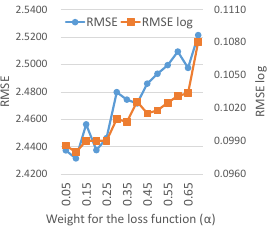}
    \caption{RMSE and RMSE$_{log}$ ($\downarrow$)}
    \label{fig:second}
\end{subfigure}
\hfill
\begin{subfigure}{0.30\textwidth}
    \includegraphics[width=\textwidth]{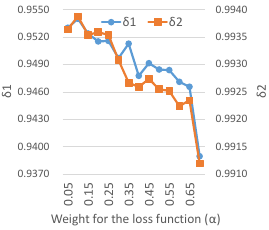}
    \caption{$\delta_1$ and $\delta_2$ ($\uparrow$)}
    \label{fig:third}
\end{subfigure}
\caption{Impact of loss function weight ($\alpha$) on TIE-KD performance, with each subfigure representing a different metric.}
\label{fig:alpha}
\end{figure*}

\begin{table*}[h!]
\centering
\caption{Effect of the standard deviation ($\sigma$) of Eq. 1 in the main paper (Sec. 3.2), with AdaBins~\cite{bhat2021adabins} as the teacher model.
Bold values indicate the best performance, and underlined values denote the second-best methods.
}
\label{table:sigma}
\footnotesize
\begin{tabular}{|c||c|c|c|c|c|c|c|}
\hline
\textbf{$\sigma$} &
  AbsRel($\downarrow$) &
  SqRel($\downarrow$) &
  RMSE($\downarrow$) &
  RMSE$_{log}$($\downarrow$) &
  $\delta_1$($\uparrow$) &
  $\delta_2$($\uparrow$) &
  $\delta_3$($\uparrow$) \\ \hline
0.1 & \underline{0.0647} & 0.2252          & 2.4495          & 0.0984          & 0.9530          & 0.9934          & 0.9984                   \\ \hline
0.2 & \textbf{0.0646} & 0.2227          & 2.4410          & 0.0981          & 0.9530          & 0.9936          & 0.9984                   \\ \hline
0.3 & 0.0652          & 0.2216          & 2.4336          & 0.0984          & \underline{0.9531} & \underline{0.9937} & \underline{0.9984}          \\ \hline
0.4 & 0.0650          & 0.2212          & 2.4363          & \textbf{0.0980} & 0.9529          & \textbf{0.9940} & 0.9986                   \\ \hline
0.5 & 0.0656          & 0.2213          & 2.4329          & 0.0985          & 0.9523          & 0.9938          & 0.9987                   \\ \hline
0.6 & 0.0652          & \underline{0.2186} & 2.4473          & 0.0986          & \underline{0.9531} & \underline{0.9939} & \textbf{0.9985} \\ \hline
0.7 & 0.0655          & 0.2234          & 2.4544          & 0.0985          & 0.9530          & 0.9937          & 0.9985                   \\ \hline
0.8 &
  0.0654 &
  \textbf{0.2179} &
  \underline{2.4315} &
  \textbf{0.0980} &
  \textbf{0.9540} &
  \underline{0.9939} &
  \textbf{0.9985} \\ \hline
0.9 & 0.0664          & 0.2238          & 2.4514          & 0.0992          & 0.9521          & \underline{0.9939} & 0.9985                   \\ \hline
1   & 0.0667          & 0.2213          & 2.4422          & 0.0991          & 0.9530          & \underline{0.9939} & 0.9985                   \\ \hline
1.1 & 0.0668          & 0.2209          & 2.4387          & 0.0991          & 0.9517          & 0.9938          & 0.9986                   \\ \hline
1.2 & 0.0684          & 0.2238          & 2.4600          & 0.1004          & 0.9505          & 0.9936          & 0.9985                   \\ \hline
1.3 & 0.0705          & 0.2237          & 2.4330          & 0.1008          & 0.9513          & 0.9938          & 0.9985                   \\ \hline
1.4 & 0.0709          & 0.2248          & \textbf{2.4307} & 0.1013          & 0.9500          & 0.9938          & 0.9987                   \\ \hline
1.5 & 0.0713          & 0.2256          & 2.4424          & 0.1016          & 0.9503          & 0.9938          & 0.9986                   \\ \hline
\end{tabular}%
\end{table*}
\begin{figure*}[h!]
\centering
\begin{subfigure}{0.30\textwidth}
    \includegraphics[width=\textwidth]{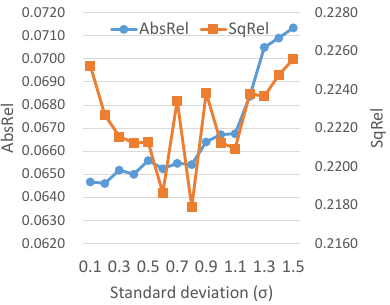}
    \caption{AbsRel and SqRel ($\downarrow$)}
    \label{fig:first}
\end{subfigure}
\hfill
\begin{subfigure}{0.30\textwidth}
    \includegraphics[width=\textwidth]{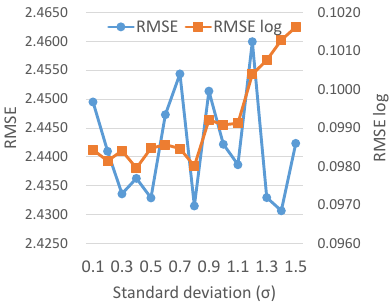}
    \caption{RMSE and RMSE$_{log}$ ($\downarrow$)}
    \label{fig:second}
\end{subfigure}
\hfill
\begin{subfigure}{0.30\textwidth}
    \includegraphics[width=\textwidth]{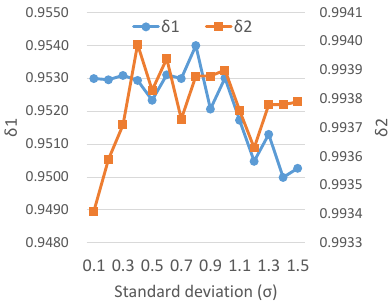}
    \caption{$\delta_1$ and $\delta_2$ ($\uparrow$)}
    \label{fig:third}
\end{subfigure}
        
\caption{Performance variation of the TIE-KD framework with respect to the standard deviation ($\sigma$) used in depth probability map generation.
Each subfigure presents a different metric.
}
\label{fig:sigma}
\end{figure*}

\subsubsection{Effect of the Weight ($\alpha$) for the Loss Function}\label{subsubsec:abal_weight}
The performance of our TIE-KD framework with varying weight \(\alpha\) is detailed in Table~\ref{table:alpha} and Fig.~\ref{fig:alpha}, as specified in Eq. 4 (Sec. 3.3).
By systematically adjusting \(\alpha\) between 0.05 to 0.7 in increments of 0.05 using AdaBins~\cite{bhat2021adabins} as the teacher, we discovered that an \(\alpha\) value of 0.1 generally yielded optimal results across most metrics.


\subsubsection{Effect of the Standard Deviation ($\sigma$) for Depth Probability Map Generation}\label{subsubsec:abal_dev}
Table~\ref{table:sigma} and Fig.~\ref{fig:sigma} illustrate the performance variations in our TIE-KD framework relative to different standard deviation ($\sigma$) values used in Eq. 1 for generating the depth probability map (refer to Section 3.2).
While optimal $\sigma$ values varied across different evaluation metrics, a $\sigma$ value of 0.1 generally yielded good overall performance.
%
%


\section{Conclusion}\label{sec:conclusion}

In this work, we introduced the Teacher-Independent Explainable Knowledge Distillation (TIE-KD) framework, a novel approach to monocular depth estimation.
Central to TIE-KD is the Depth Probability Map (DPM), which enables an efficient distillation process that is not constrained by the architectural compatibility between teacher and student models.
Rigorous testing on the KITTI dataset has shown that TIE-KD surpasses traditional response-based KD methods, demonstrating not only a closer alignment with the teacher model’s performance but also adaptability to various student model architectures.


Our aspirations for the TIE-KD framework are to pave the way for more effective and scalable solutions in monocular depth estimation, contributing significantly to the practical deployment of deep learning models, especially in settings where computational resources are limited.

\ifCLASSOPTIONcompsoc
\fi

\ifCLASSOPTIONcaptionsoff
  \newpage
\fi



%

\bibliographystyle{IEEEtran}




%


\begin{IEEEbiography}[{\includegraphics[width=1in,height=1.25in,clip,keepaspectratio]{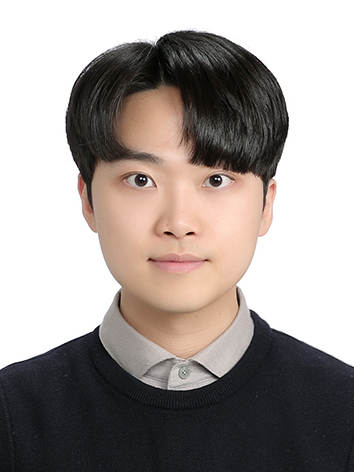}}]
{Sangwon Choi} was received the M.S. degree with the School of Computer Engineering,
Korea University of Technology and Education
(KOREATECH). His research interests include lightweight deep learning, knowledge distillation, autonomous driving.
\end{IEEEbiography}

\begin{IEEEbiography}[{\includegraphics[width=1in,height=1.25in,clip,keepaspectratio]{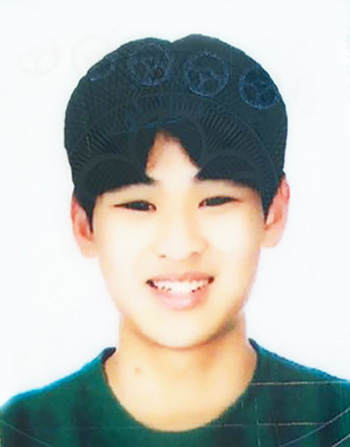}}]
{Daejune Choi} was received the Bachelor's degree with the School of Computer Engineering,
Korea University of Technology and Education
(KOREATECH). His research interests include lightweight deep learning, knowledge distillation, single-image super-resolution.
\end{IEEEbiography}


\begin{IEEEbiography}[{\includegraphics[width=1in,height=1.25in,clip,keepaspectratio]{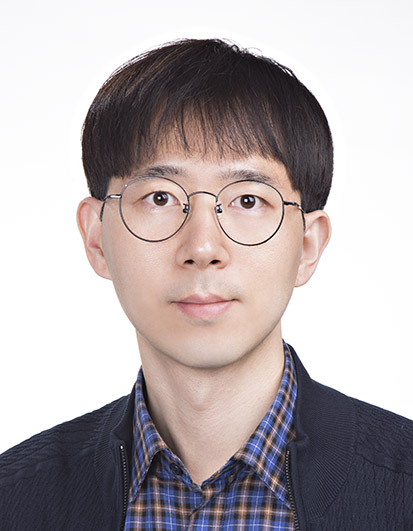}}]{Duksu Kim} is currently an assistant professor in the School of Computer Engineering at KOREATECH (Korea University of Technology and Education).
He received his B.S. from SungKyunKwan University in 2008.
He received his Ph.D. from KAIST (Korea Advanced Institute of Science and Technology) in Computer Science in 2014.
He spent several years as a senior researcher at KISTI National Supercomputing Center.
His research interest is designing heterogeneous parallel computing algorithms for various applications, including proximity computation, scientific visualization, and machine learning.
Some of his work received the distinguished paper award at Pacific Graphics 2009, and an ACM student research competition award in 2009, and was selected as the spotlight paper for the September issue of IEEE Transactions on Visualization and Computer Graphics (TVCG) in 2013.
He is a young professional member of IEEE and a professional member of ACM.
\end{IEEEbiography}




\newpage
\onecolumn
\appendices
\section{Additional Qualitative Comparisons}
\label{sec:sample:appendix}
This section presents additional qualitative comparisons among teacher models, the baseline, response-based KD methods (Res-KD) employing various loss functions, and TIE-KD with distinct loss function configurations.

\subsection{Teacher model: AdaBins~\cite{bhat2021adabins}}
\begin{figure*}[hb!]
    \centering
    \resizebox{0.84\textwidth}{!}{%
    \renewcommand{\arraystretch}{1}
    \setlength{\tabcolsep}{0pt}
    \begin{tabular}{c@{\hspace{0.5em}}@{\hspace{0.5em}}r@{\hspace{0.5em}}c@{\hspace{0.5em}}c@{\hspace{0.5em}}c}
            \multicolumn{2}{c}{ \normalsize Input\,}  & 
                \raisebox{-0.4\height}{\includegraphics[width=0.3\textwidth, height=0.07\textheight]{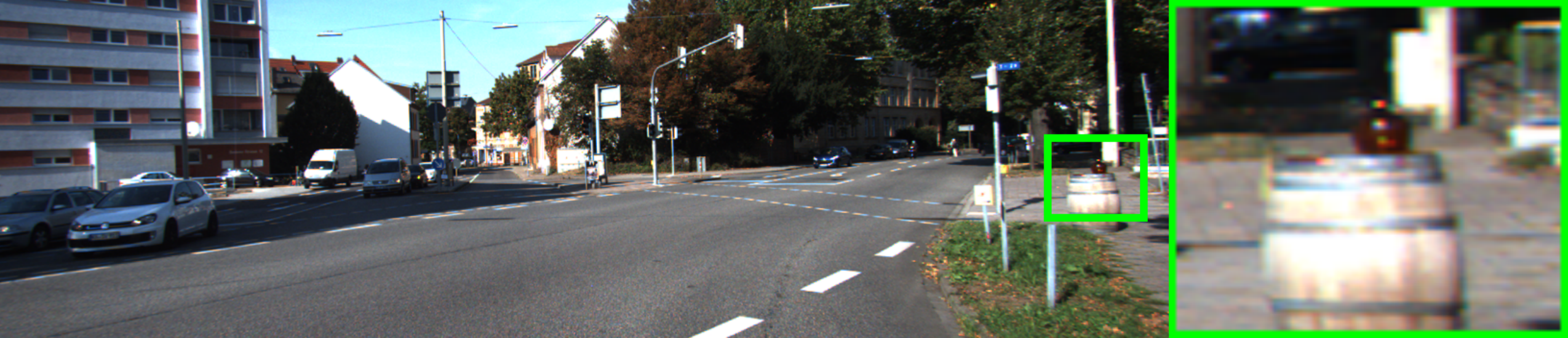}}& \vspace{0.5pt}
                \raisebox{-0.4\height}{\includegraphics[width=0.3\textwidth, height=0.07\textheight]{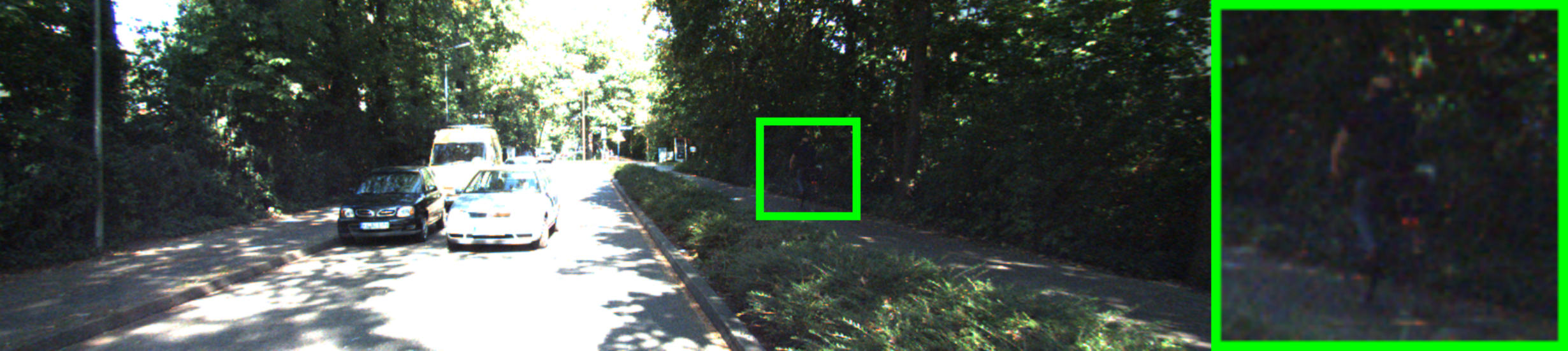}}& \vspace{0.5pt}
                \raisebox{-0.4\height}{\includegraphics[width=0.3\textwidth, height=0.07\textheight]{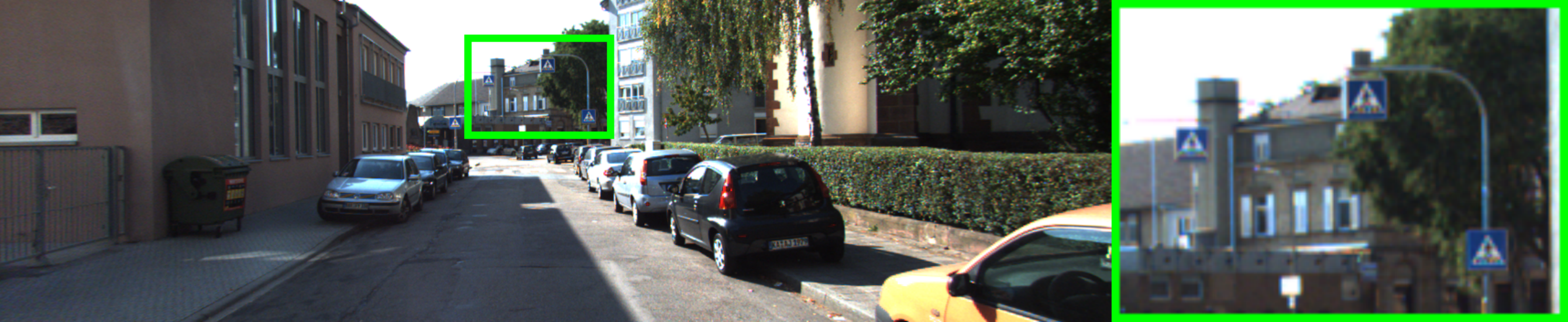}} \\ 
            \multicolumn{2}{c}{ \normalsize Baseline\,}& 
                \raisebox{-0.4\height}{\includegraphics[width=0.3\textwidth, height=0.07\textheight]{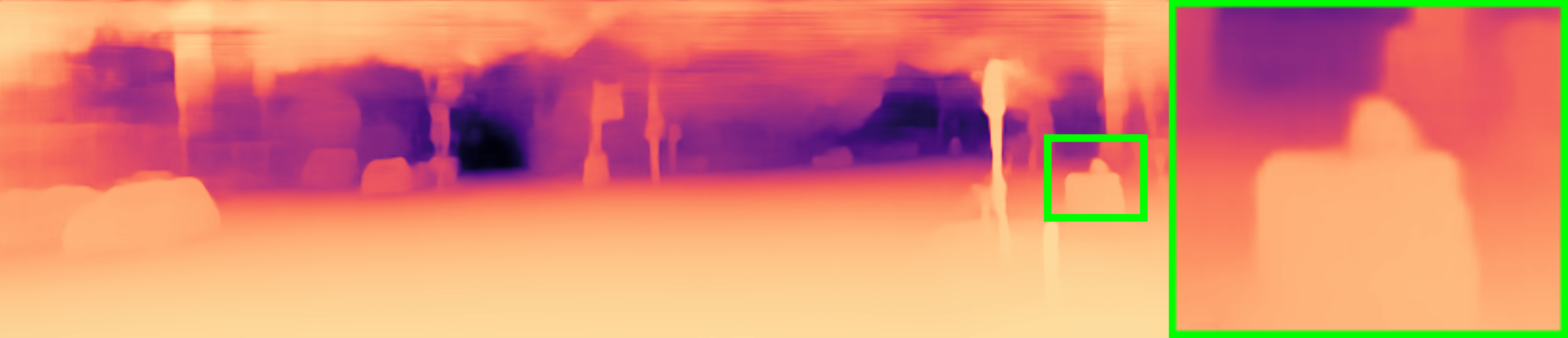}}& \vspace{0.5pt} 
                \raisebox{-0.4\height}{\includegraphics[width=0.3\textwidth, height=0.07\textheight]{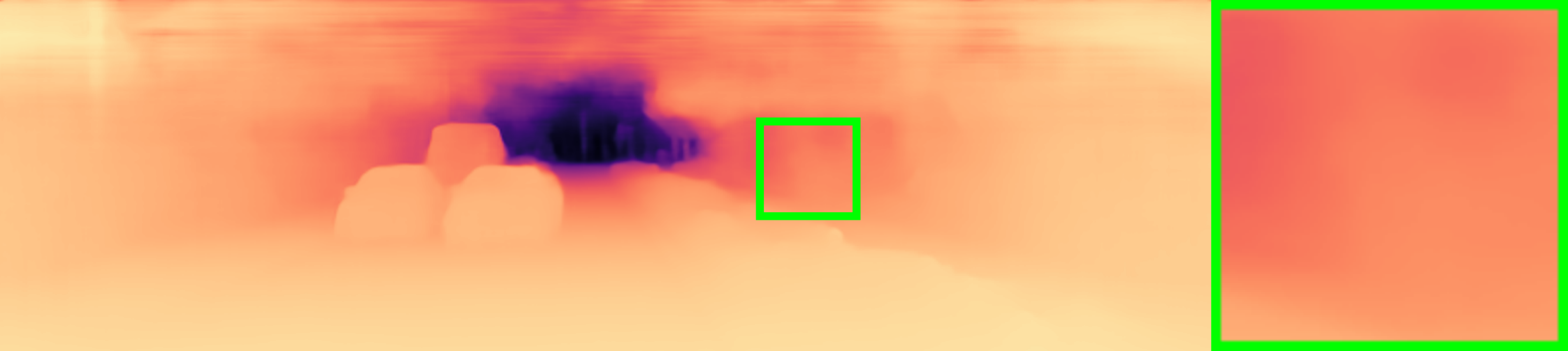}}& \vspace{0.5pt} 
                \raisebox{-0.4\height}{\includegraphics[width=0.3\textwidth, height=0.07\textheight]{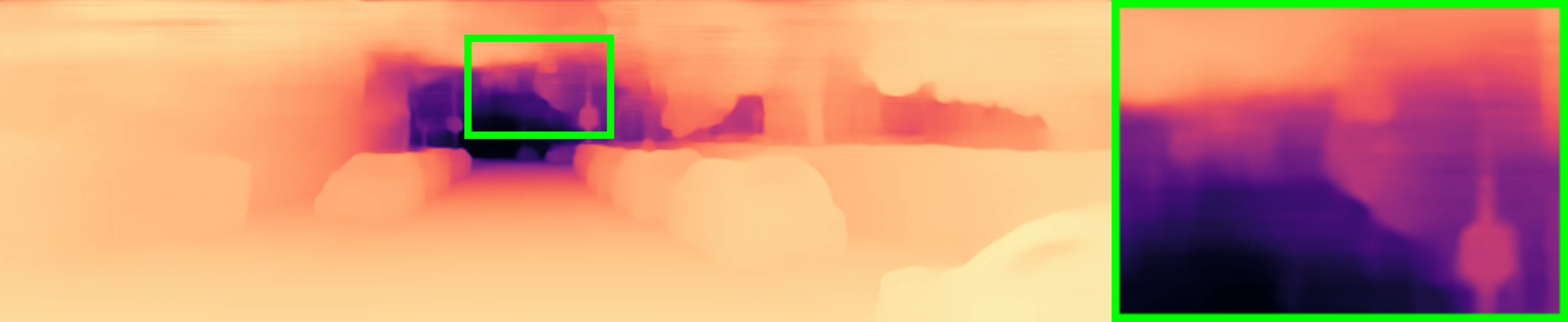}} \\ 
            \multicolumn{2}{c}{ \normalsize Teacher\,}& 
                \raisebox{-0.4\height}{\includegraphics[width=0.3\textwidth, height=0.07\textheight]{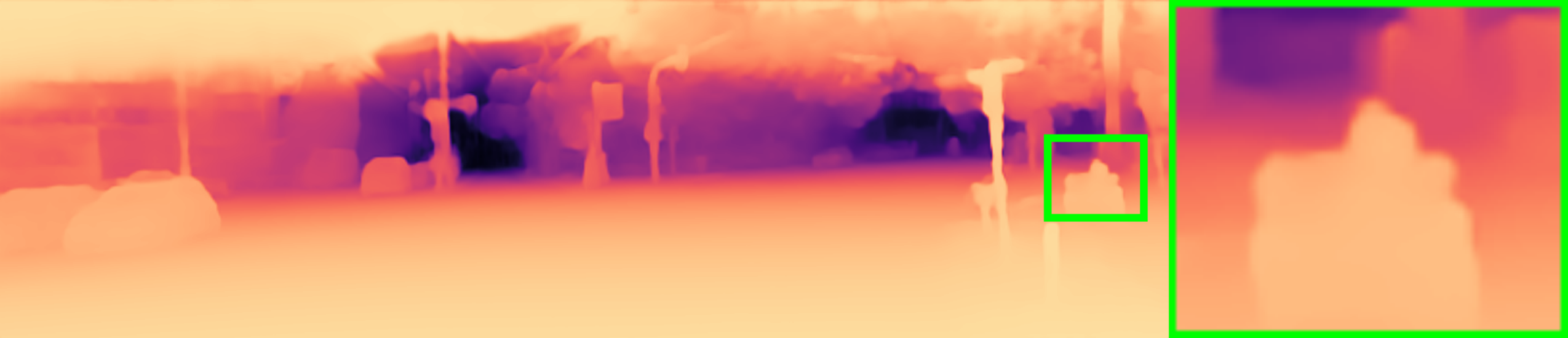}}& \vspace{0.5pt} 
                \raisebox{-0.4\height}{\includegraphics[width=0.3\textwidth, height=0.07\textheight]{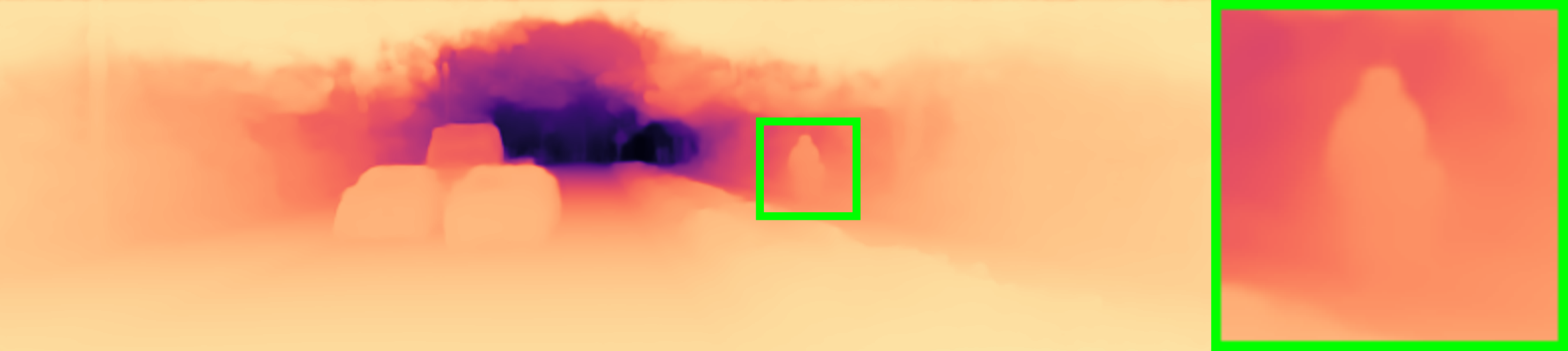}}& \vspace{0.5pt} 
                \raisebox{-0.4\height}{\includegraphics[width=0.3\textwidth, height=0.07\textheight]{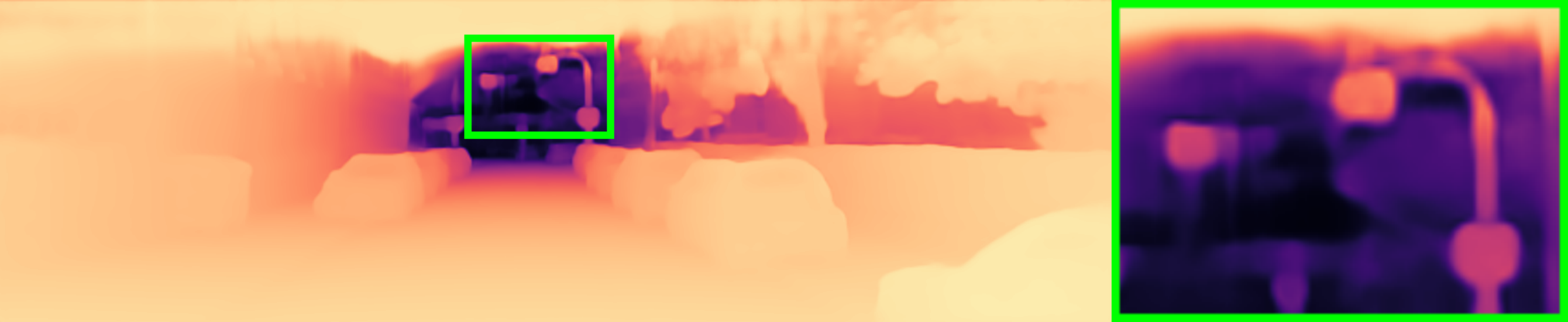}} \\ 

        \bottomrule[0.3pt]
        
        \toprule[0.3pt]
            \multirow{5}{*}{\rotatebox{90}{\parbox[c]{0.37\linewidth}{\centering \normalsize Res-KD}}} 
            &  {SSIM}\,&  
                \raisebox{-0.4\height}{\includegraphics[width=0.3\textwidth, height=0.07\textheight]{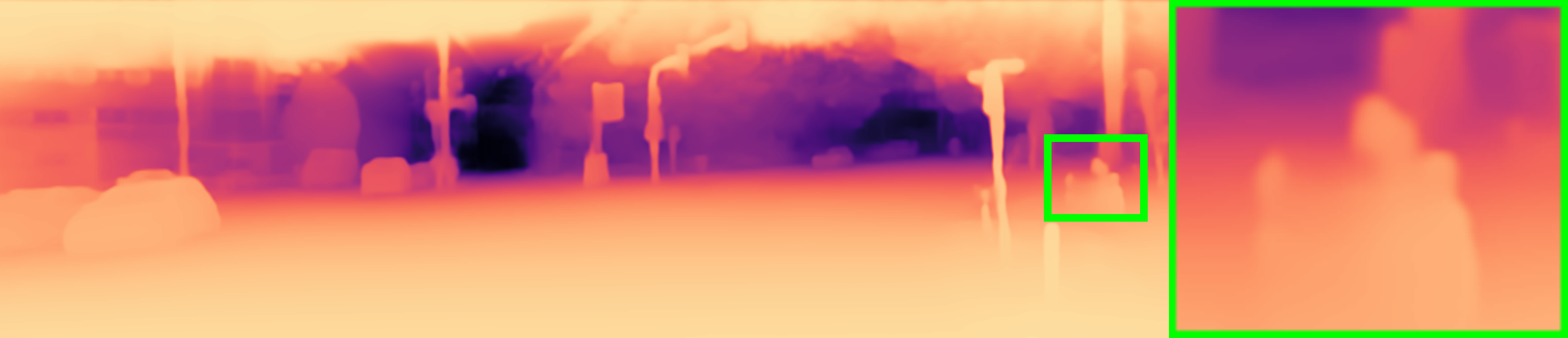}}& \vspace{0.5pt}
                \raisebox{-0.4\height}{\includegraphics[width=0.3\textwidth, height=0.07\textheight]{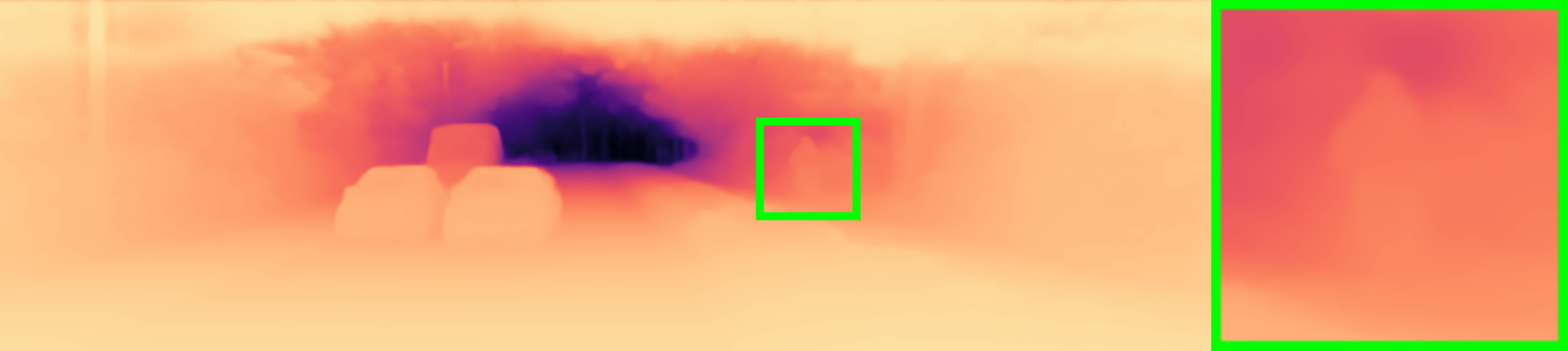}}& \vspace{0.5pt} 
                \raisebox{-0.4\height}{\includegraphics[width=0.3\textwidth, height=0.07\textheight]{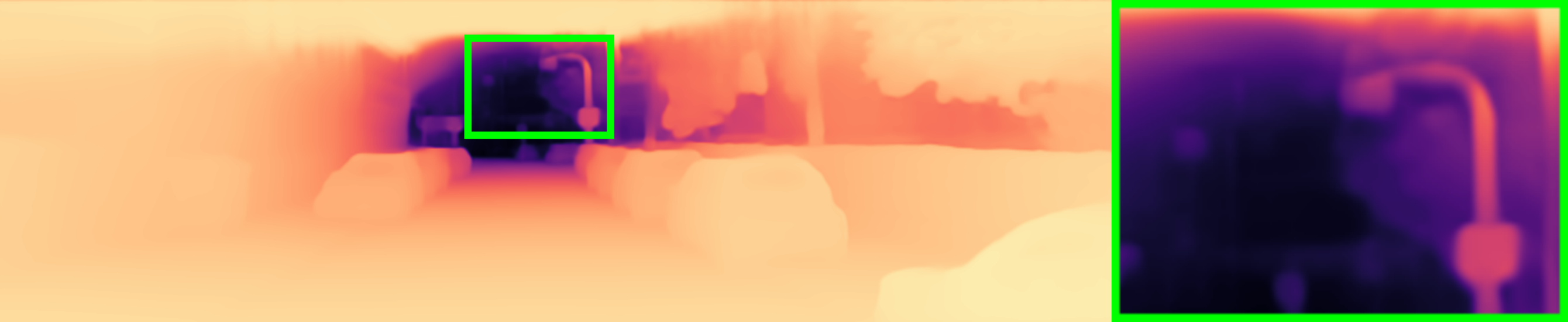}} \\ 
            &  {MSE}\,&  
                \raisebox{-0.4\height}{\includegraphics[width=0.3\textwidth, height=0.07\textheight]{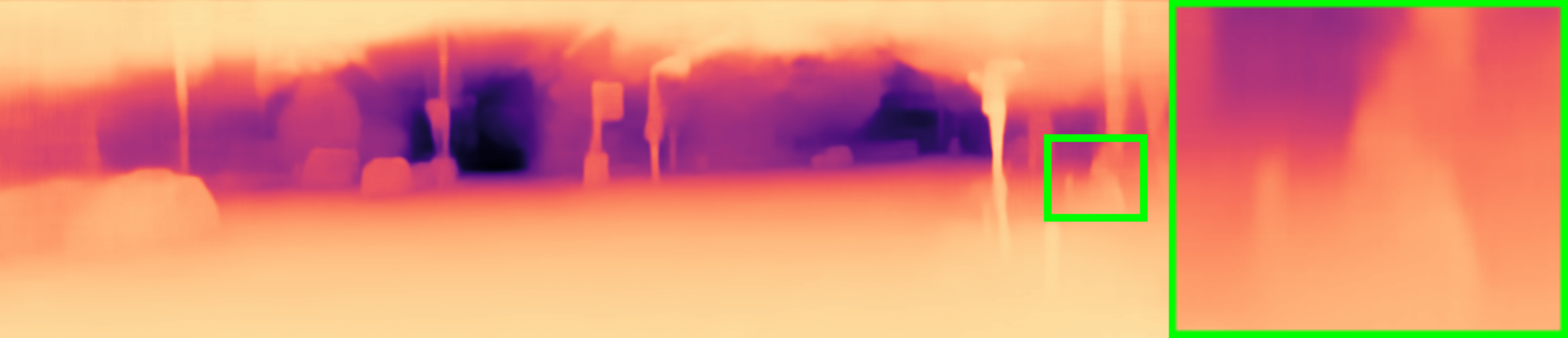}}& \vspace{0.5pt}
                \raisebox{-0.4\height}{\includegraphics[width=0.3\textwidth, height=0.07\textheight]{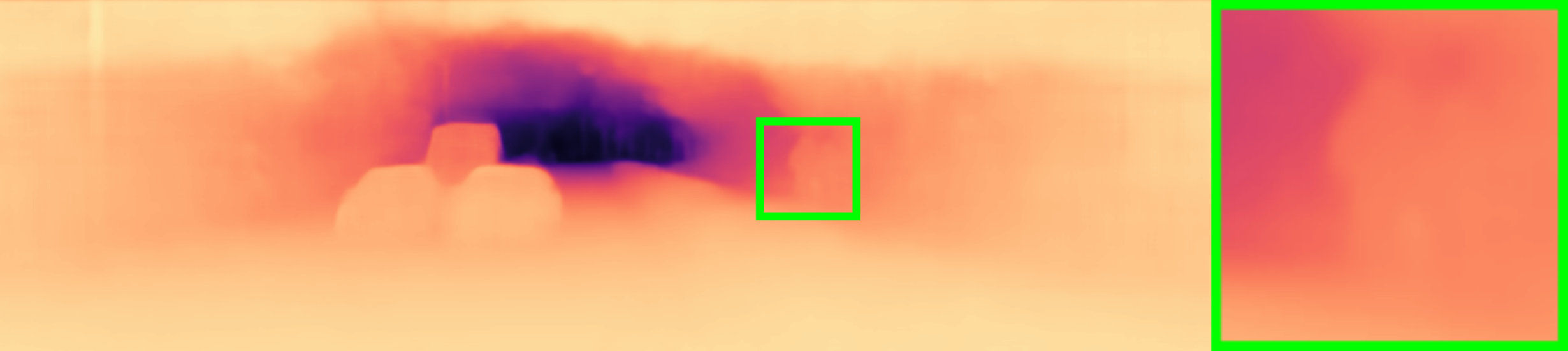}}& \vspace{0.5pt} 
                \raisebox{-0.4\height}{\includegraphics[width=0.3\textwidth, height=0.07\textheight]{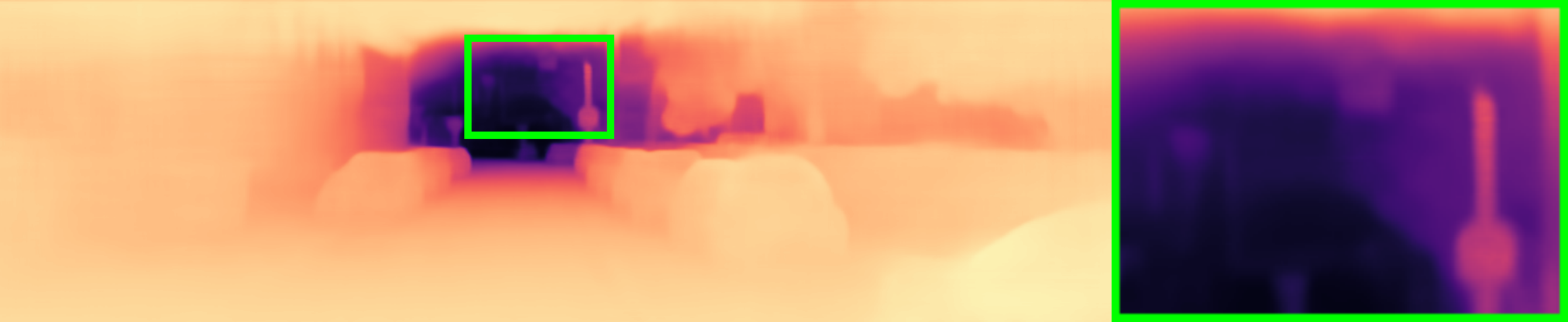}} \\ 
            &  {SI}\,&  
                \raisebox{-0.4\height}{\includegraphics[width=0.3\textwidth, height=0.07\textheight]{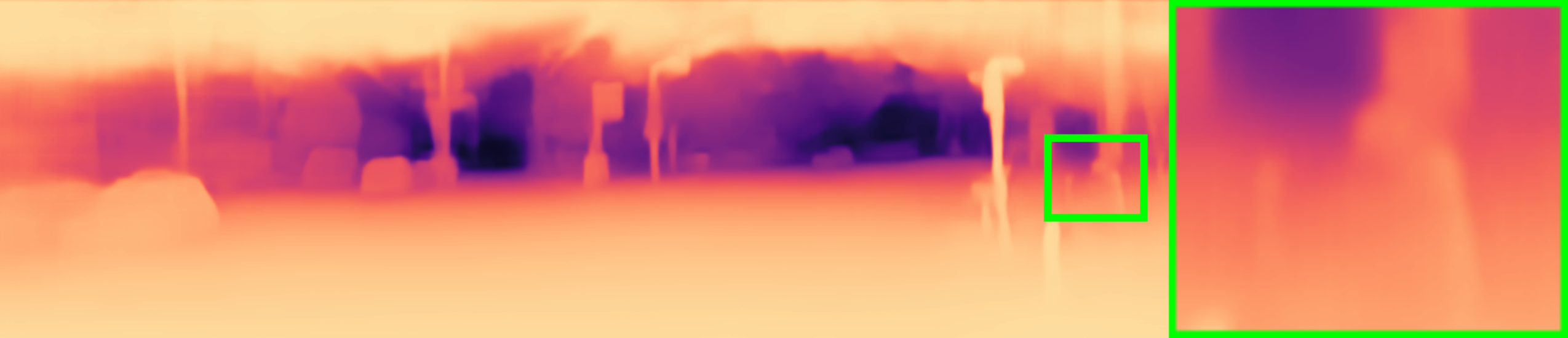}}& \vspace{0.5pt} 
                \raisebox{-0.4\height}{\includegraphics[width=0.3\textwidth, height=0.07\textheight]{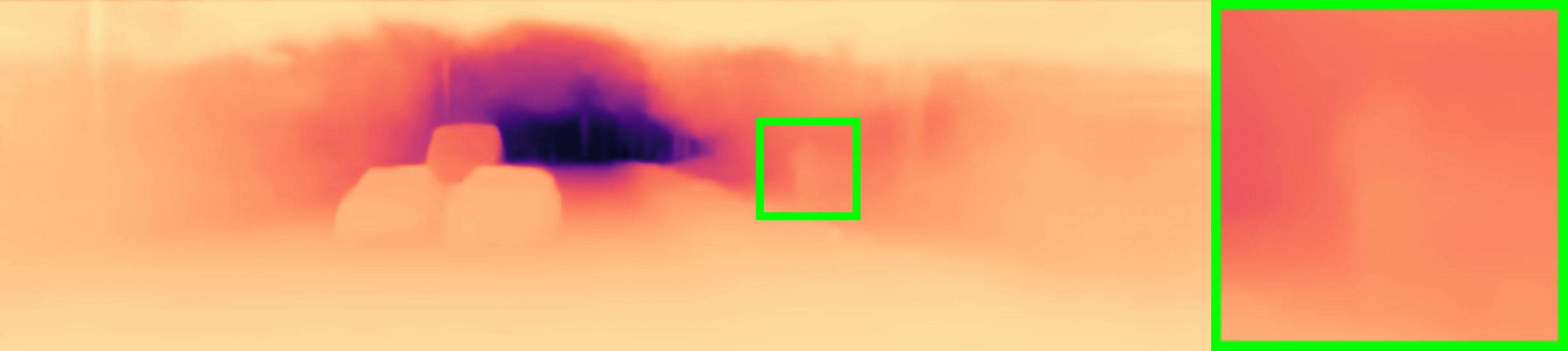}}& \vspace{0.5pt} 
                \raisebox{-0.4\height}{\includegraphics[width=0.3\textwidth, height=0.07\textheight]{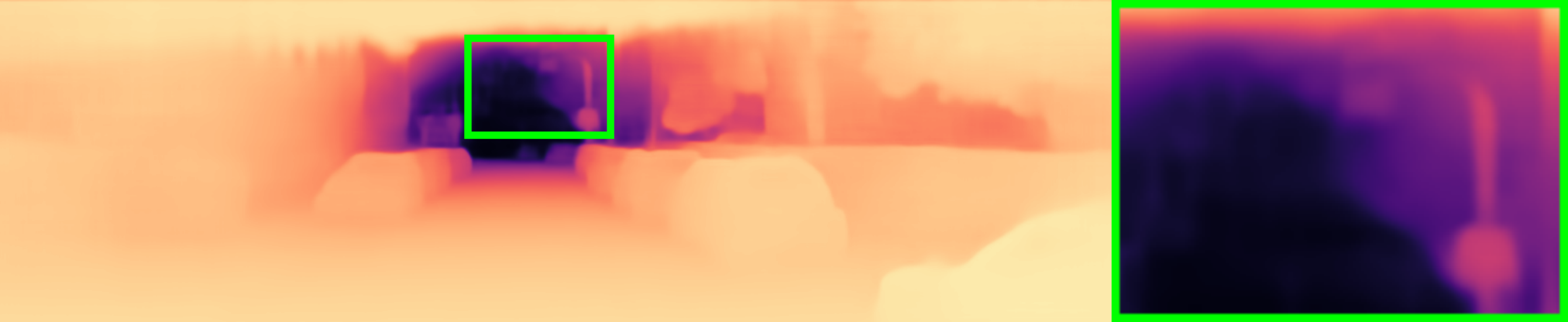}} \\ 
            &  {SSIM, SI}\,&  
                \raisebox{-0.4\height}{\includegraphics[width=0.3\textwidth, height=0.07\textheight]{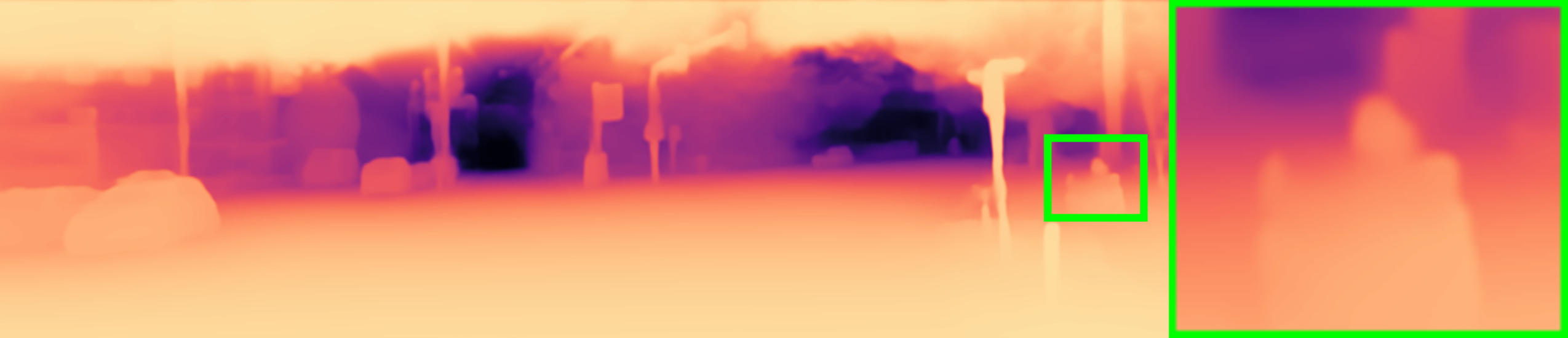}}&  \vspace{0.5pt} 
                \raisebox{-0.4\height}{\includegraphics[width=0.3\textwidth, height=0.07\textheight]{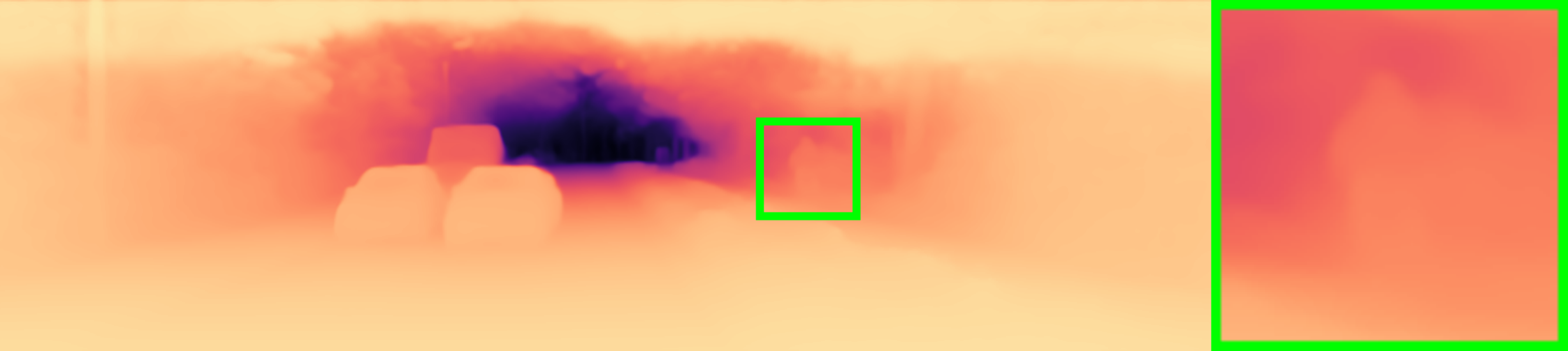}}& \vspace{0.5pt} 
                \raisebox{-0.4\height}{\includegraphics[width=0.3\textwidth, height=0.07\textheight]{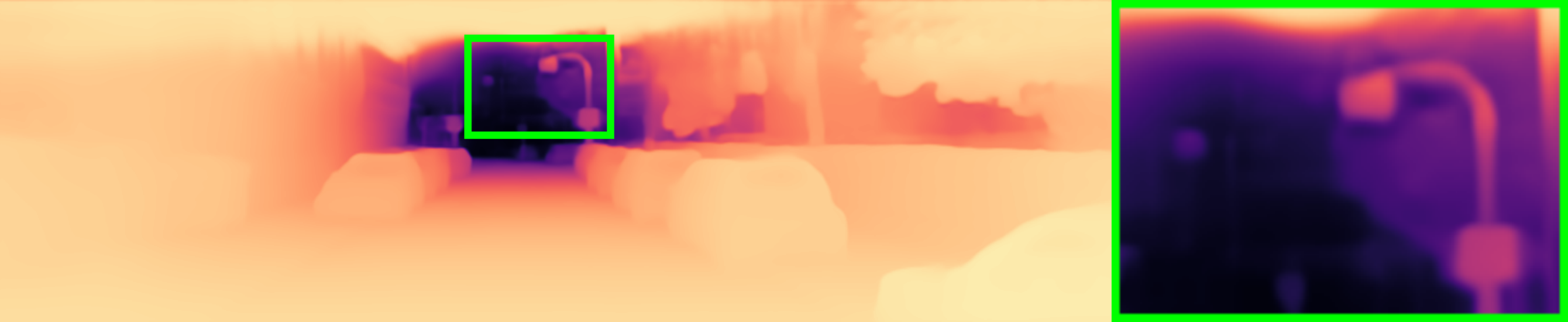}} \\ 
            &  {SSIM, MSE}\,&  
                \raisebox{-0.4\height}{\includegraphics[width=0.3\textwidth, height=0.07\textheight]{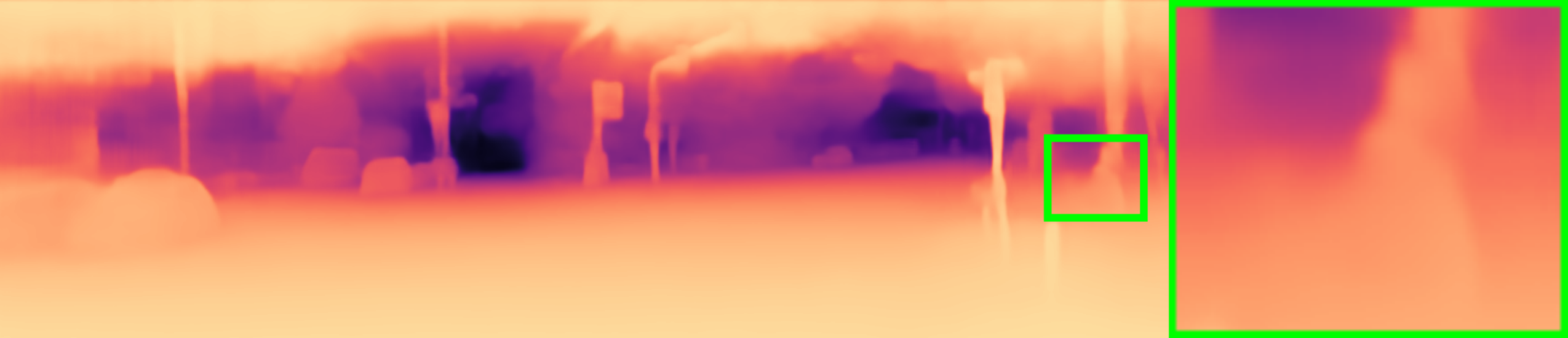}}& \vspace{0.5pt} 
                \raisebox{-0.4\height}{\includegraphics[width=0.3\textwidth, height=0.07\textheight]{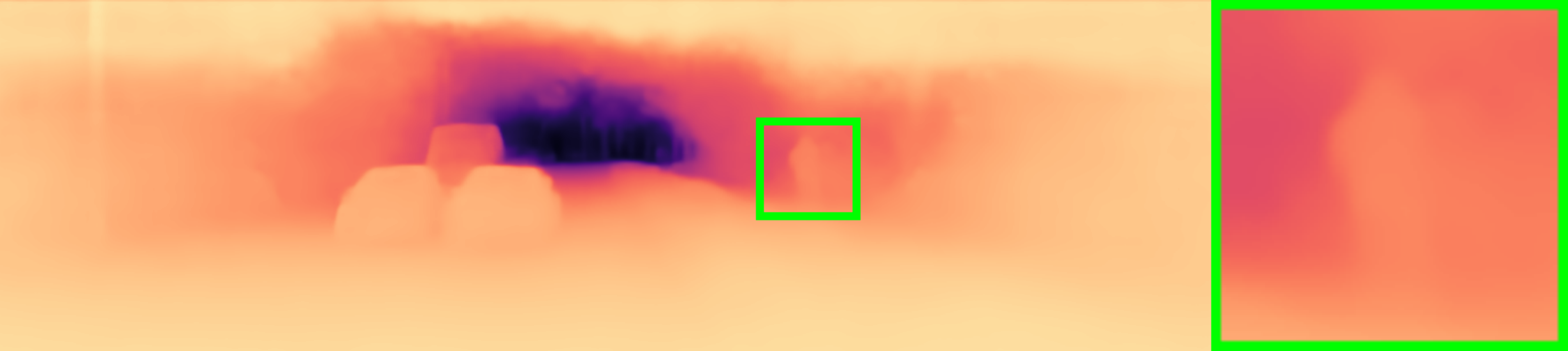}}& \vspace{0.5pt} 
                \raisebox{-0.4\height}{\includegraphics[width=0.3\textwidth, height=0.07\textheight]{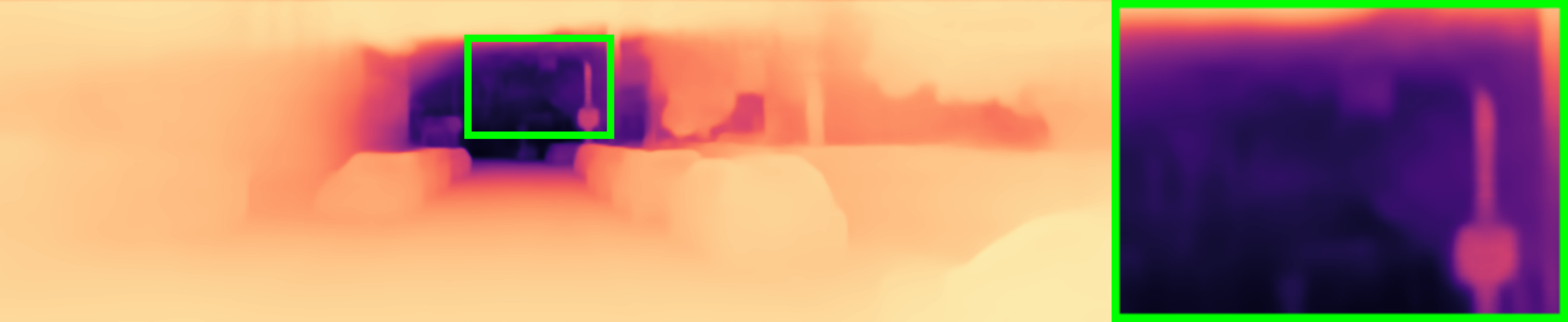}} \\ 
        \bottomrule[0.3pt]
        
        \toprule[0.3pt]
            \multirow{3}{*}{\rotatebox{90}{\parbox[c]{0.2\linewidth}{\centering \normalsize \bf{TIE-KD}}}} 
            & $L_{DPM}$\,&  
                \raisebox{-0.4\height}{\includegraphics[width=0.3\textwidth, height=0.07\textheight]{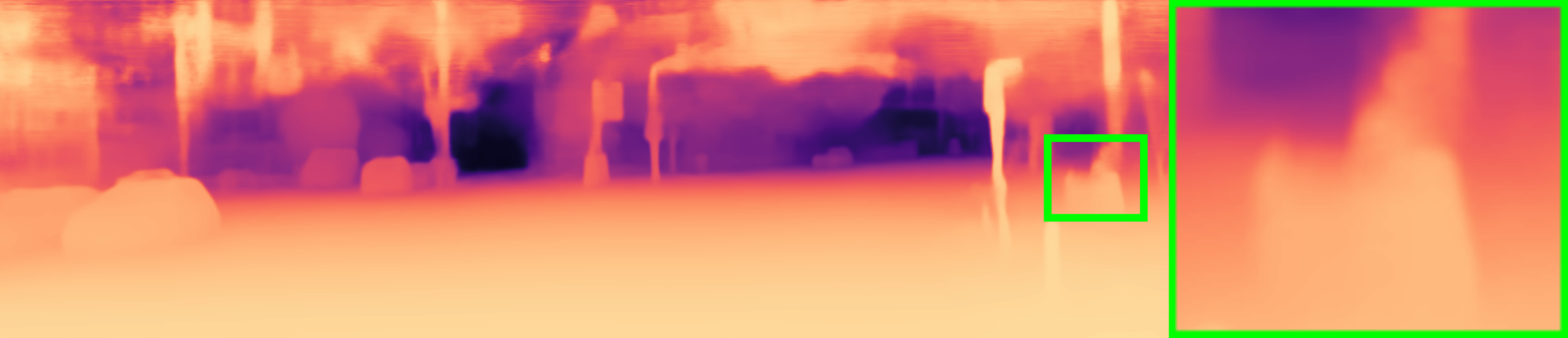}}& \vspace{0.5pt} 
                \raisebox{-0.4\height}{\includegraphics[width=0.3\textwidth, height=0.07\textheight]{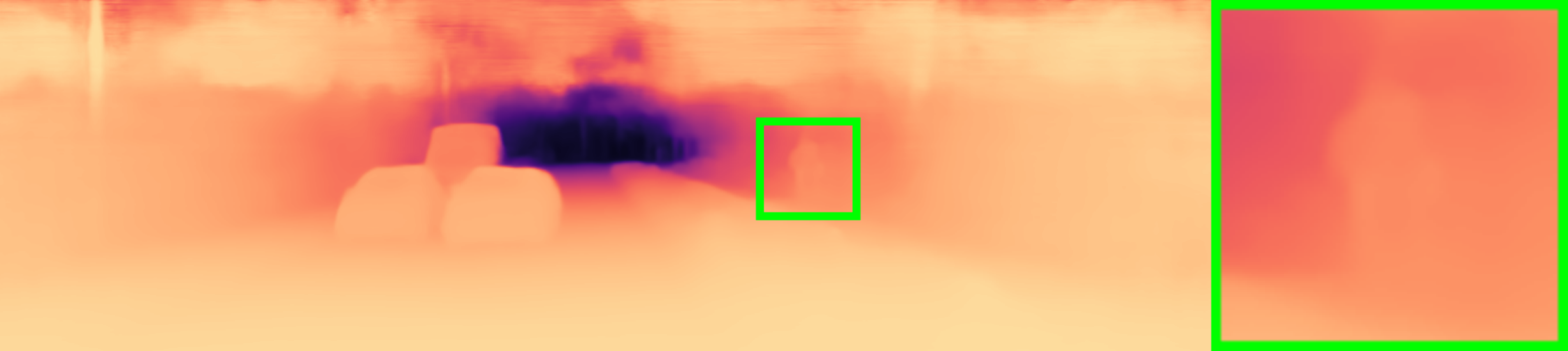}}& \vspace{0.5pt} 
                \raisebox{-0.4\height}{\includegraphics[width=0.3\textwidth, height=0.07\textheight]{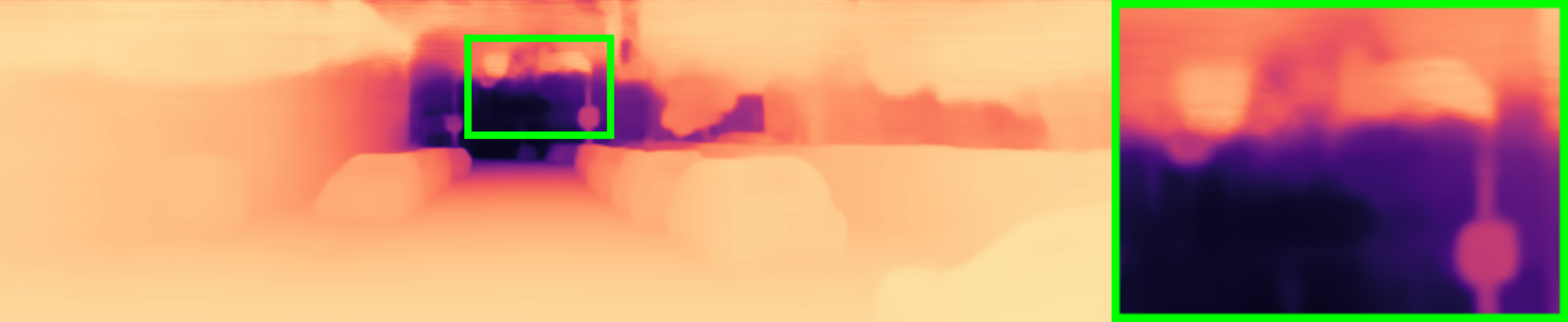}} \\ 
            & $L_{depth}$\,&  
                \raisebox{-0.4\height}{\includegraphics[width=0.3\textwidth, height=0.07\textheight]{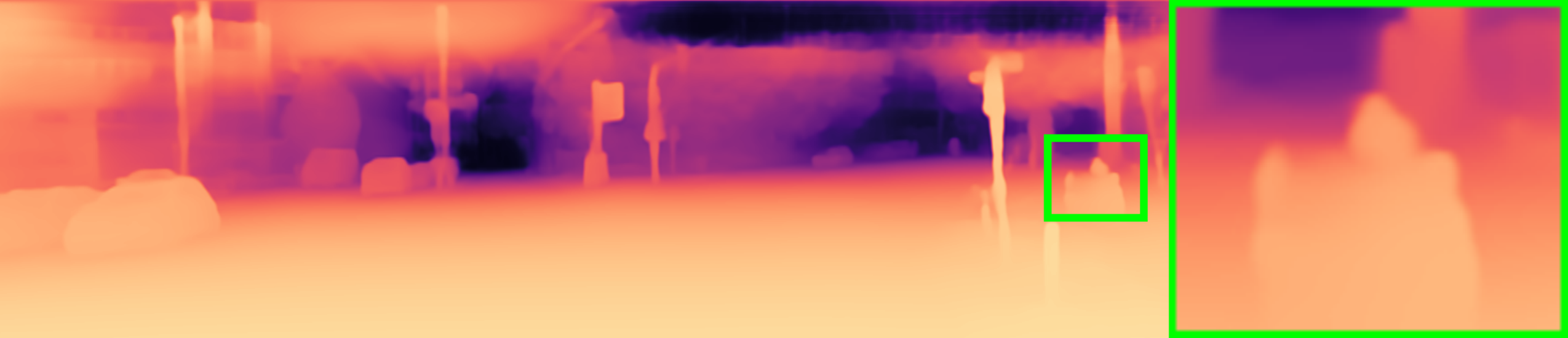}}& \vspace{0.5pt} 
                \raisebox{-0.4\height}{\includegraphics[width=0.3\textwidth, height=0.07\textheight]{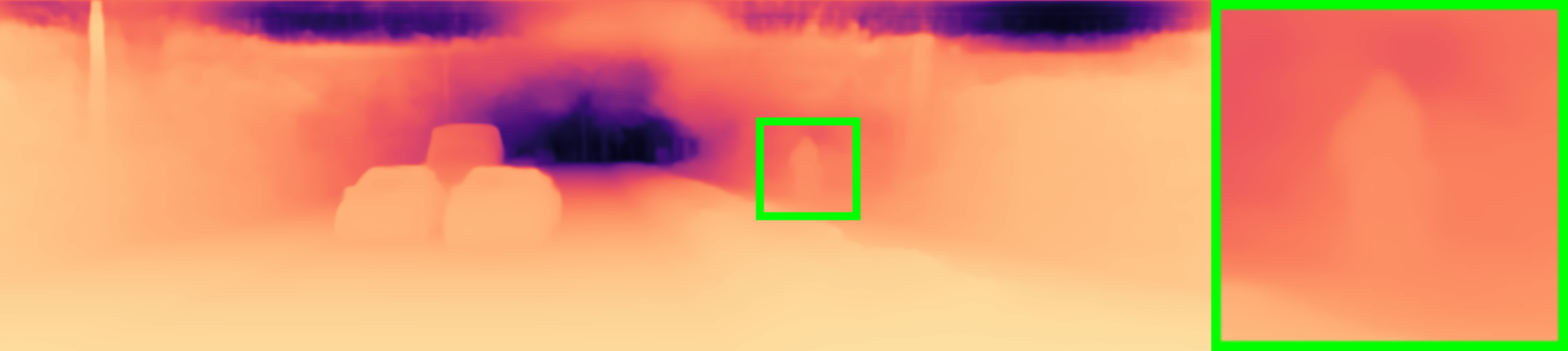}}& \vspace{0.5pt} 
                \raisebox{-0.4\height}{\includegraphics[width=0.3\textwidth, height=0.07\textheight]{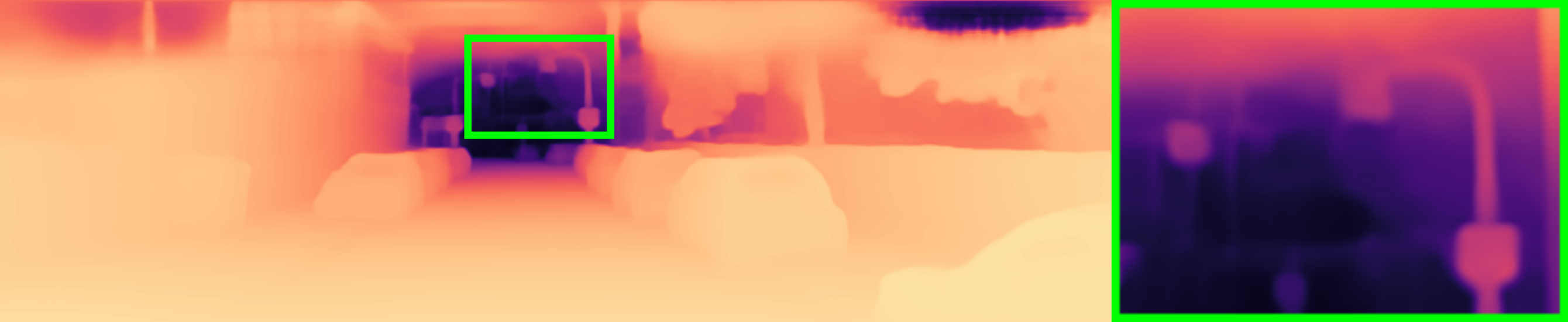}} \\ 
            & $L_{DPM, depth}$\,&  
                \raisebox{-0.4\height}{\includegraphics[width=0.3\textwidth, height=0.07\textheight]{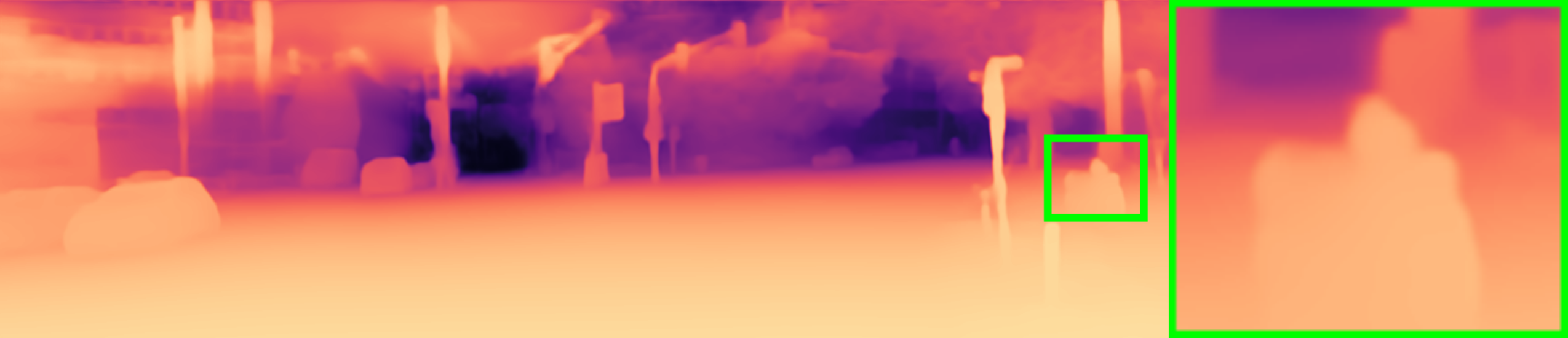}}& \vspace{0.5pt} 
                \raisebox{-0.4\height}{\includegraphics[width=0.3\textwidth, height=0.07\textheight]{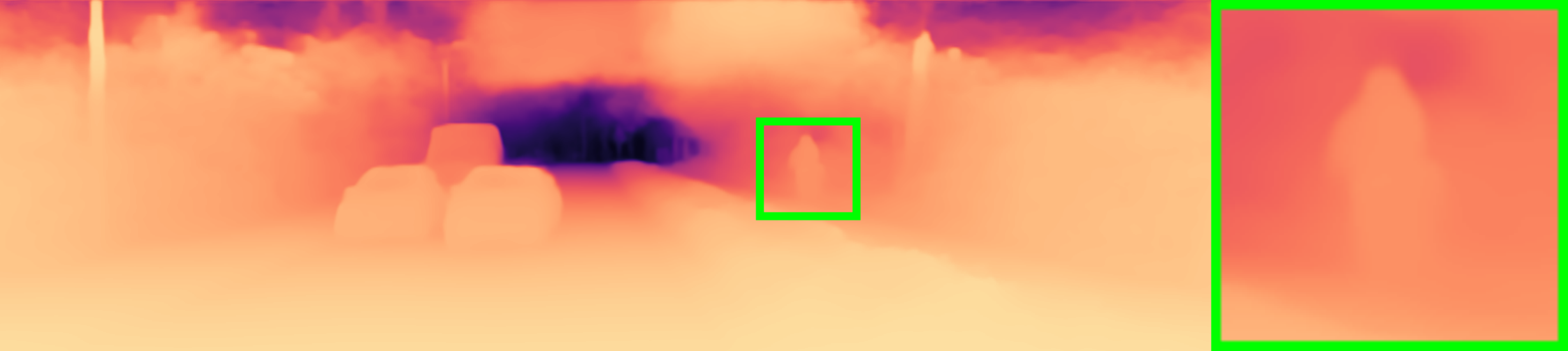}}& \vspace{0.5pt} 
                \raisebox{-0.4\height}{\includegraphics[width=0.3\textwidth, height=0.07\textheight]{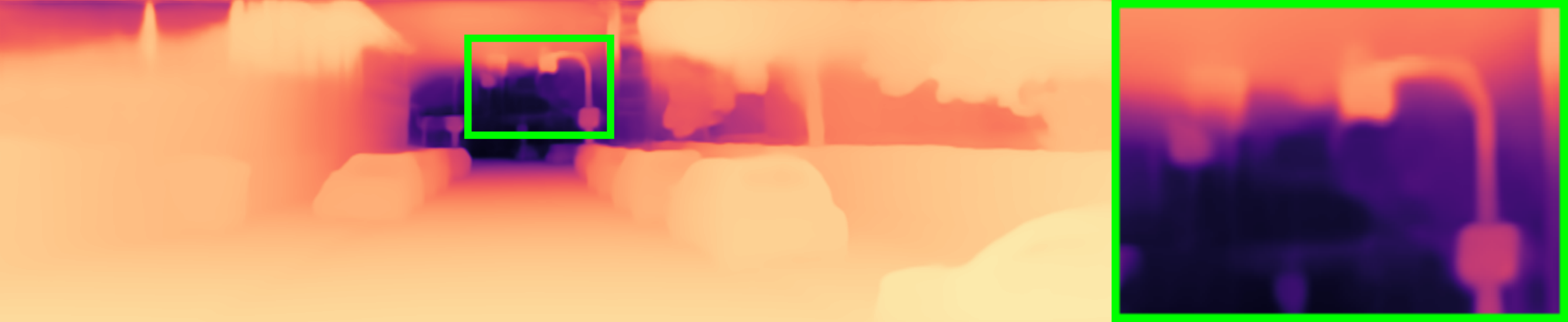}} \\ 

    \end{tabular}
    }
    \caption{Visual comparison of depth maps produced by different models for three distinct scenes, with a focus on detail variations within areas marked by green boxes.
    The top row presents the input images.
    The second row illustrates the depth maps generated by the teacher model, here adabins~\cite{bhat2021adabins}.
    The third row depicts the baseline model's output.
    Subsequent rows display the results of the student models trained using various response-based knowledge distillation methods (Res-KD) with different loss function combinations, and the bottom rows show the depth maps from students trained using our proposed TIE-KD framework with different loss function configurations.}
    \label{fig:adabins_results}
\end{figure*}
\newpage

\subsection{Teacher model: BTS~\cite{lee2019big}}
\begin{figure*}[hb!]
    \centering
    \resizebox{0.84\textwidth}{!}{%
    \renewcommand{\arraystretch}{1}
    \setlength{\tabcolsep}{0pt}
    \begin{tabular}{c@{\hspace{0.5em}}@{\hspace{0.5em}}r@{\hspace{0.5em}}c@{\hspace{0.5em}}c@{\hspace{0.5em}}c}
            \multicolumn{2}{c}{ \normalsize Input\,}  & 
                \raisebox{-0.4\height}{\includegraphics[width=0.3\textwidth, height=0.07\textheight]{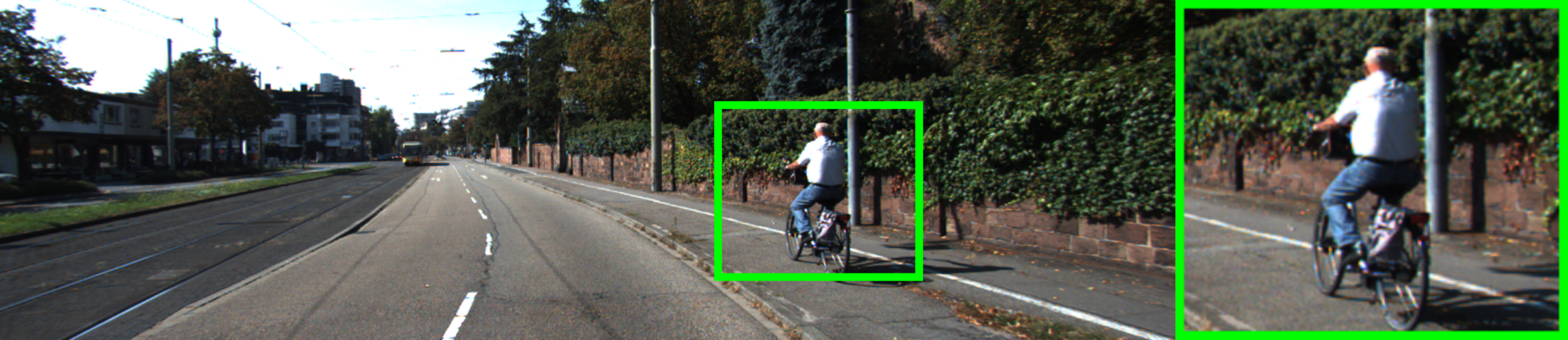}}& \vspace{0.5pt}
                \raisebox{-0.4\height}{\includegraphics[width=0.3\textwidth, height=0.07\textheight]{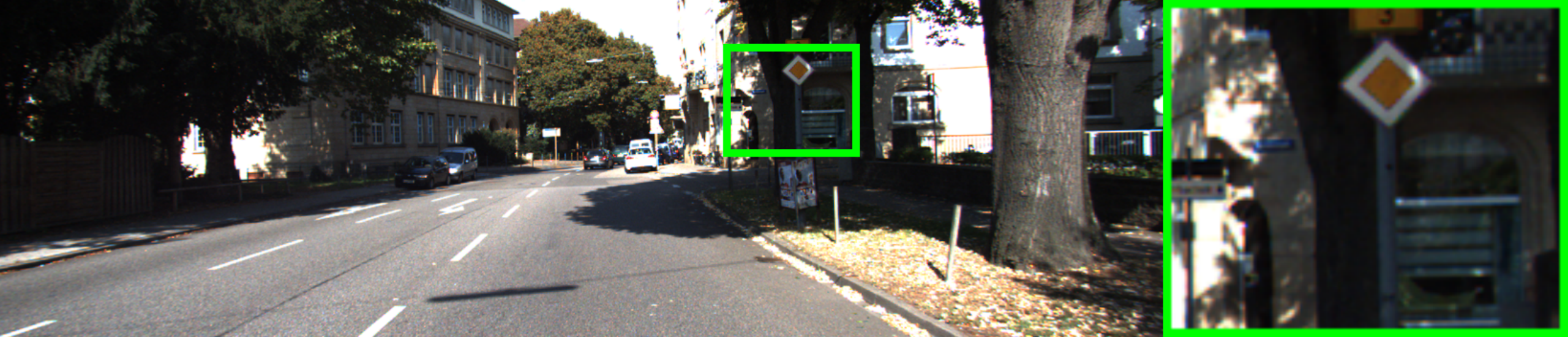}}& \vspace{0.5pt}
                \raisebox{-0.4\height}{\includegraphics[width=0.3\textwidth, height=0.07\textheight]{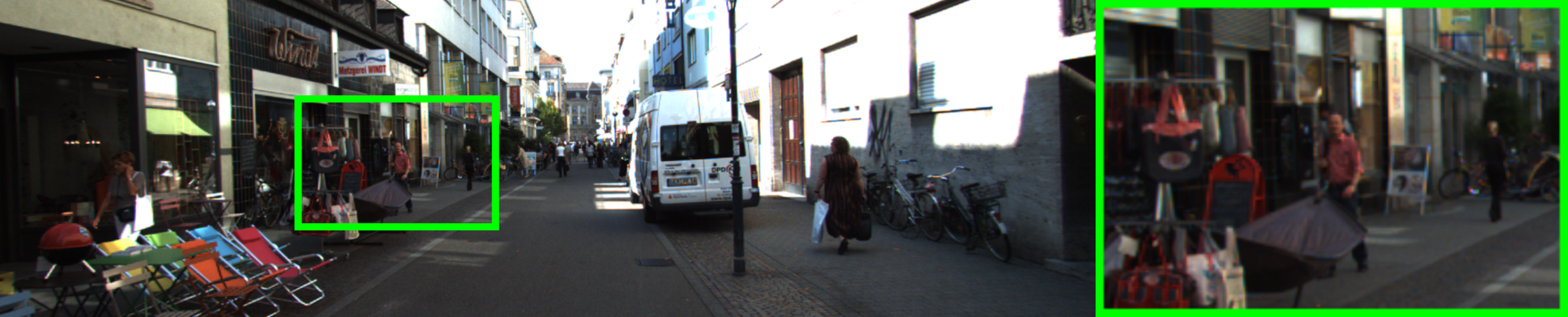}} \\ 
            \multicolumn{2}{c}{ \normalsize Baseline\,}& 
                \raisebox{-0.4\height}{\includegraphics[width=0.3\textwidth, height=0.07\textheight]{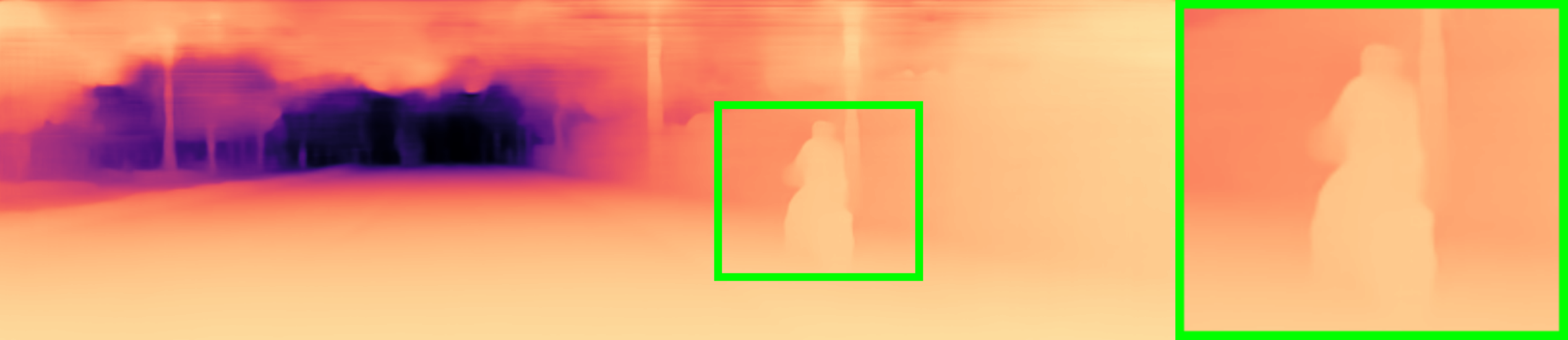}}& \vspace{0.5pt} 
                \raisebox{-0.4\height}{\includegraphics[width=0.3\textwidth, height=0.07\textheight]{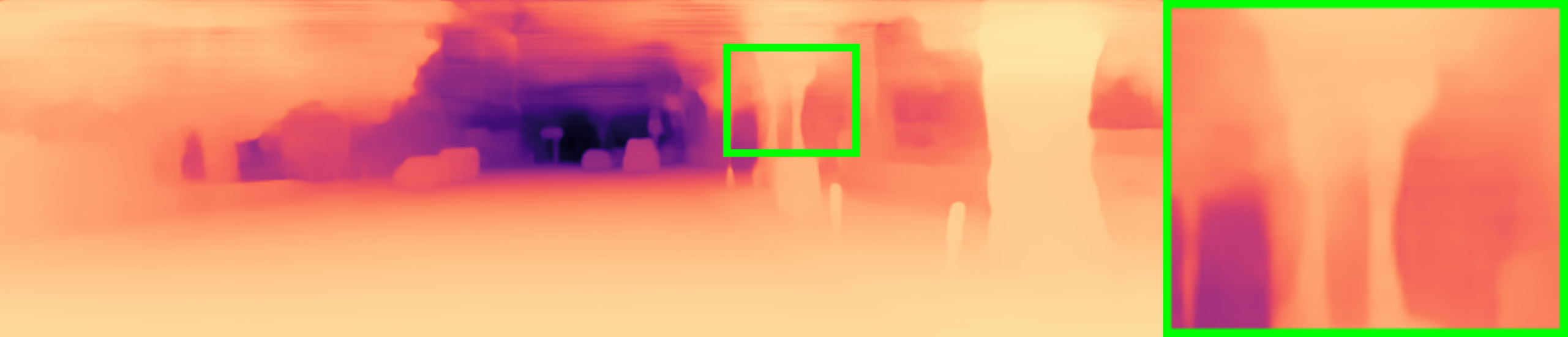}}& \vspace{0.5pt} 
                \raisebox{-0.4\height}{\includegraphics[width=0.3\textwidth, height=0.07\textheight]{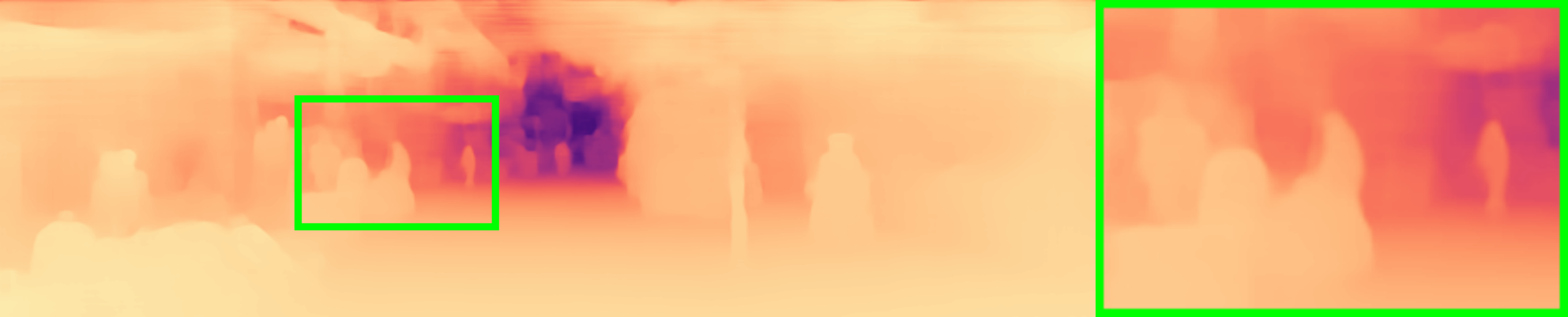}} \\ 
            \multicolumn{2}{c}{ \normalsize Teacher\,}& 
                \raisebox{-0.4\height}{\includegraphics[width=0.3\textwidth, height=0.07\textheight]{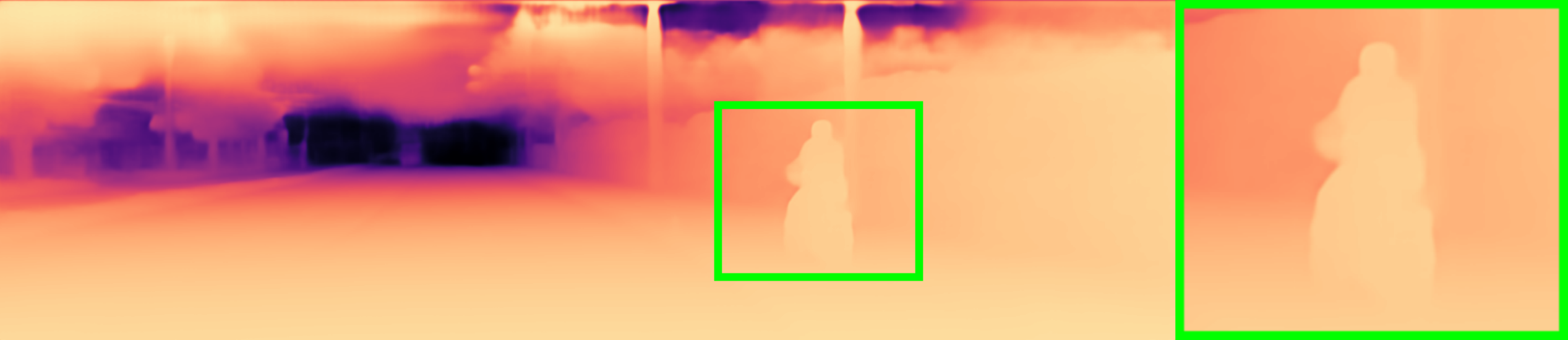}}& \vspace{0.5pt} 
                \raisebox{-0.4\height}{\includegraphics[width=0.3\textwidth, height=0.07\textheight]{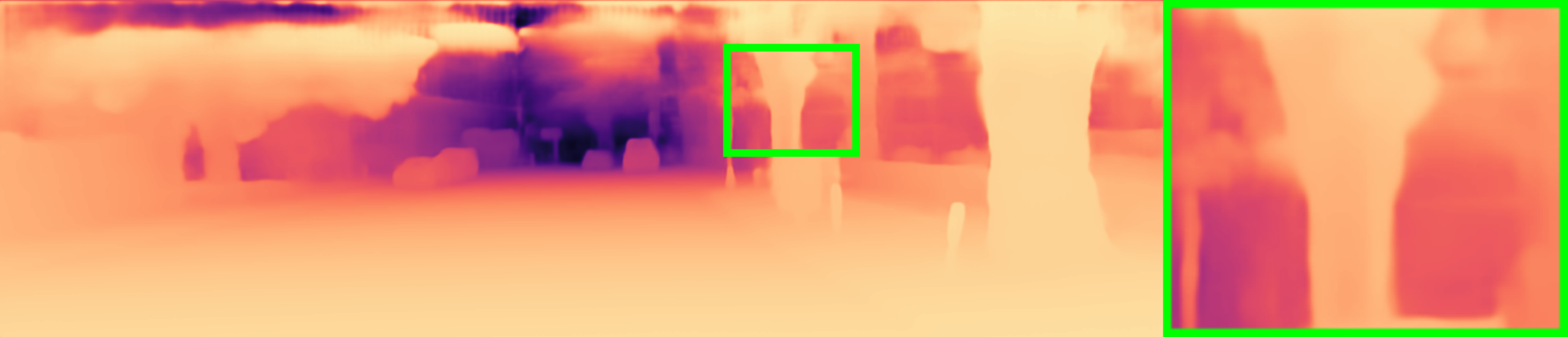}}& \vspace{0.5pt} 
                \raisebox{-0.4\height}{\includegraphics[width=0.3\textwidth, height=0.07\textheight]{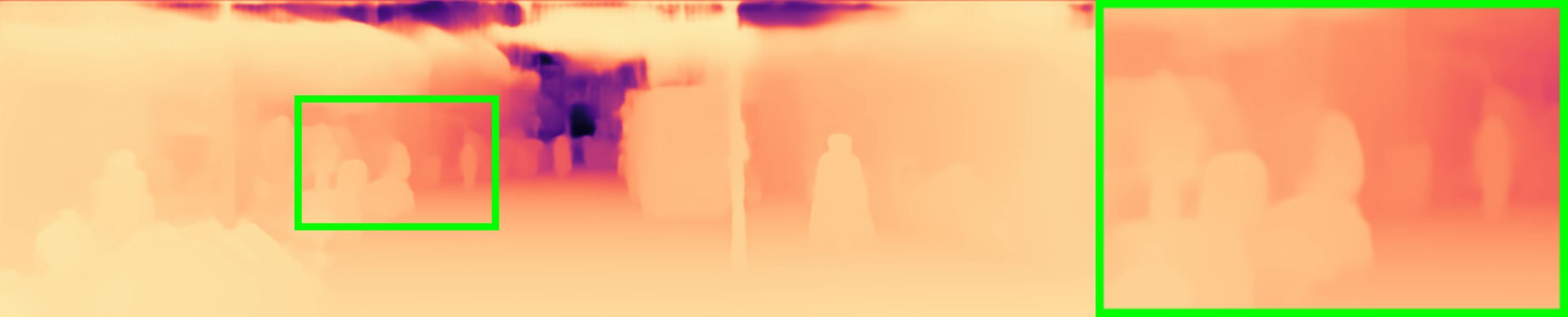}} \\ 

        \bottomrule[0.3pt]
        
        \toprule[0.3pt]
            \multirow{5}{*}{\rotatebox{90}{\parbox[c]{0.37\linewidth}{\centering \normalsize Res-KD}}} 
            &  {SSIM}\,&  
                \raisebox{-0.4\height}{\includegraphics[width=0.3\textwidth, height=0.07\textheight]{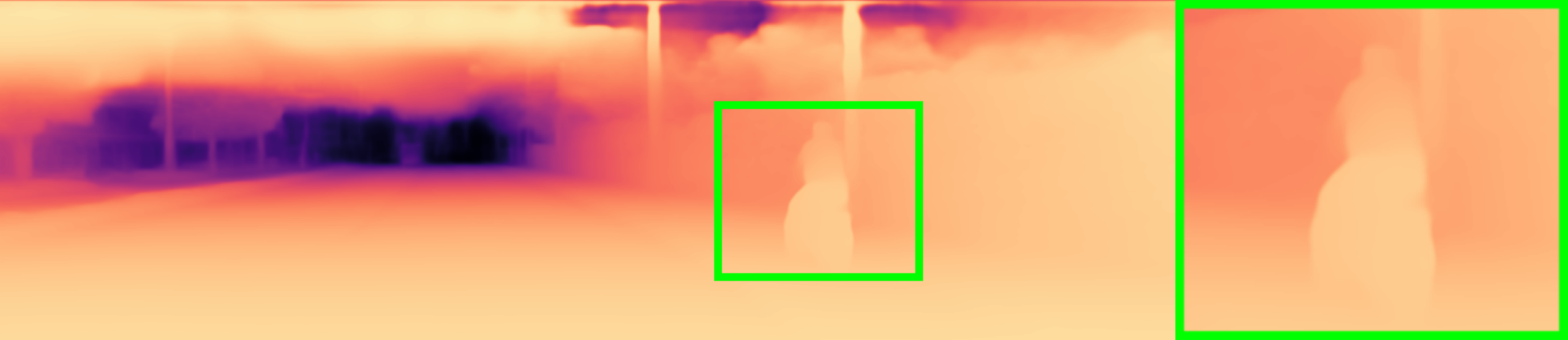}}& \vspace{0.5pt}
                \raisebox{-0.4\height}{\includegraphics[width=0.3\textwidth, height=0.07\textheight]{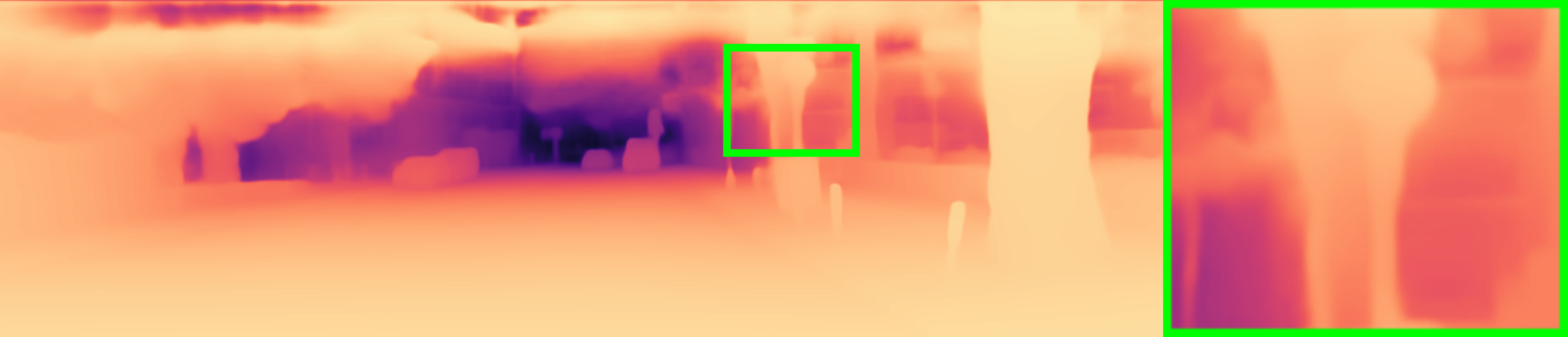}}& \vspace{0.5pt} 
                \raisebox{-0.4\height}{\includegraphics[width=0.3\textwidth, height=0.07\textheight]{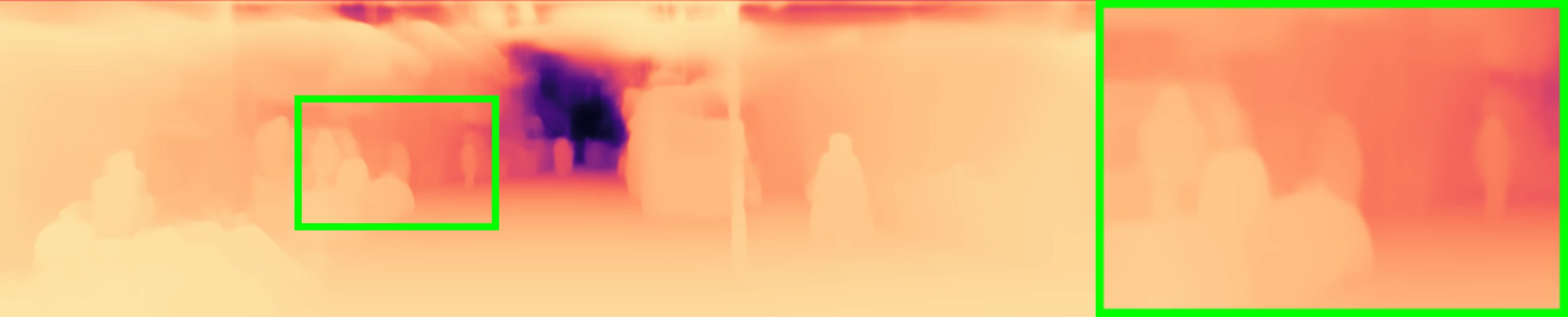}} \\ 
            &  {MSE}\,&  
                \raisebox{-0.4\height}{\includegraphics[width=0.3\textwidth, height=0.07\textheight]{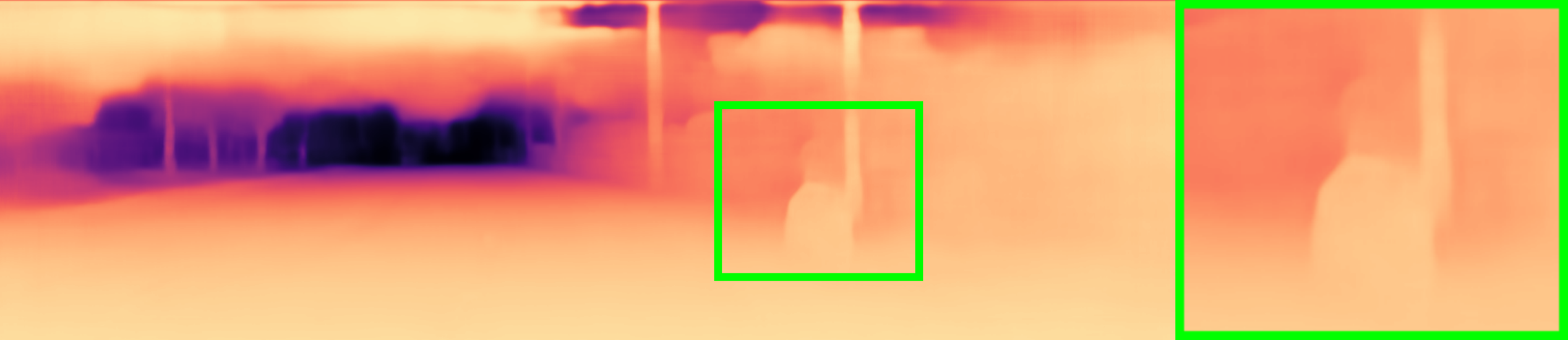}}& \vspace{0.5pt}
                \raisebox{-0.4\height}{\includegraphics[width=0.3\textwidth, height=0.07\textheight]{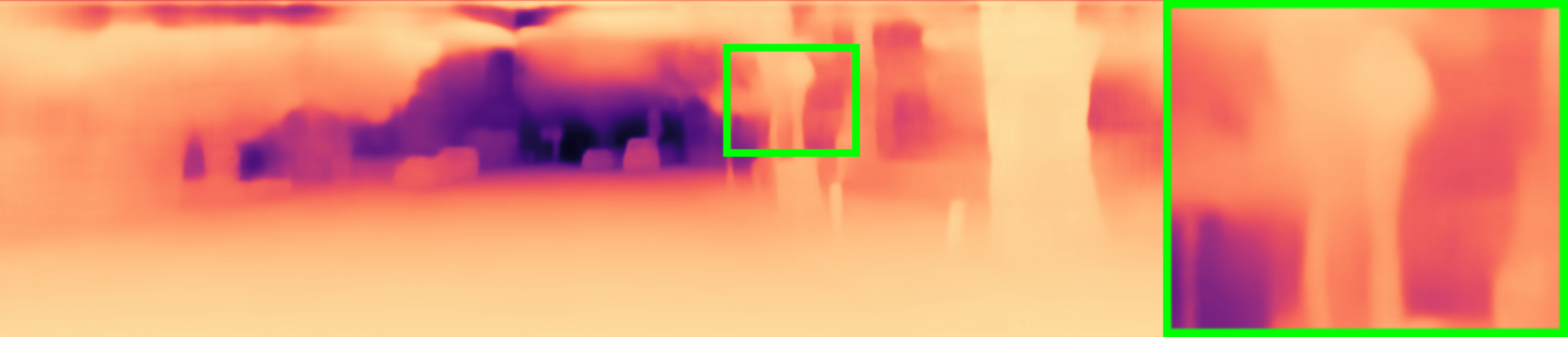}}& \vspace{0.5pt} 
                \raisebox{-0.4\height}{\includegraphics[width=0.3\textwidth, height=0.07\textheight]{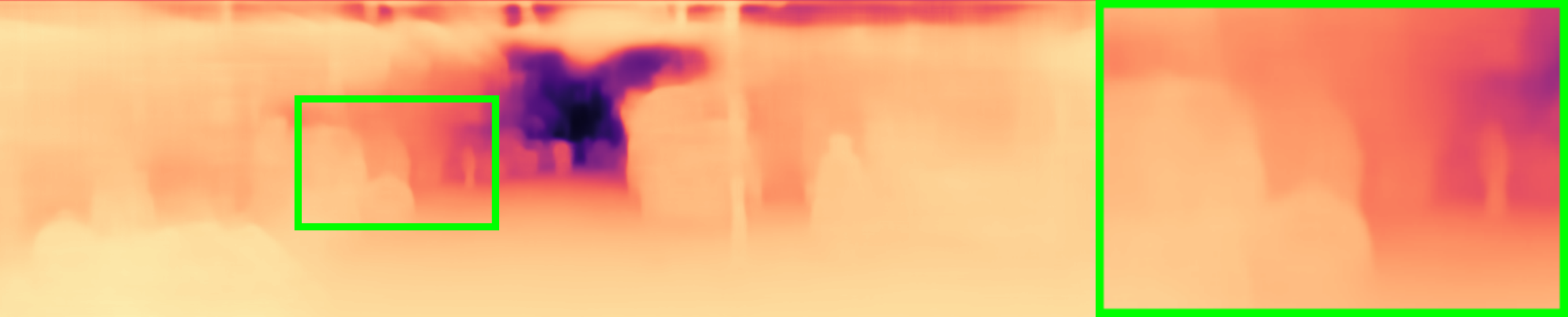}} \\ 
            &  {SI}\,&  
                \raisebox{-0.4\height}{\includegraphics[width=0.3\textwidth, height=0.07\textheight]{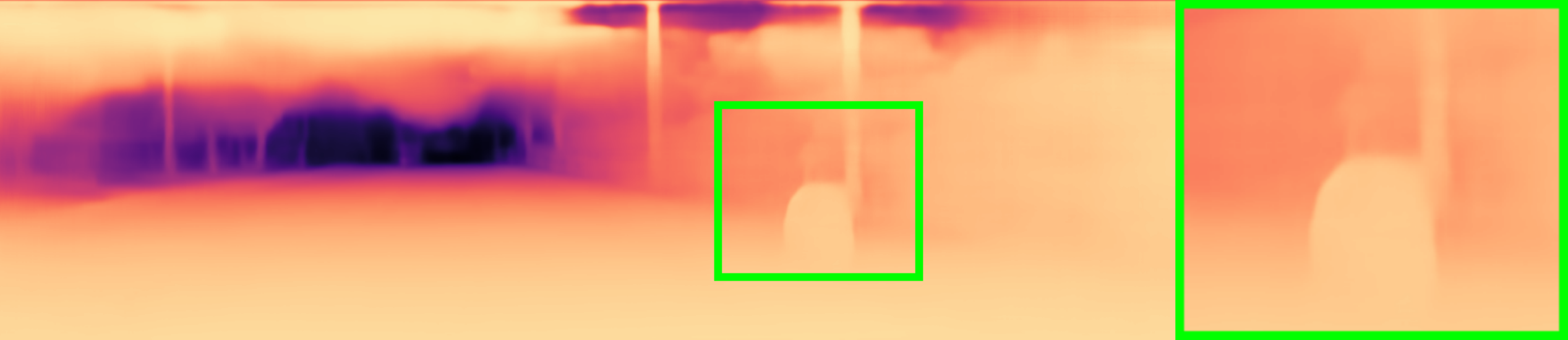}}& \vspace{0.5pt} 
                \raisebox{-0.4\height}{\includegraphics[width=0.3\textwidth, height=0.07\textheight]{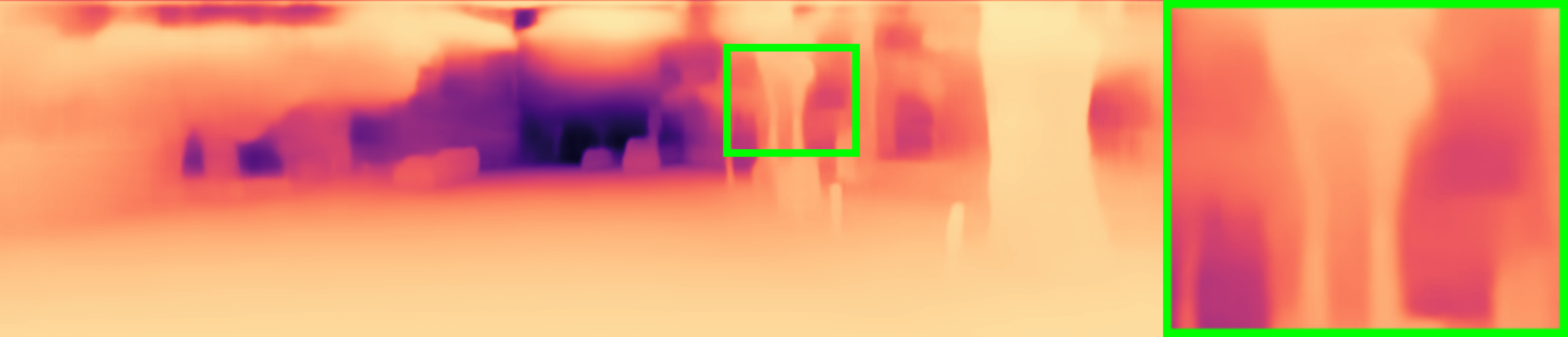}}& \vspace{0.5pt} 
                \raisebox{-0.4\height}{\includegraphics[width=0.3\textwidth, height=0.07\textheight]{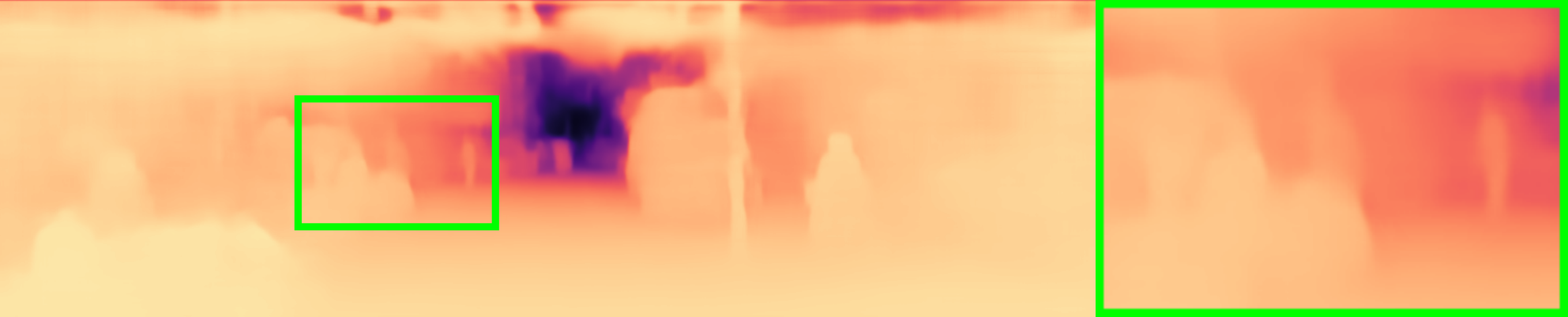}} \\ 
            &  {SSIM, SI}\,&  
                \raisebox{-0.4\height}{\includegraphics[width=0.3\textwidth, height=0.07\textheight]{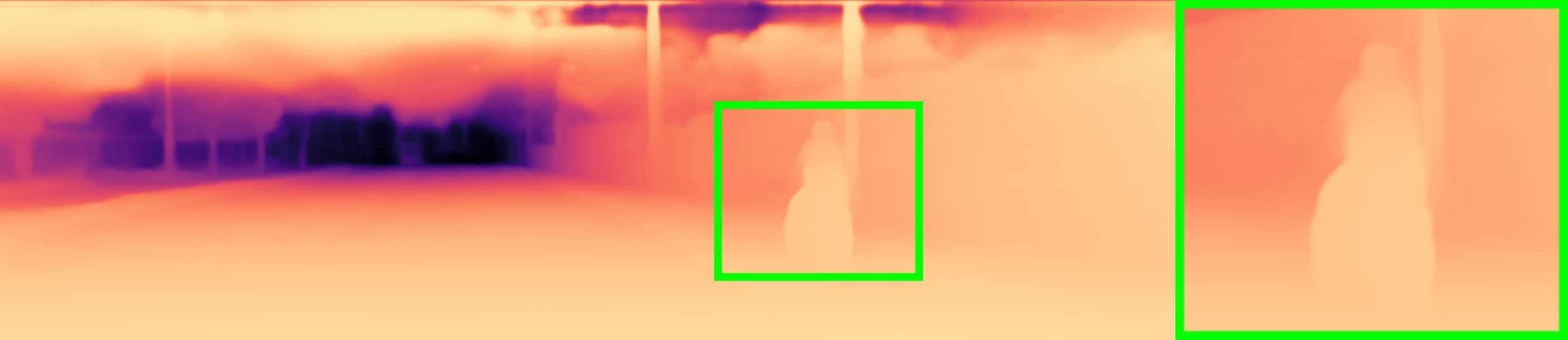}}&  \vspace{0.5pt} 
                \raisebox{-0.4\height}{\includegraphics[width=0.3\textwidth, height=0.07\textheight]{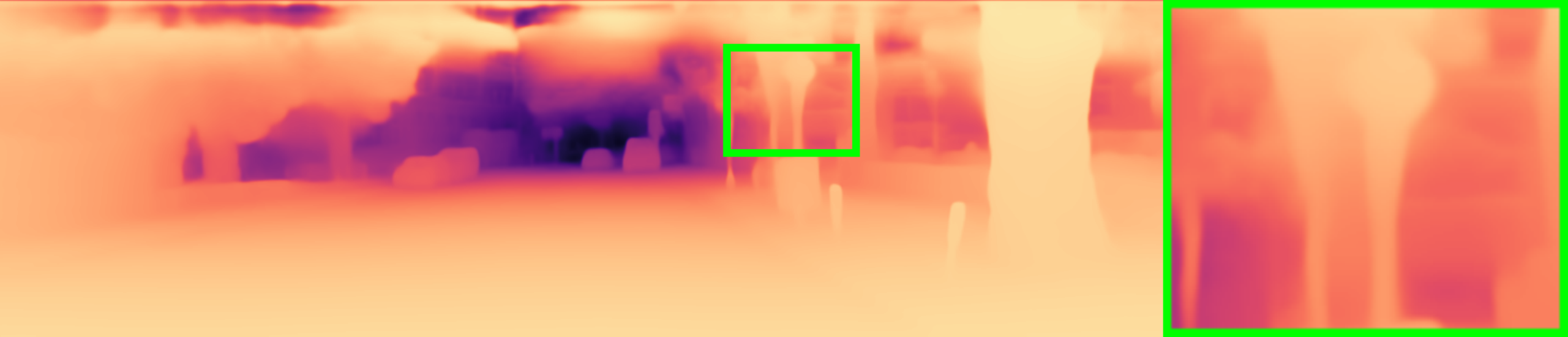}}& \vspace{0.5pt} 
                \raisebox{-0.4\height}{\includegraphics[width=0.3\textwidth, height=0.07\textheight]{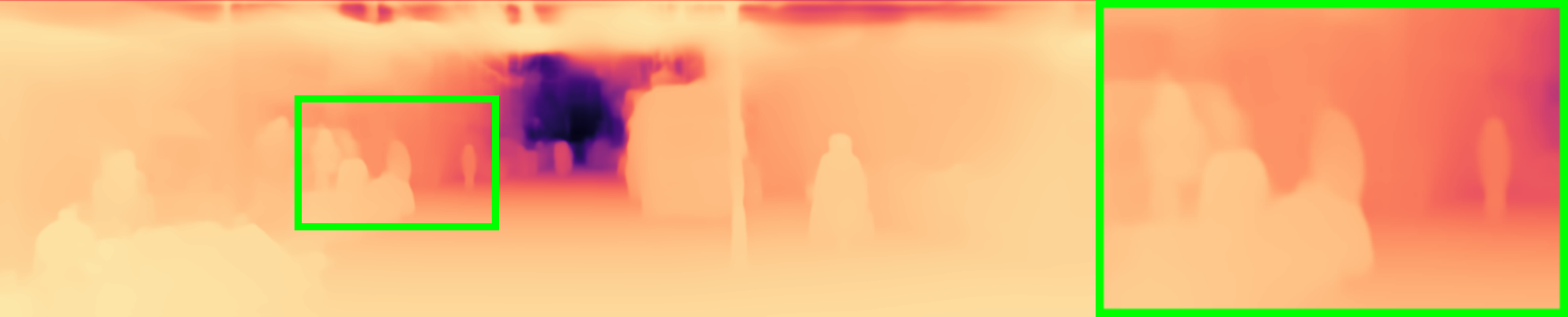}} \\ 
            &  {SSIM, MSE}\,&  
                \raisebox{-0.4\height}{\includegraphics[width=0.3\textwidth, height=0.07\textheight]{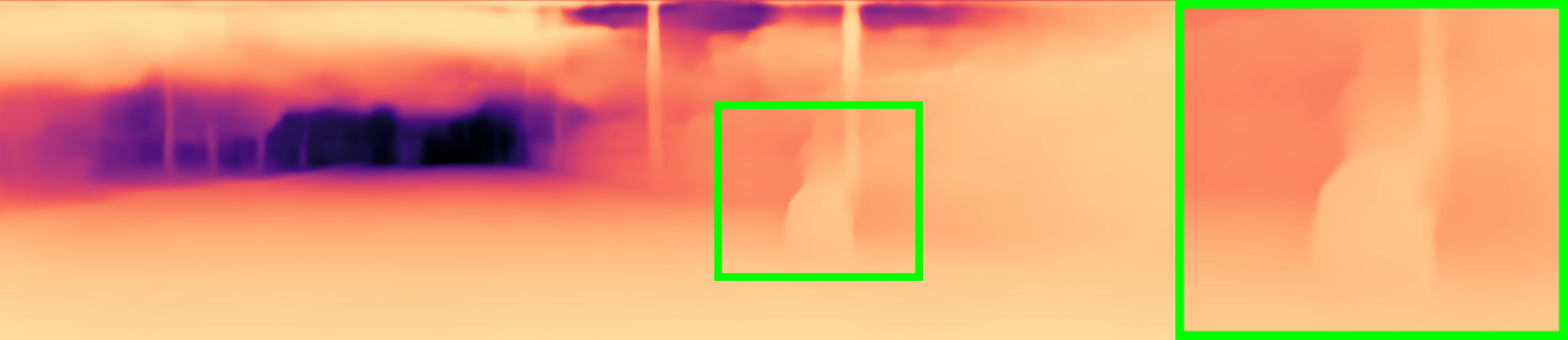}}& \vspace{0.5pt} 
                \raisebox{-0.4\height}{\includegraphics[width=0.3\textwidth, height=0.07\textheight]{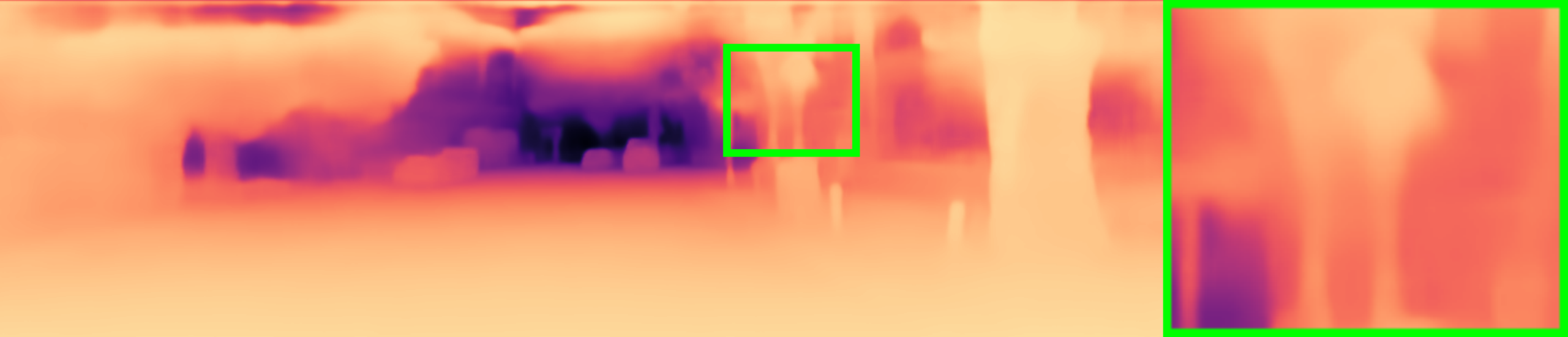}}& \vspace{0.5pt} 
                \raisebox{-0.4\height}{\includegraphics[width=0.3\textwidth, height=0.07\textheight]{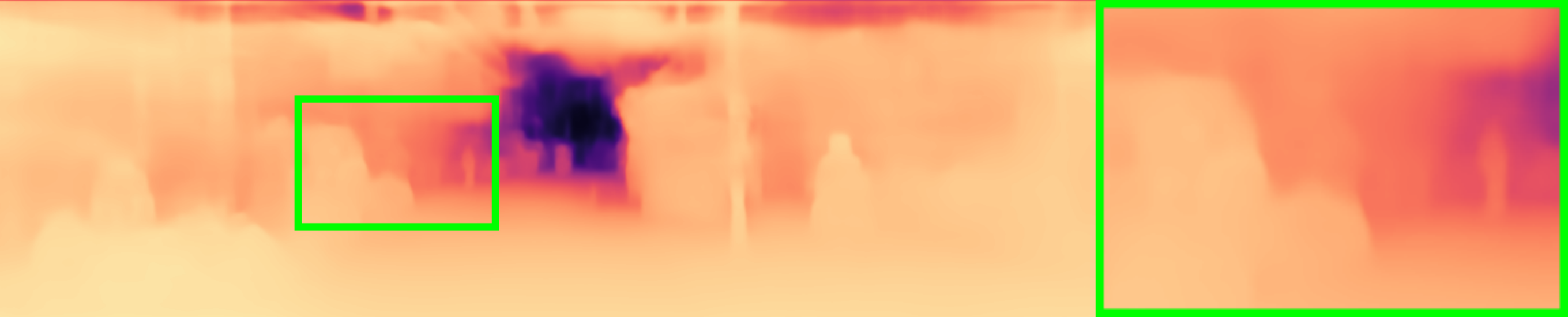}} \\ 
        \bottomrule[0.3pt]
        
        \toprule[0.3pt]
            \multirow{3}{*}{\rotatebox{90}{\parbox[c]{0.2\linewidth}{\centering \normalsize \bf{TIE-KD}}}} 
            & $L_{DPM}$\,&  
                \raisebox{-0.4\height}{\includegraphics[width=0.3\textwidth, height=0.07\textheight]{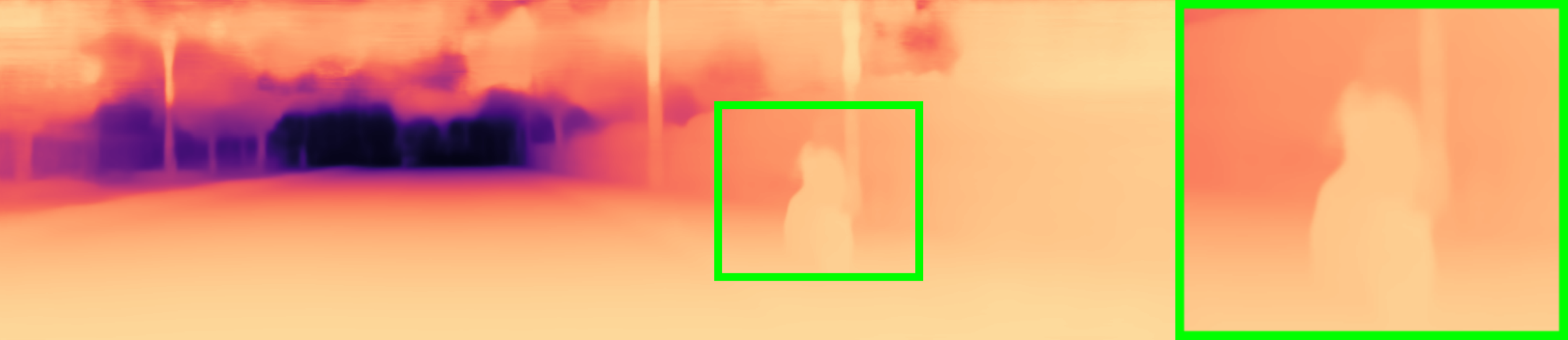}}& \vspace{0.5pt} 
                \raisebox{-0.4\height}{\includegraphics[width=0.3\textwidth, height=0.07\textheight]{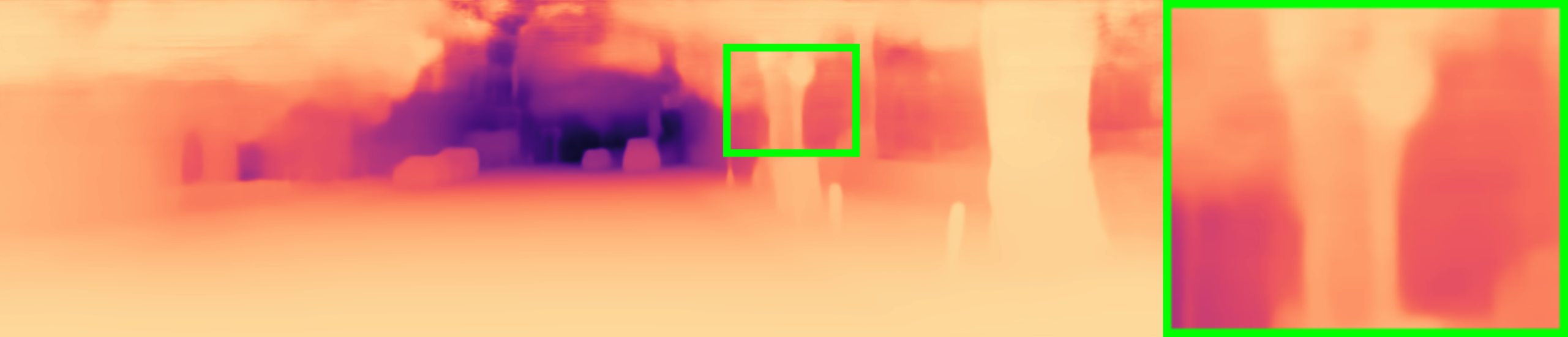}}& \vspace{0.5pt} 
                \raisebox{-0.4\height}{\includegraphics[width=0.3\textwidth, height=0.07\textheight]{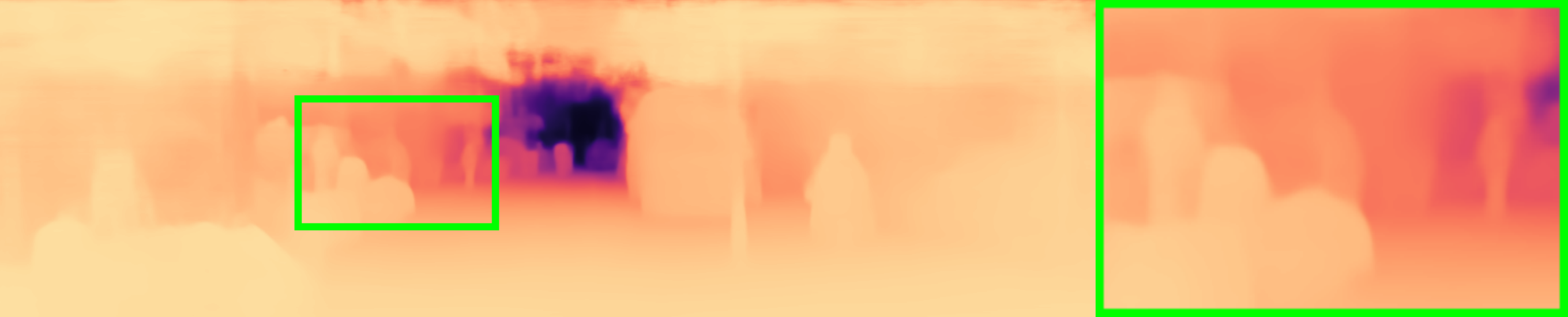}} \\ 
            & $L_{depth}$\,&  
                \raisebox{-0.4\height}{\includegraphics[width=0.3\textwidth, height=0.07\textheight]{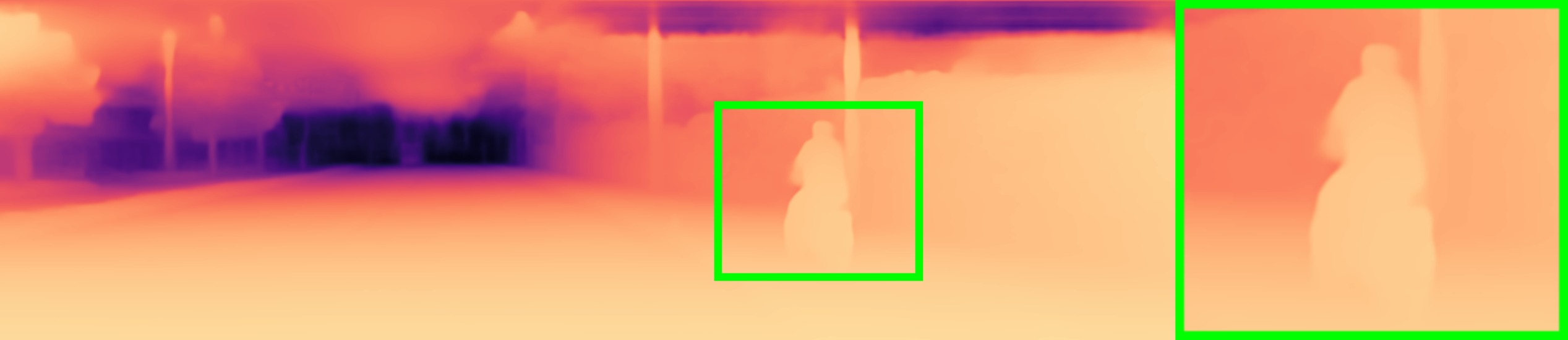}}& \vspace{0.5pt} 
                \raisebox{-0.4\height}{\includegraphics[width=0.3\textwidth, height=0.07\textheight]{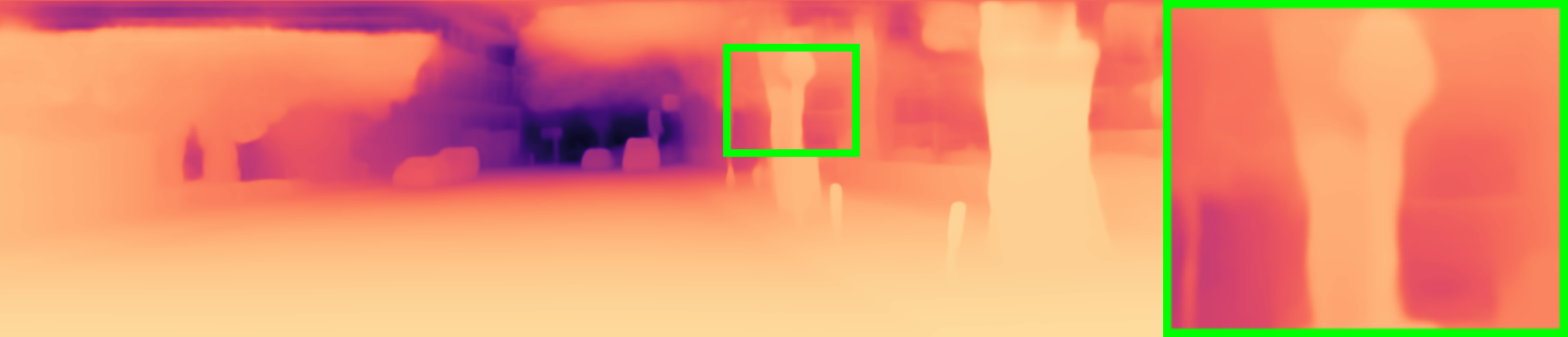}}& \vspace{0.5pt} 
                \raisebox{-0.4\height}{\includegraphics[width=0.3\textwidth, height=0.07\textheight]{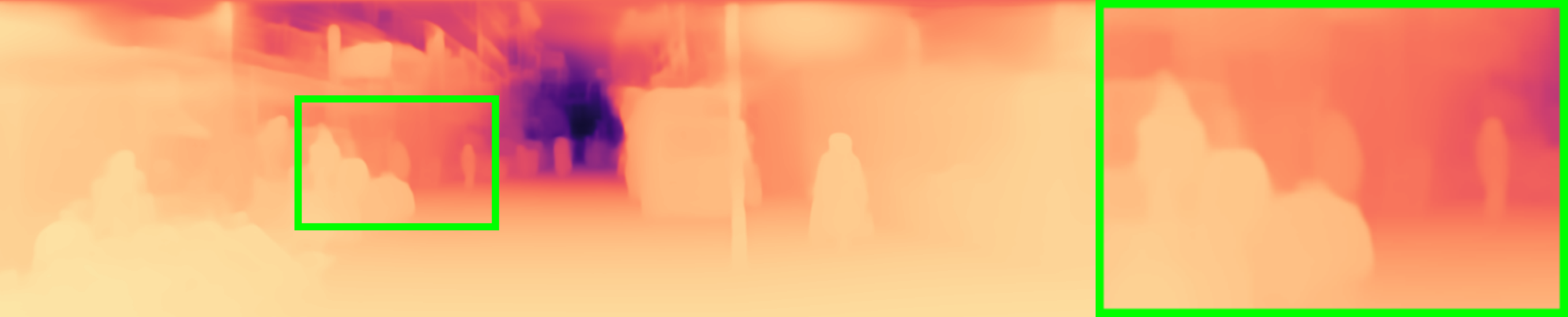}} \\ 
            & $L_{DPM, depth}$\,&  
                \raisebox{-0.4\height}{\includegraphics[width=0.3\textwidth, height=0.07\textheight]{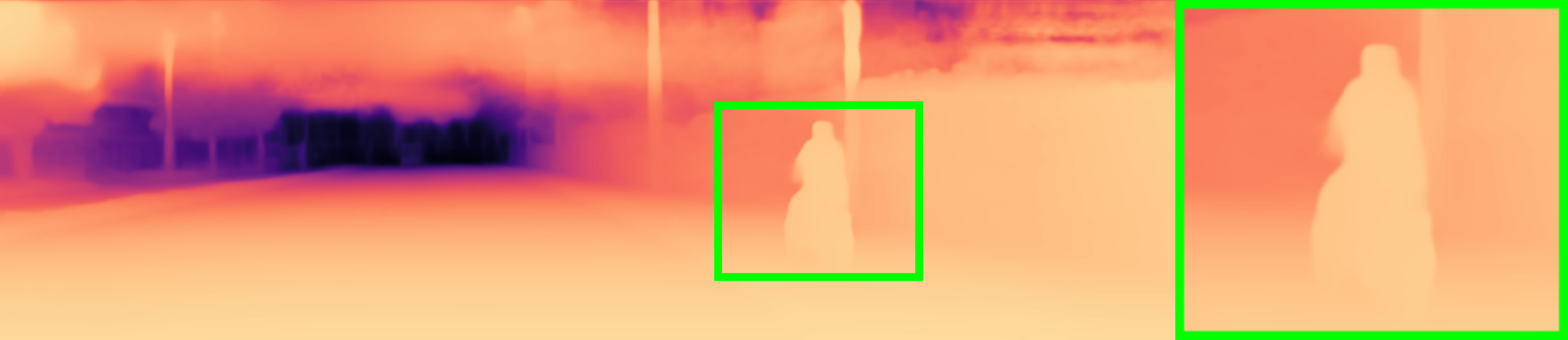}}& \vspace{0.5pt} 
                \raisebox{-0.4\height}{\includegraphics[width=0.3\textwidth, height=0.07\textheight]{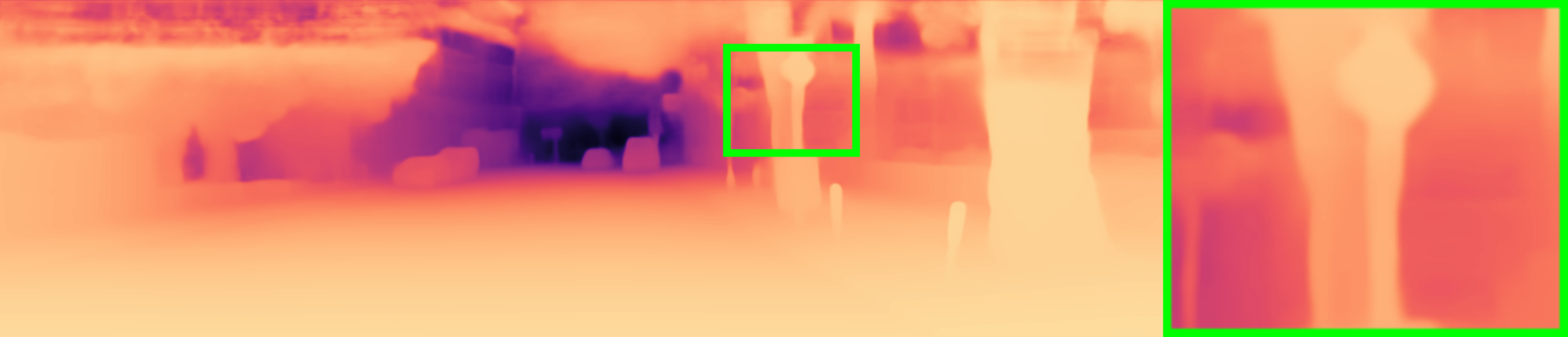}}& \vspace{0.5pt} 
                \raisebox{-0.4\height}{\includegraphics[width=0.3\textwidth, height=0.07\textheight]{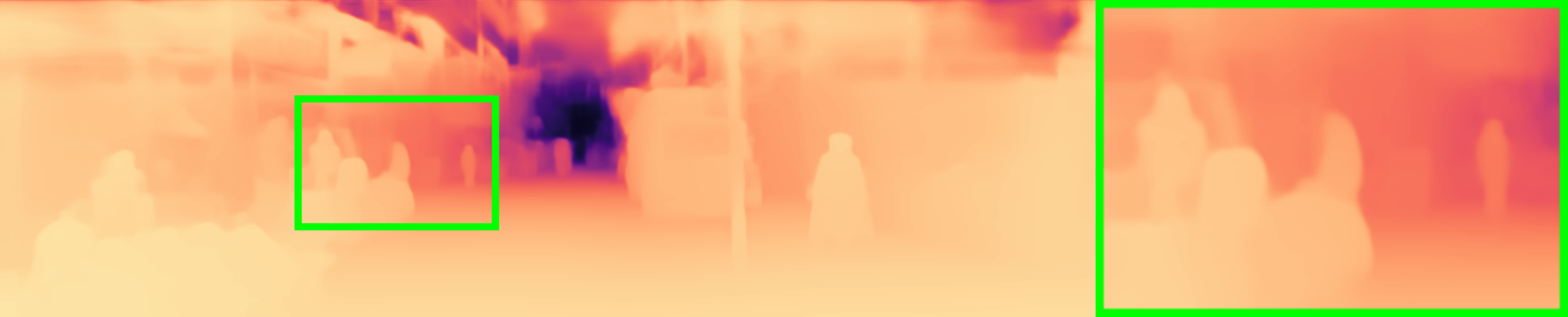}} \\ 

    \end{tabular}
    }
    \caption{Visual comparison of depth maps produced by different models for three distinct scenes, with a focus on detail variations within areas marked by green boxes.
    The top row presents the input images.
    The second row illustrates the depth maps generated by the teacher model, here BTS~\cite{lee2019big}.
    The third row depicts the baseline model's output.
    Subsequent rows display the results of the student models trained using various response-based knowledge distillation methods (Res-KD) with different loss function combinations, and the bottom rows show the depth maps from students trained using our proposed TIE-KD framework with different loss function configurations.}
    \label{fig:bts_results}
\end{figure*}
\newpage

\subsection{Teacher model: DepthFormer~\cite{li2023depthformer}}
\begin{figure*}[hb!]
    \centering
    \resizebox{0.84\textwidth}{!}{%
    \renewcommand{\arraystretch}{1}
    \setlength{\tabcolsep}{0pt}
    \begin{tabular}{c@{\hspace{0.5em}}@{\hspace{0.5em}}r@{\hspace{0.5em}}c@{\hspace{0.5em}}c@{\hspace{0.5em}}c}
            \multicolumn{2}{c}{ \normalsize Input\,}  & 
                \raisebox{-0.4\height}{\includegraphics[width=0.3\textwidth, height=0.07\textheight]{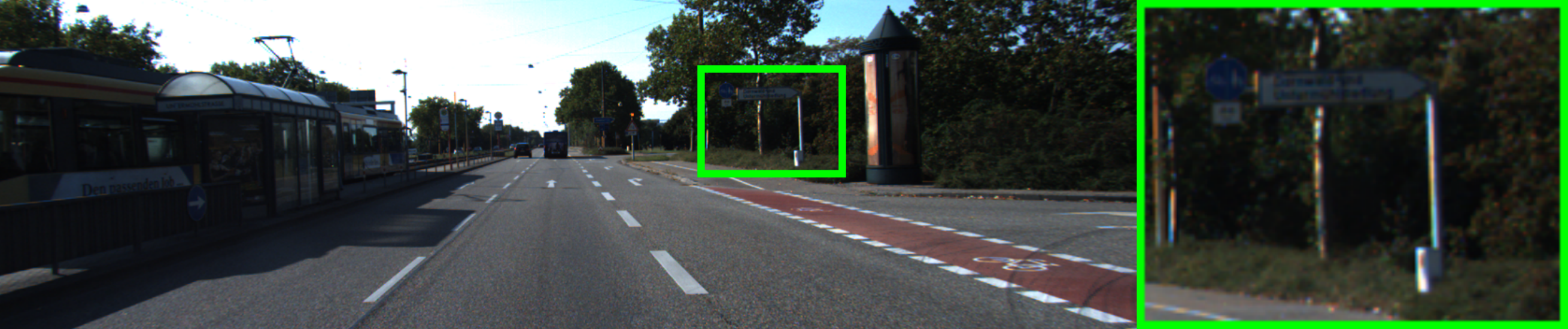}}& \vspace{0.5pt}
                \raisebox{-0.4\height}{\includegraphics[width=0.3\textwidth, height=0.07\textheight]{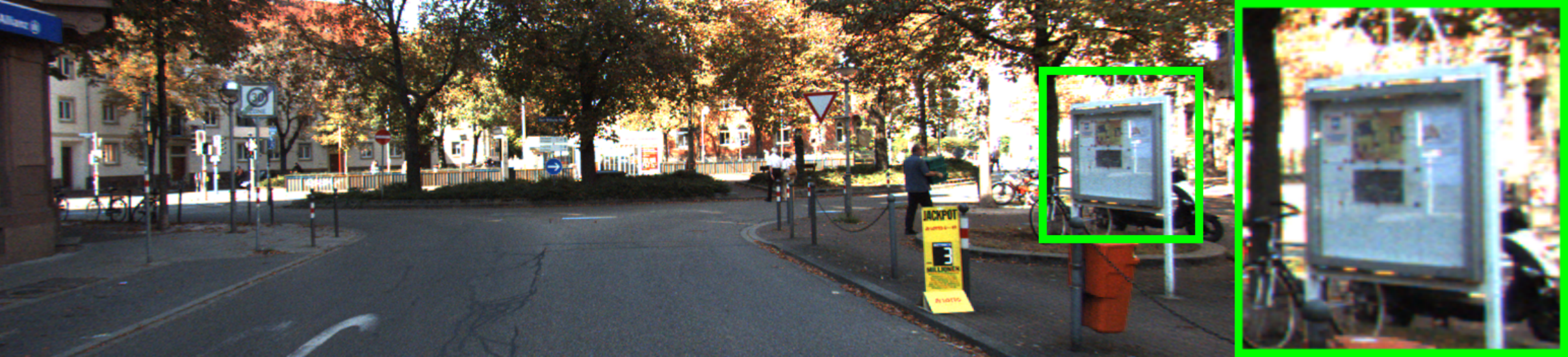}}& \vspace{0.5pt}
                \raisebox{-0.4\height}{\includegraphics[width=0.3\textwidth, height=0.07\textheight]{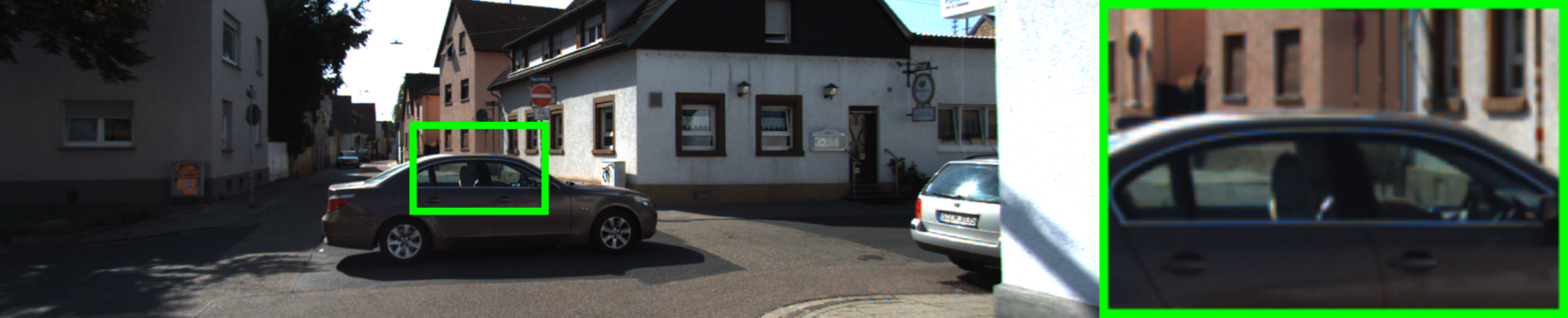}} \\ 
            \multicolumn{2}{c}{ \normalsize Baseline\,}& 
                \raisebox{-0.4\height}{\includegraphics[width=0.3\textwidth, height=0.07\textheight]{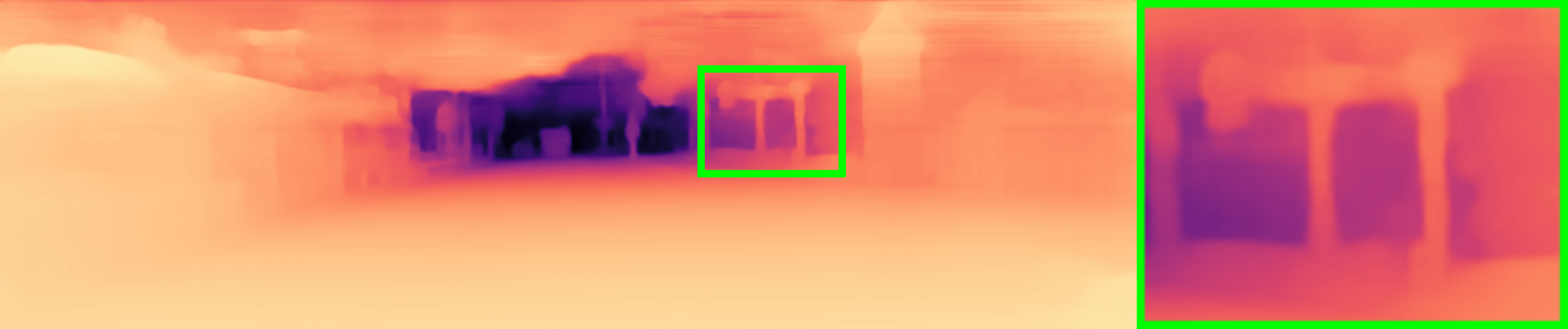}}& \vspace{0.5pt} 
                \raisebox{-0.4\height}{\includegraphics[width=0.3\textwidth, height=0.07\textheight]{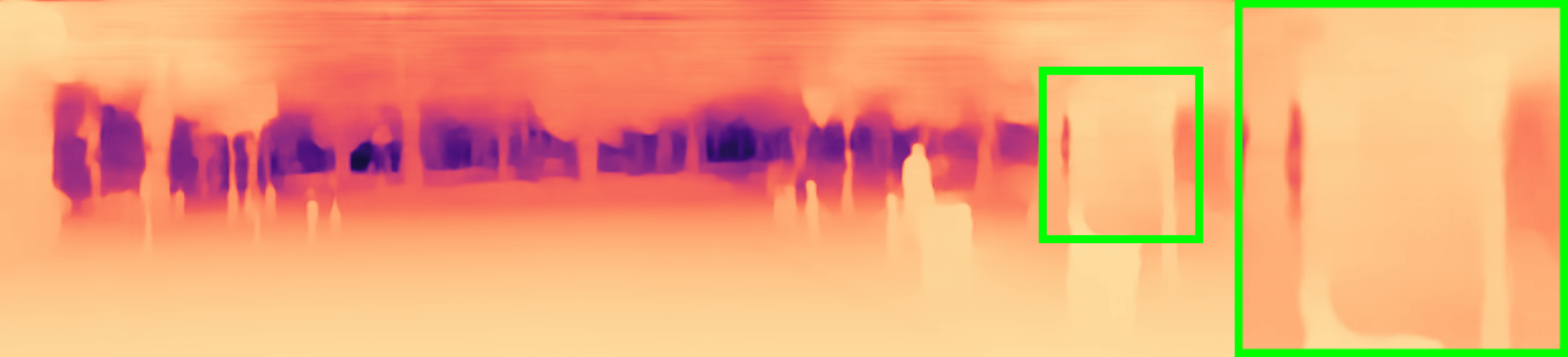}}& \vspace{0.5pt} 
                \raisebox{-0.4\height}{\includegraphics[width=0.3\textwidth, height=0.07\textheight]{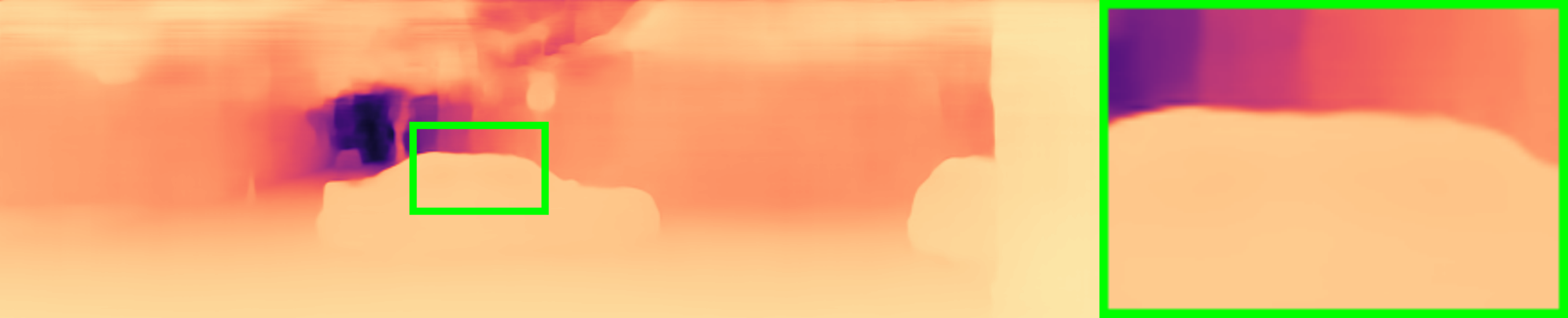}} \\ 
            \multicolumn{2}{c}{ \normalsize Teacher\,}& 
                \raisebox{-0.4\height}{\includegraphics[width=0.3\textwidth, height=0.07\textheight]{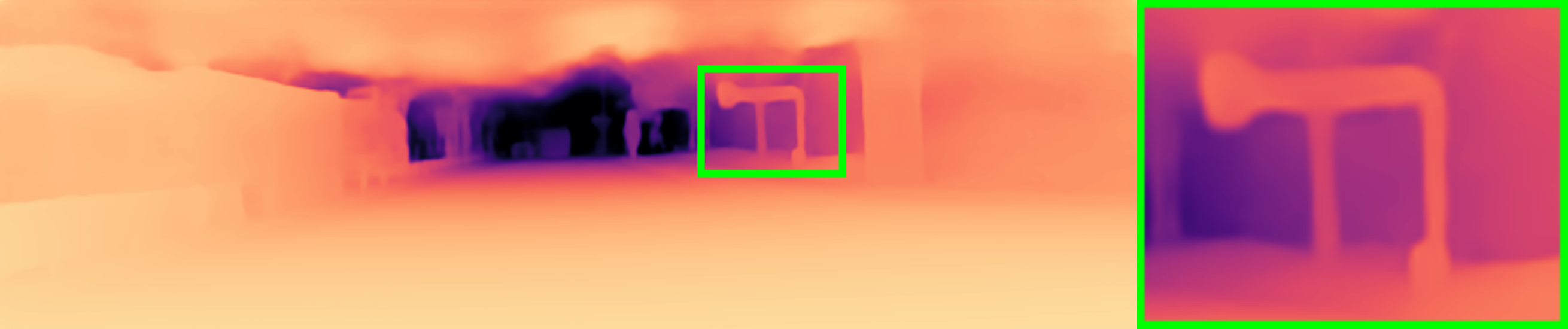}}& \vspace{0.5pt} 
                \raisebox{-0.4\height}{\includegraphics[width=0.3\textwidth, height=0.07\textheight]{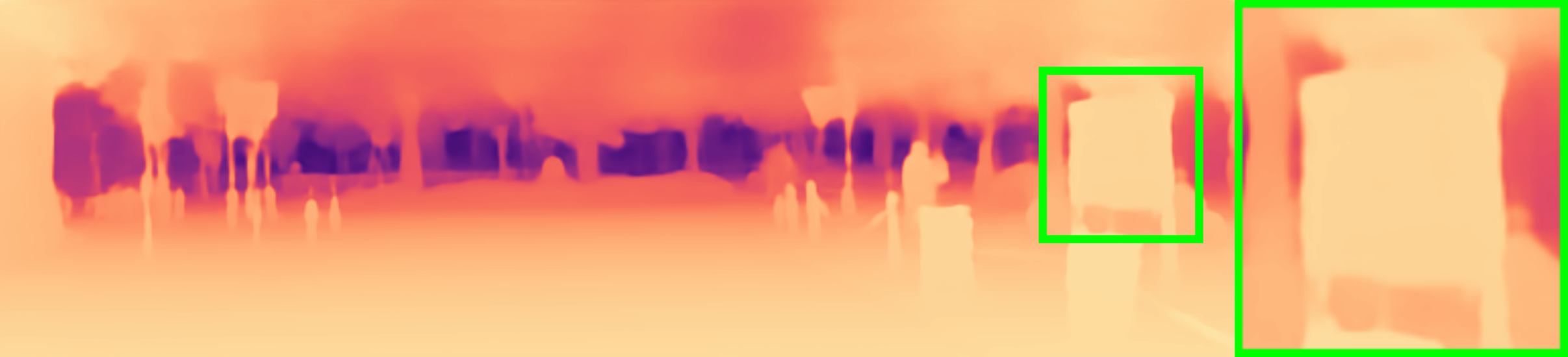}}& \vspace{0.5pt} 
                \raisebox{-0.4\height}{\includegraphics[width=0.3\textwidth, height=0.07\textheight]{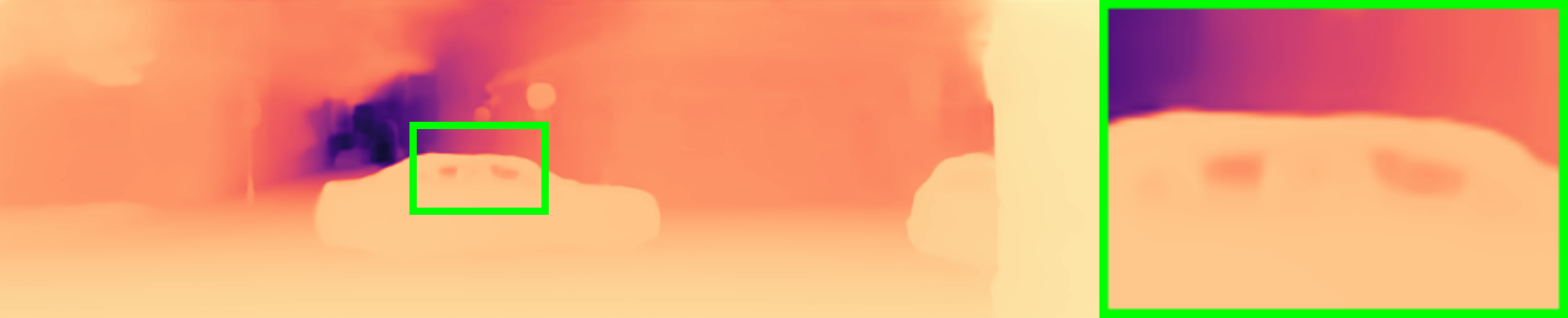}} \\ 

        \bottomrule[0.3pt]
        
        \toprule[0.3pt]
            \multirow{5}{*}{\rotatebox{90}{\parbox[c]{0.37\linewidth}{\centering \normalsize Res-KD}}} 
            &  {SSIM}\,&  
                \raisebox{-0.4\height}{\includegraphics[width=0.3\textwidth, height=0.07\textheight]{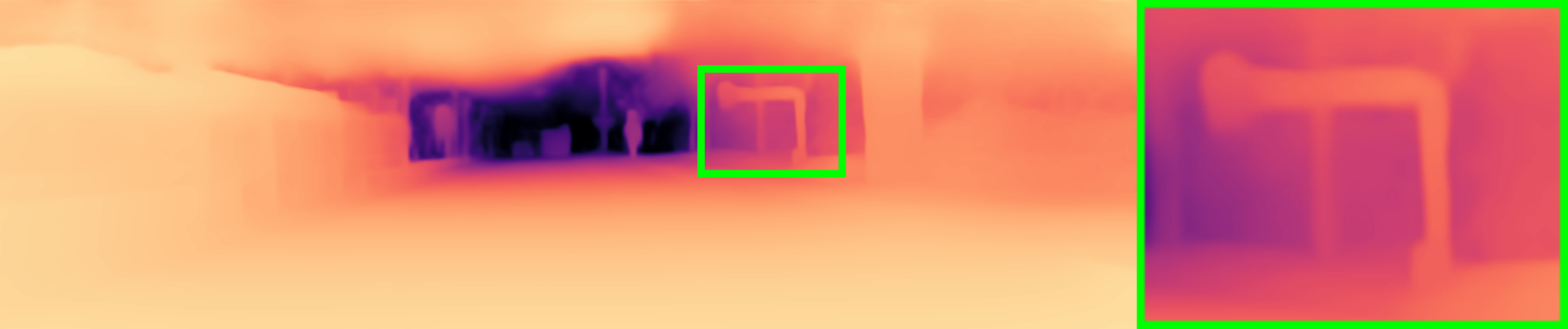}}& \vspace{0.5pt}
                \raisebox{-0.4\height}{\includegraphics[width=0.3\textwidth, height=0.07\textheight]{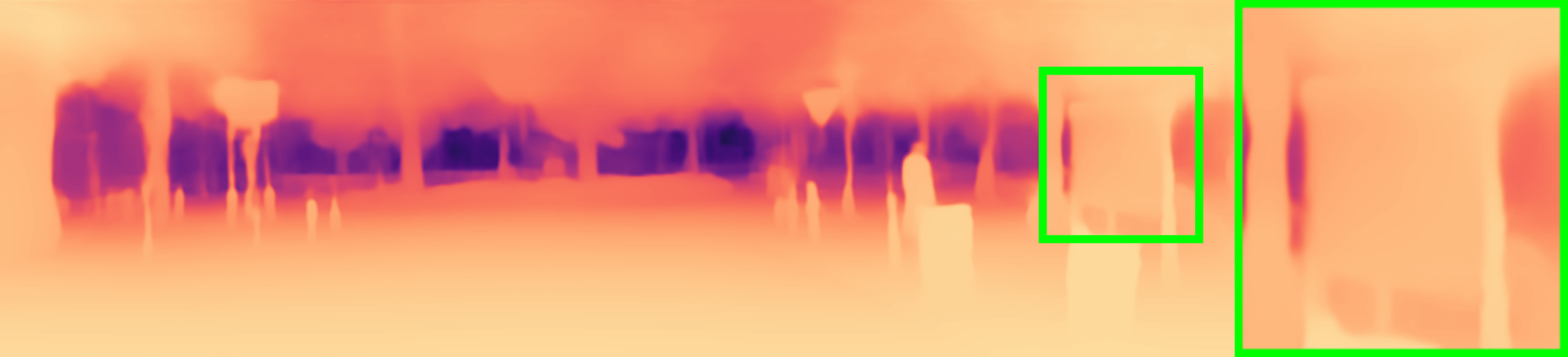}}& \vspace{0.5pt} 
                \raisebox{-0.4\height}{\includegraphics[width=0.3\textwidth, height=0.07\textheight]{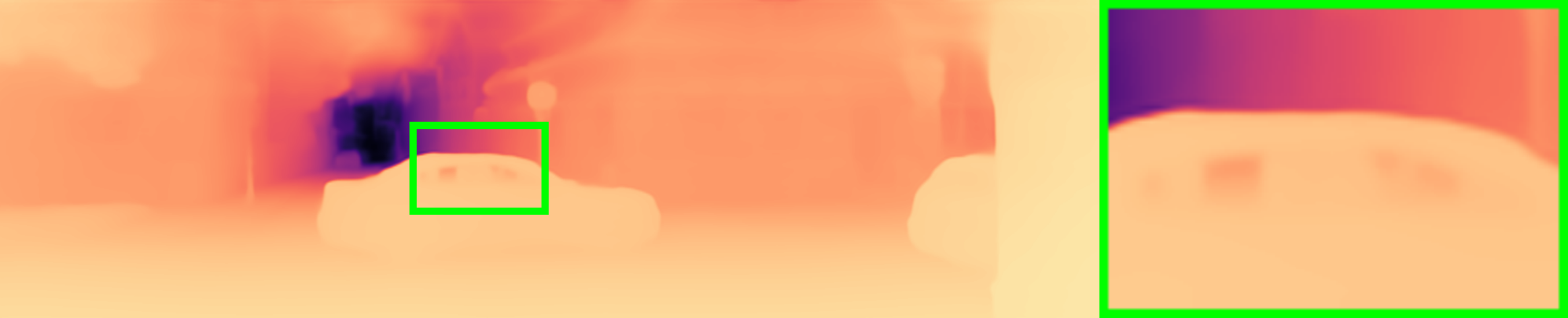}} \\ 
            &  {MSE}\,&  
                \raisebox{-0.4\height}{\includegraphics[width=0.3\textwidth, height=0.07\textheight]{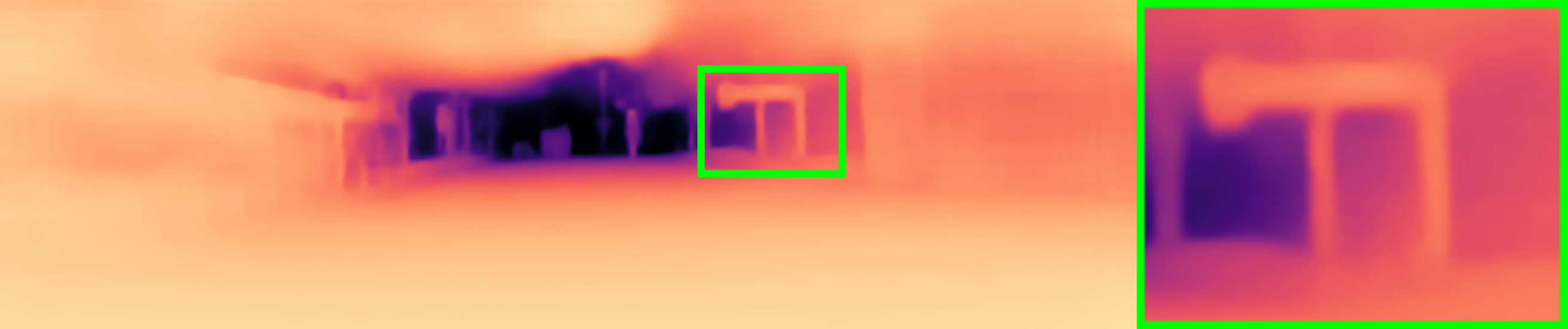}}& \vspace{0.5pt}
                \raisebox{-0.4\height}{\includegraphics[width=0.3\textwidth, height=0.07\textheight]{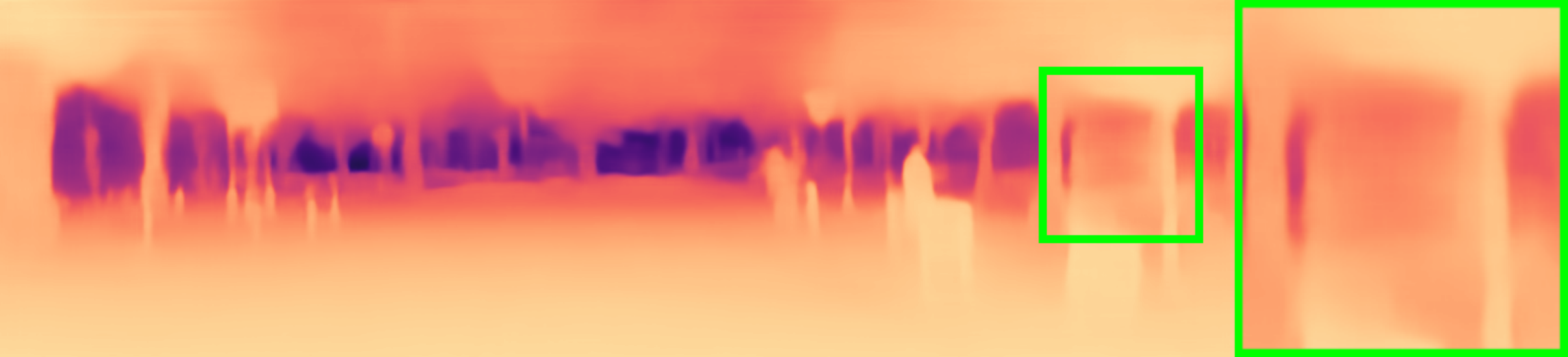}}& \vspace{0.5pt} 
                \raisebox{-0.4\height}{\includegraphics[width=0.3\textwidth, height=0.07\textheight]{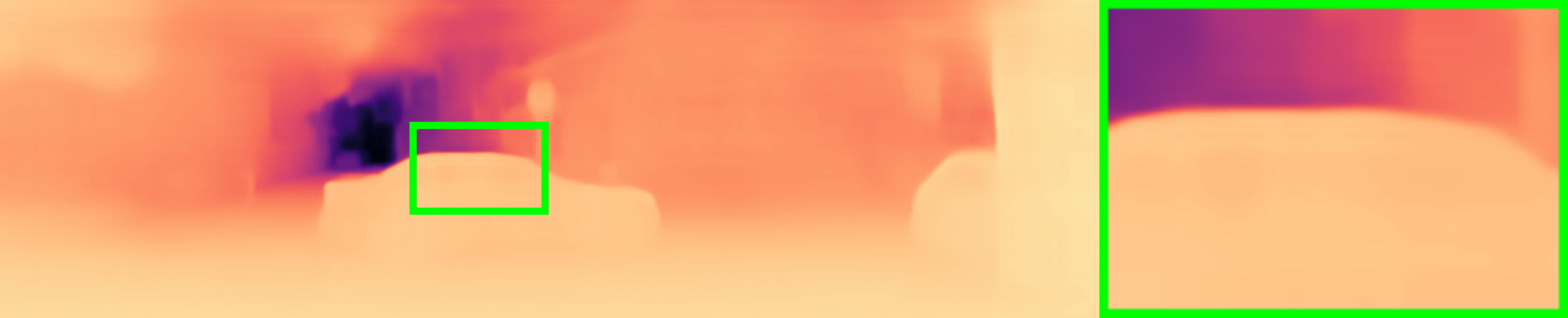}} \\ 
            &  {SI}\,&  
                \raisebox{-0.4\height}{\includegraphics[width=0.3\textwidth, height=0.07\textheight]{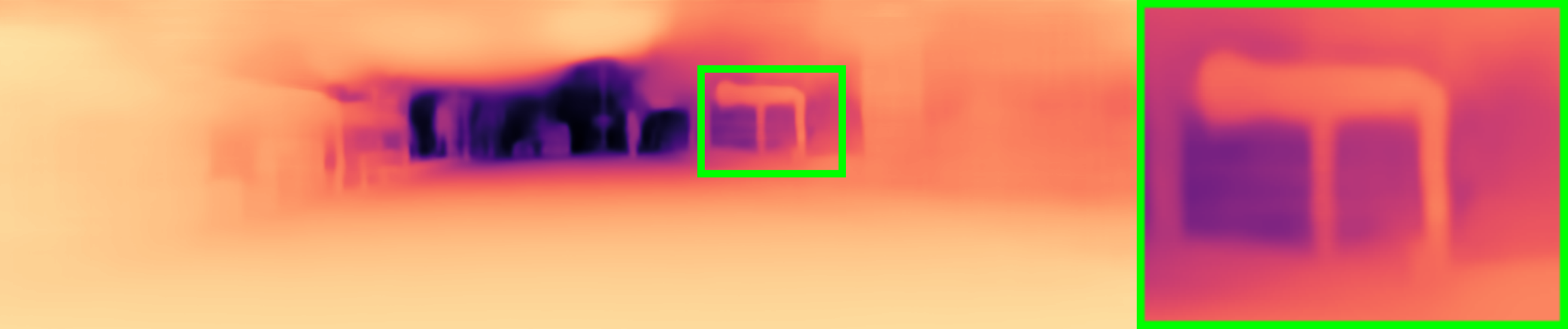}}& \vspace{0.5pt} 
                \raisebox{-0.4\height}{\includegraphics[width=0.3\textwidth, height=0.07\textheight]{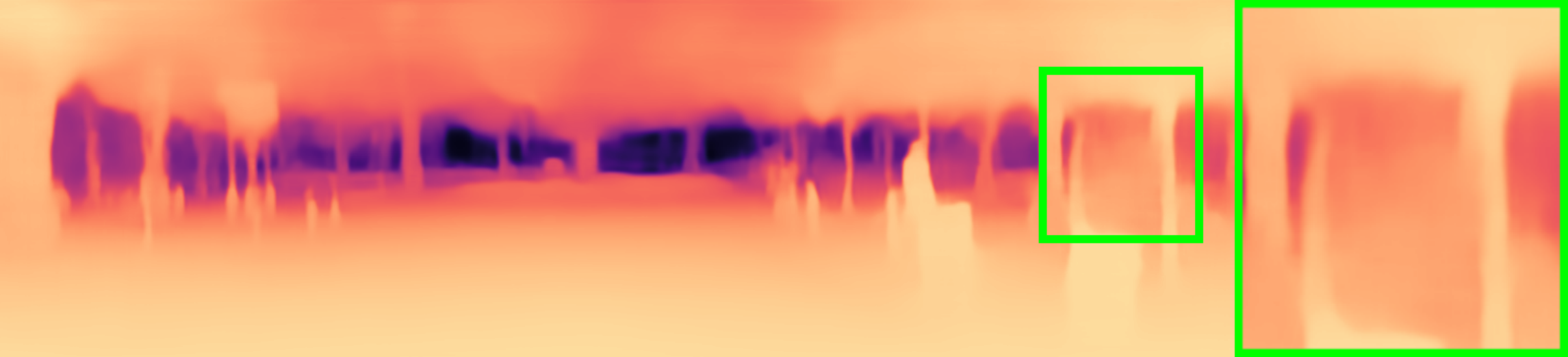}}& \vspace{0.5pt} 
                \raisebox{-0.4\height}{\includegraphics[width=0.3\textwidth, height=0.07\textheight]{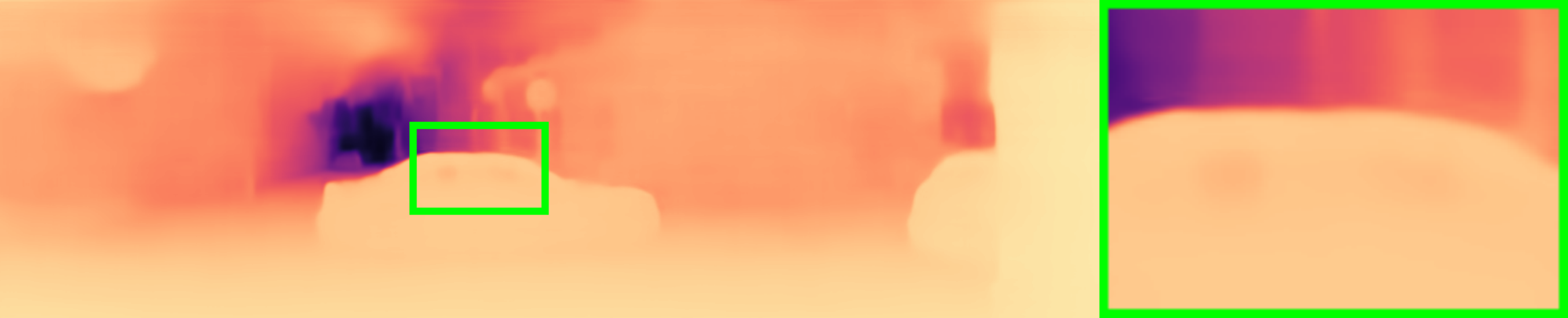}} \\ 
            &  {SSIM, SI}\,&  
                \raisebox{-0.4\height}{\includegraphics[width=0.3\textwidth, height=0.07\textheight]{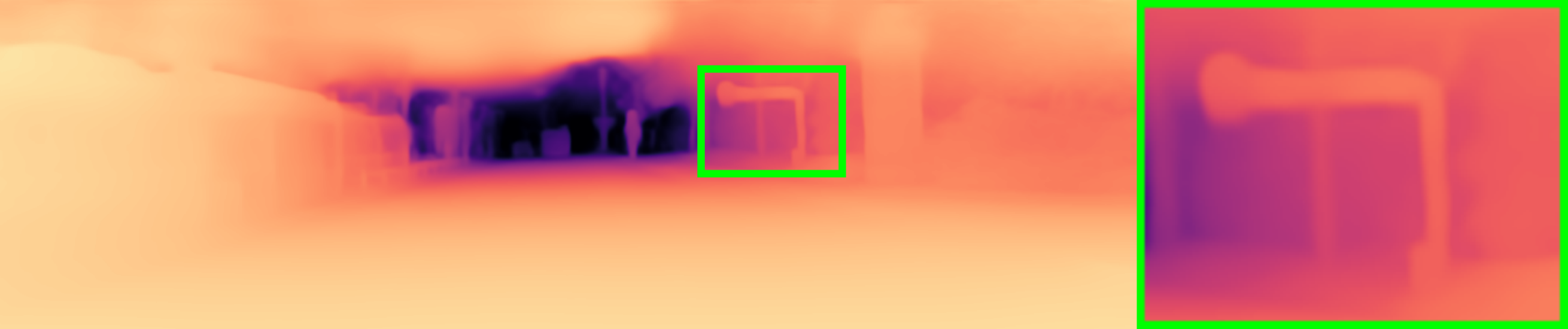}}&  \vspace{0.5pt} 
                \raisebox{-0.4\height}{\includegraphics[width=0.3\textwidth, height=0.07\textheight]{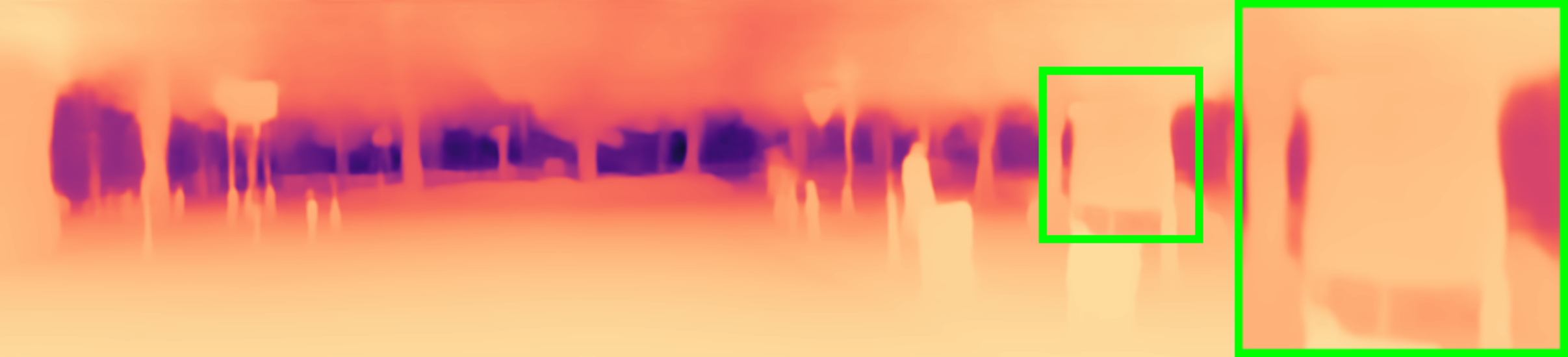}}& \vspace{0.5pt} 
                \raisebox{-0.4\height}{\includegraphics[width=0.3\textwidth, height=0.07\textheight]{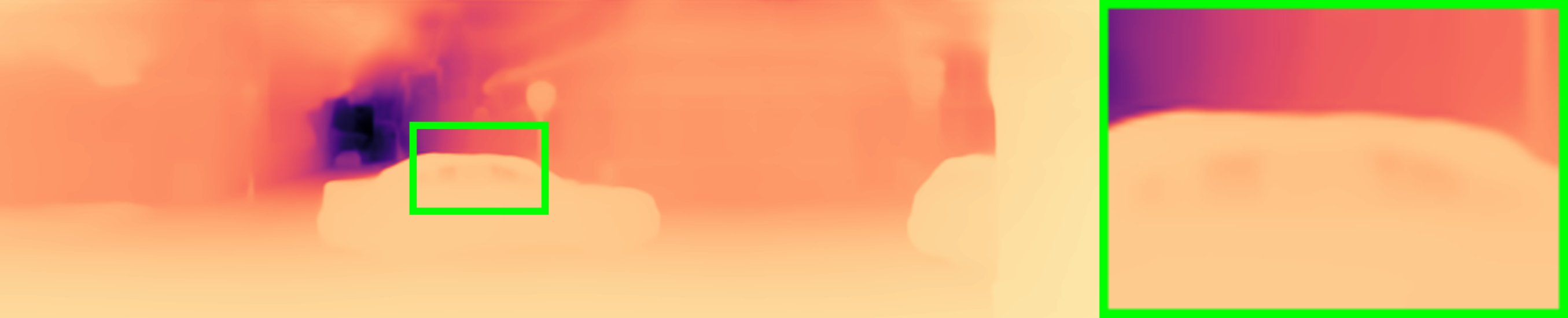}} \\ 
            &  {SSIM, MSE}\,&  
                \raisebox{-0.4\height}{\includegraphics[width=0.3\textwidth, height=0.07\textheight]{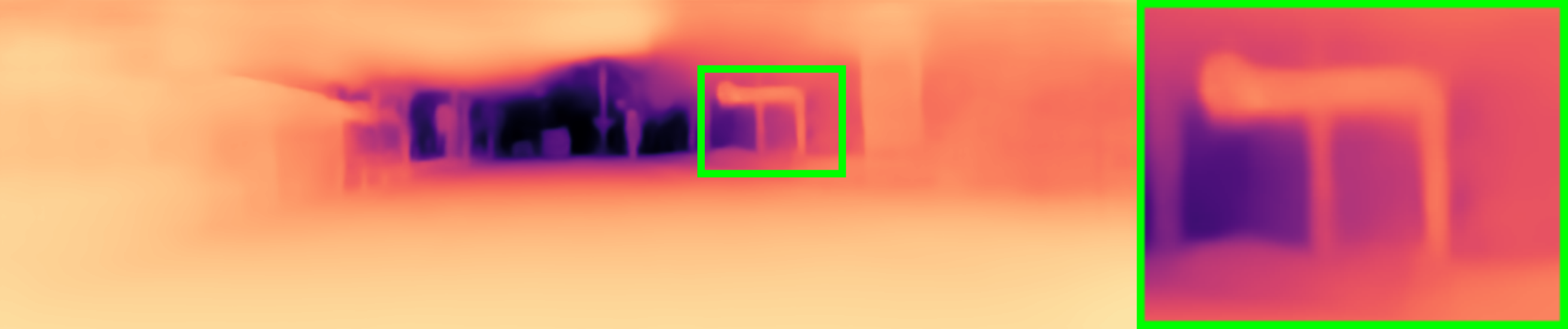}}& \vspace{0.5pt} 
                \raisebox{-0.4\height}{\includegraphics[width=0.3\textwidth, height=0.07\textheight]{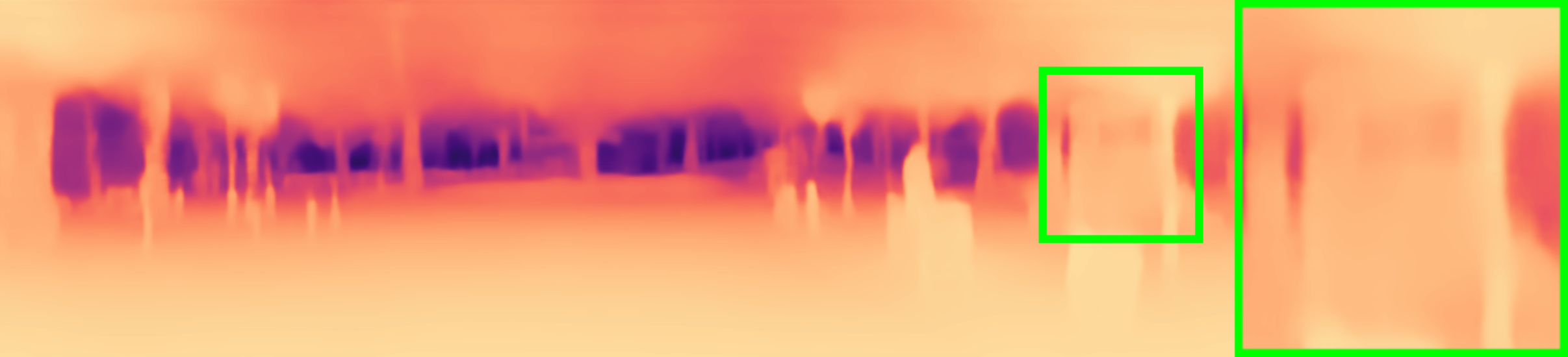}}& \vspace{0.5pt} 
                \raisebox{-0.4\height}{\includegraphics[width=0.3\textwidth, height=0.07\textheight]{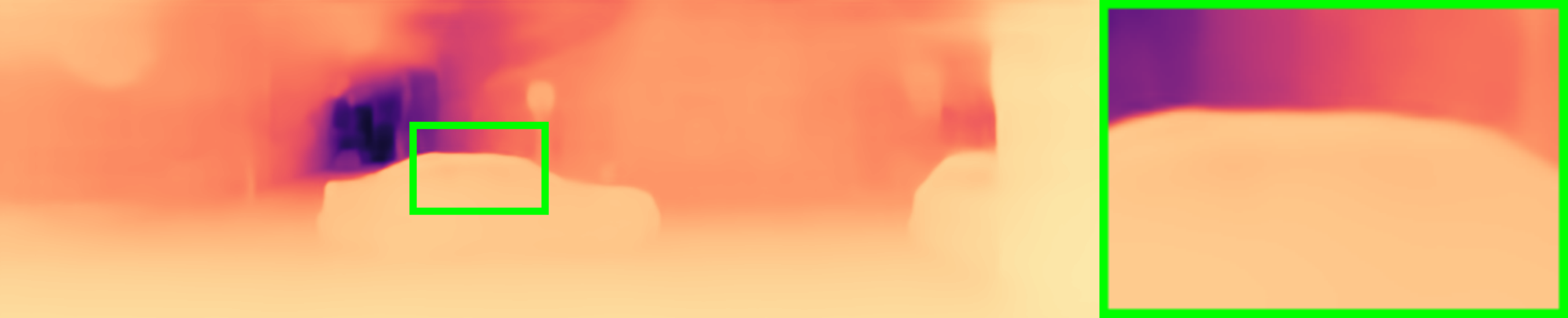}} \\ 
        \bottomrule[0.3pt]
        
        \toprule[0.3pt]
            \multirow{3}{*}{\rotatebox{90}{\parbox[c]{0.2\linewidth}{\centering \normalsize \bf{TIE-KD}}}} 
            & $L_{DPM}$\,&  
                \raisebox{-0.4\height}{\includegraphics[width=0.3\textwidth, height=0.07\textheight]{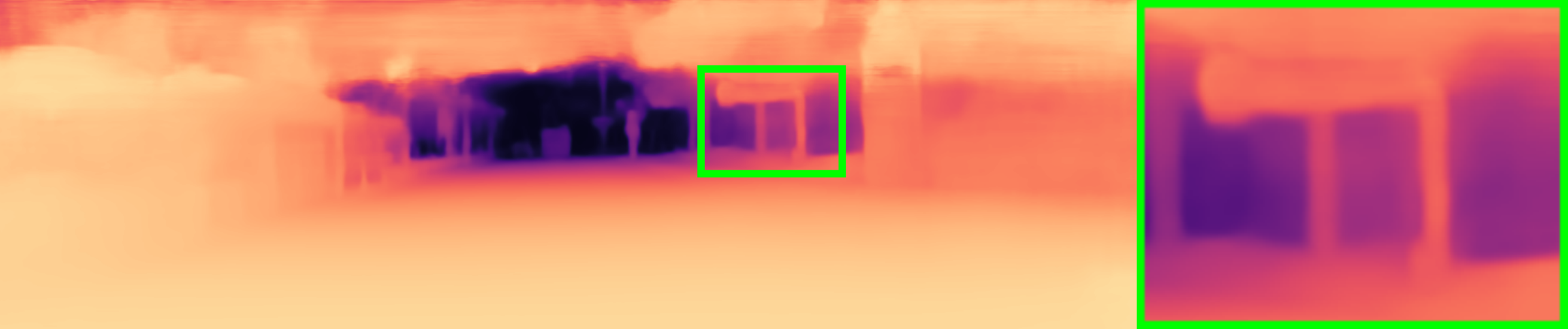}}& \vspace{0.5pt} 
                \raisebox{-0.4\height}{\includegraphics[width=0.3\textwidth, height=0.07\textheight]{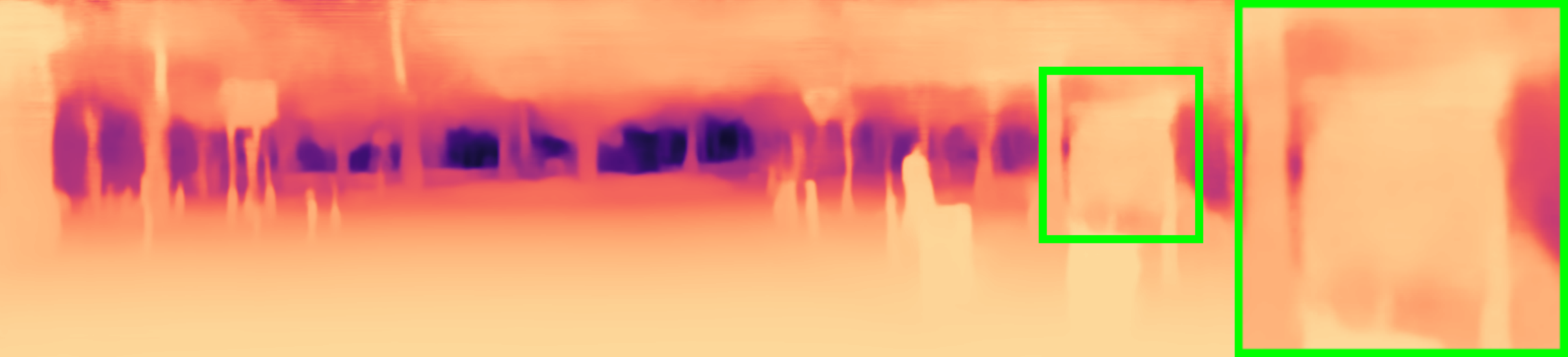}}& \vspace{0.5pt} 
                \raisebox{-0.4\height}{\includegraphics[width=0.3\textwidth, height=0.07\textheight]{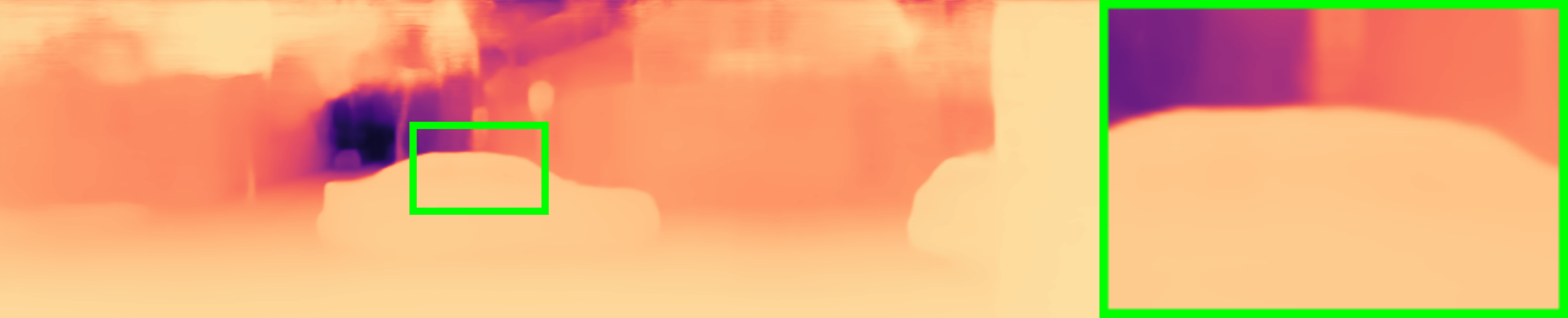}} \\ 
            & $L_{depth}$\,&  
                \raisebox{-0.4\height}{\includegraphics[width=0.3\textwidth, height=0.07\textheight]{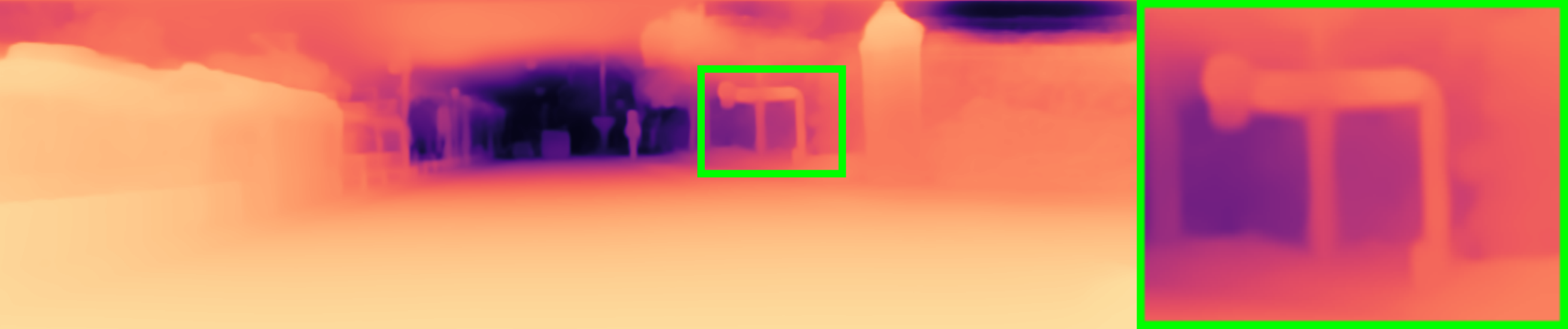}}& \vspace{0.5pt} 
                \raisebox{-0.4\height}{\includegraphics[width=0.3\textwidth, height=0.07\textheight]{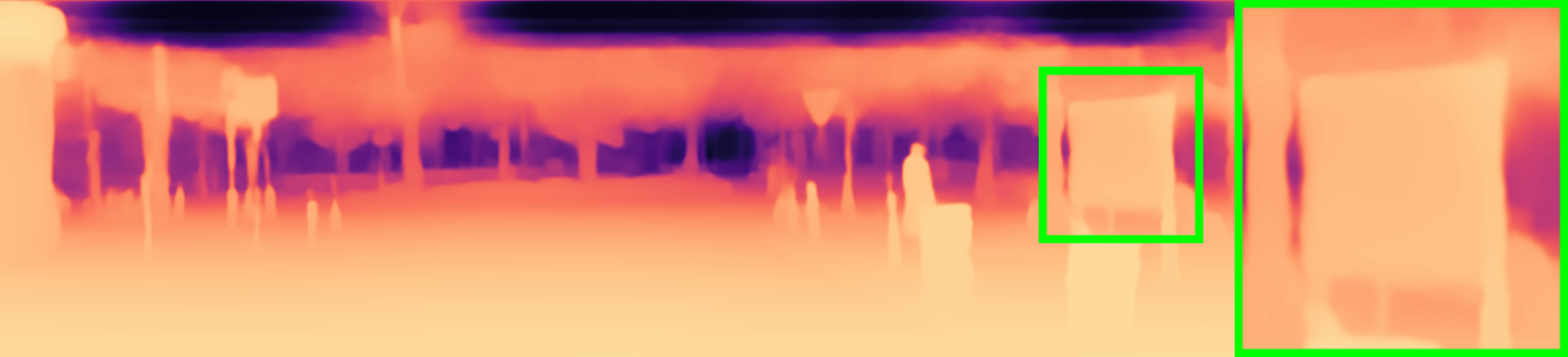}}& \vspace{0.5pt} 
                \raisebox{-0.4\height}{\includegraphics[width=0.3\textwidth, height=0.07\textheight]{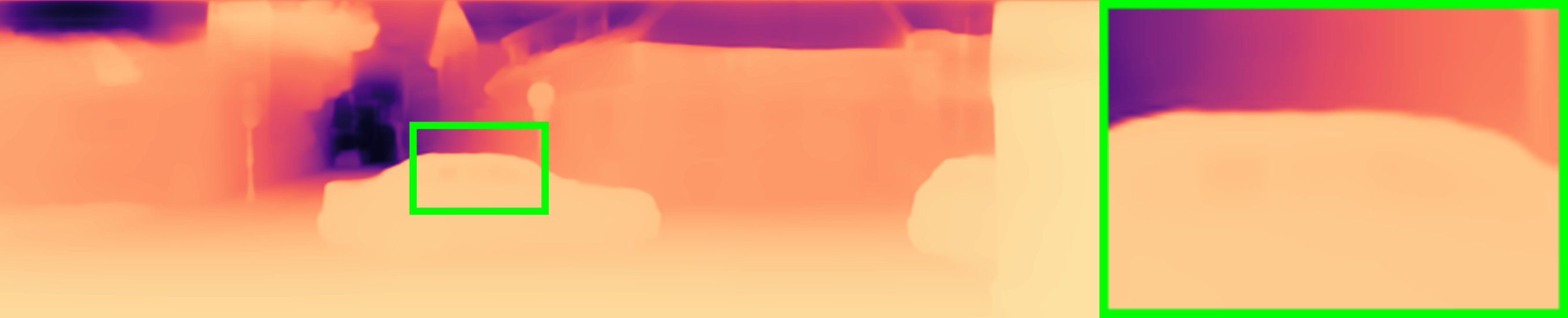}} \\ 
            & $L_{DPM, depth}$\,&  
                \raisebox{-0.4\height}{\includegraphics[width=0.3\textwidth, height=0.07\textheight]{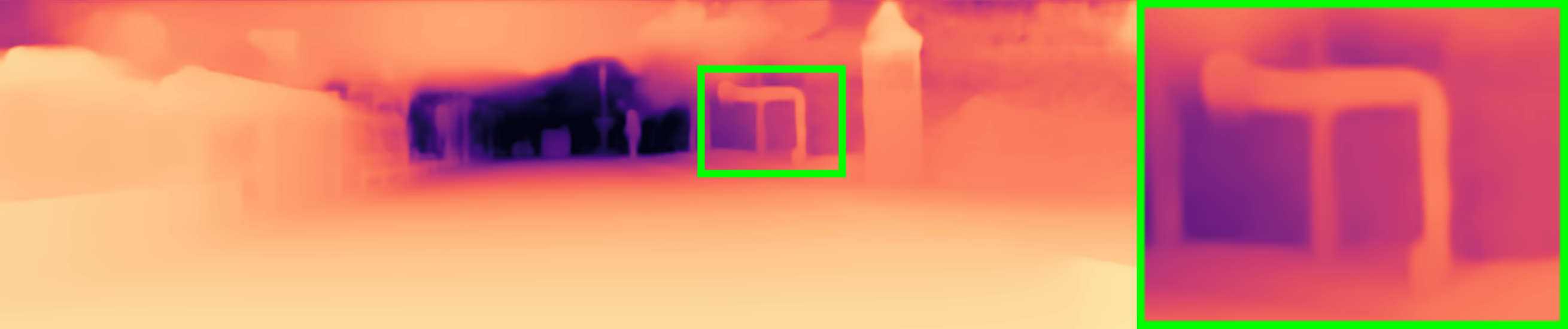}}& \vspace{0.5pt} 
                \raisebox{-0.4\height}{\includegraphics[width=0.3\textwidth, height=0.07\textheight]{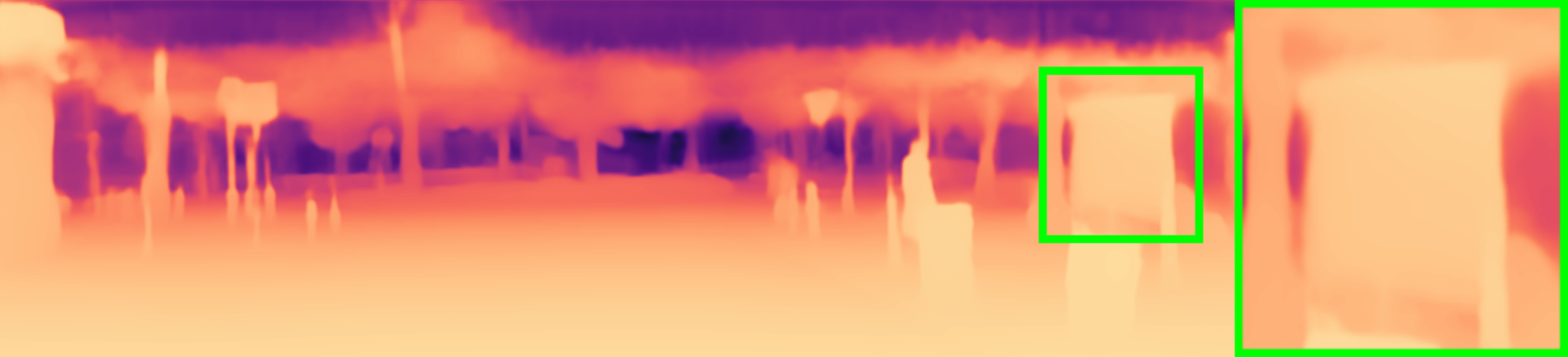}}& \vspace{0.5pt} 
                \raisebox{-0.4\height}{\includegraphics[width=0.3\textwidth, height=0.07\textheight]{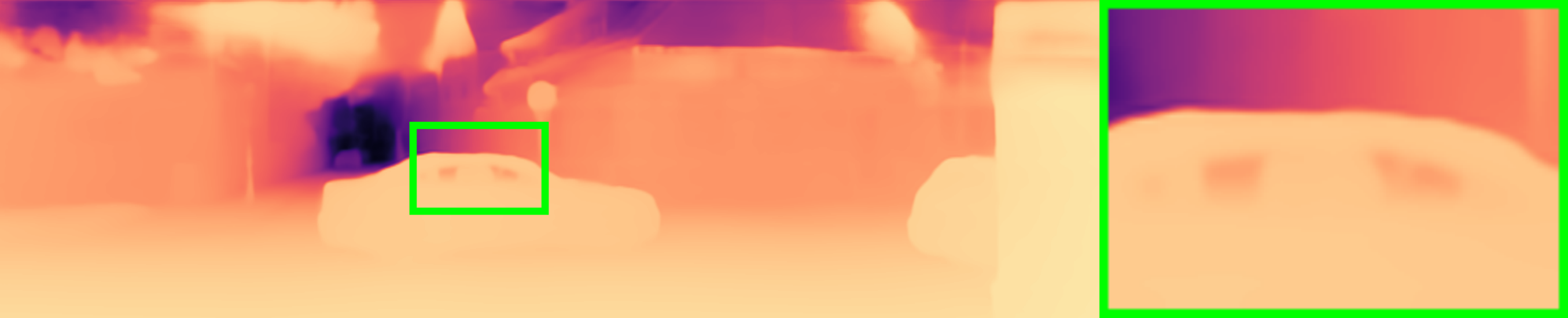}} \\ 

    \end{tabular}
    }
    \caption{Visual comparison of depth maps produced by different models for three distinct scenes, with a focus on detail variations within areas marked by green boxes.
    The top row presents the input images.
    The second row illustrates the depth maps generated by the teacher model, here DepthFormer~\cite{li2023depthformer}.
    The third row depicts the baseline model's output.
    Subsequent rows display the results of the student models trained using various response-based knowledge distillation methods (Res-KD) with different loss function combinations, and the bottom rows show the depth maps from students trained using our proposed TIE-KD framework with different loss function configurations.}
    \label{fig:depthformer_results}
\end{figure*}
\newpage
\twocolumn

\end{document}